\documentclass[Afour,sageh,times]{Format/sagej}

\usepackage{moreverb}
\usepackage[hyphens,spaces,obeyspaces]{url}

\usepackage[printonlyused,withpage,nolist,nohyperlinks]{acronym}
\usepackage[colorlinks,bookmarksopen,bookmarksnumbered,citecolor=red,urlcolor=red,draft]{hyperref}
\usepackage[ruled,vlined,linesnumbered]{algorithm2e}
\usepackage{amsfonts, amsmath, amssymb}
\usepackage{array}
\usepackage{float}
\usepackage[utf8]{inputenc} % required for names accents
\usepackage[T1]{fontenc}    % required for names accents
\usepackage{mathtools}
\usepackage{subfig}

\usepackage{dblfloatfix} % for image spanning both colums at the bottom

\usepackage{soul}

\usepackage{multicol}
\usepackage{multirow}
\usepackage{booktabs}

\usepackage{tikz}
\usetikzlibrary{shapes.geometric}

\usepackage[colorinlistoftodos,prependcaption,textsize=tiny]{todonotes}
\usepackage{color,xcolor}

\renewcommand\cite{\citep}
\makeatletter
\long\def\@makefntext#1{\parindent 1em%
\noindent{\textsuperscript{\@thefnmark}}#1}
\makeatother

\makeatletter
\def\blfootnote{\xdef\@thefnmark{}\@footnotetext}
\makeatother

\let\originalleft\left
\let\originalright\right
\renewcommand{\left}{\mathopen{}\mathclose\bgroup\originalleft}
\renewcommand{\right}{\aftergroup\egroup\originalright}

\DeclareMathOperator{\vect}{vec}

\newcommand{\identity}{\mathbb{I}}
\newcommand{\real}{\mathbb{R}}

\newcommand{\pitch}{\theta}
\newcommand{\yaw}{\psi}

\newcommand{\pitchvel}{q}
\newcommand{\yawvel}{\omega}%{r}

\newcommand{\goal}{\textrm{goal}}
\newcommand{\region}{\textrm{region}}
\newcommand{\start}{\textrm{start}}
\newcommand{\safe}{\textrm{safe}}
\newcommand{\col}{\textrm{collision}}

\newcommand{\prop}{\textrm{prop}}

\newcommand{\W}{\mathcal{W}}
\newcommand{\X}{\mathcal{X}}
\newcommand{\B}{\mathcal{B}}
\newcommand{\U}{\mathcal{U}}
\newcommand{\N}{\mathcal{N}}

\newcommand{\Xlead}{\X^\prime}
\newcommand{\Bgoal}{\B_\goal}

\newcommand{\env}{\mathcal{M}}
\newcommand{\cov}{\Sigma}
\newcommand{\traj}{\xi}
\newcommand{\locmap}{\mathcal{LM}}

\newcommand{\voxel}{\mathbf{v}}

\newcommand{\timestamp}{k}
\newcommand{\kernel}{\mathcal{K}}

\newcommand{\vx}{\mathbf{x}}

\newcommand{\vu}{\mathbf{u}}
\newcommand{\vw}{\mathbf{w}}
\newcommand{\vz}{\mathbf{z}}
\newcommand{\vr}{\mathbf{r}}

\newcommand{\belief}{\textrm{b}}

\newcommand{\reading}{\mathbf{h}}

\newcommand{\globalframe}{W}

\newcommand{\baseframe}{R}
\newcommand{\planningframe}{R^\prime}
\newcommand{\randomframe}{Y}

\newcommand{\kernelsize}{m}
\newcommand{\confidencelevel}{\alpha}
\newcommand{\criticalvalue}{t}
\newcommand{\mapresolution}{h}

\newcommand{\planningtime}{\Delta T_{MP}}
\newcommand{\planningtimescout}{\Delta T_{L}}
\newcommand{\planningtimetough}{\Delta T_{C}}

\newcommand{\jacobian}{\mathbf{J}}

\newcommand*{\sref}[1]{Section~\ref{#1}}
\newcommand*{\xref}[1]{Appendix~\ref{#1}}

\newcommand*{\eref}[1]{(\ref{#1})}
\newcommand*{\fref}[1]{Figure~\ref{#1}}
\newcommand*{\tref}[1]{Table~\ref{#1}}

\newcommand{\epa}[1]{\textcolor{magenta}{#1}}

\def\journalname{International Journal of Robotics Research}
\def\volumeyear{2020}

\begin{document}
    \runninghead{Pairet et al.}

    %\title{Online Mapping and Motion Planning under Localisation, Motion and Environmental Uncertainty}
    % \title{Online Mapping and Motion Planning under Uncertainty for Probabilistically Safe Autonomous Navigation in Unknown Environments}
    \title{Online Mapping and Motion Planning under Uncertainty for Safe Navigation in Unknown Environments}

    \author{{\`E}ric Pairet\affilnum{1}, Juan David Hern{\'a}ndez\affilnum{2}, Marc Carreras\affilnum{3}, Yvan Petillot\affilnum{1}, Morteza Lahijanian\affilnum{4}}

    \affiliation{\affilnum{1}Institute of Perception, Action and Behaviour, Edinburgh Centre for Robotics, University of Edinburgh and Heriot-Watt University, UK\\
    \affilnum{2}Department of Computer Sciences, Rice University, TX, USA\\
    \affilnum{3}Computer Vision and Robotics Institute, Univeristy of Girona, Spain \\
    \affilnum{4}Department of Aerospace Engineering Sciences, University of Colorado Boulder, CO, USA
    }

    \corrauth{{\`E}ric Pairet,
    Edinburgh Centre for Robotics,
    University of Edinburgh and Heriot-Watt University,
    Edinburgh, UK.}
    \email{eric.pairet@ed.ac.uk}

    \begin{abstract}
        Safe autonomous navigation is an essential and challenging problem for robots operating in highly unstructured or completely unknown environments. Under these conditions, not only robotic systems must deal with limited localisation information, but also their manoeuvrability is constrained by their dynamics and often suffer from uncertainty. In order to cope with these constraints, this manuscript proposes an uncertainty-based framework for mapping and planning feasible motions online with probabilistic safety-guarantees. The proposed approach deals with the motion, probabilistic safety, and online computation constraints by: (i)~incrementally mapping the surroundings to build an uncertainty-aware representation of the environment, and (ii)~iteratively (re)planning trajectories to goal that
        % are feasible (according to the robot's kinodynamic constraints) and probabilistically-safe (according to the uncertainties in the vehicle's localisation, vehicle's motion and environment awareness).
        are kinodynamically feasible and probabilistically safe through a multi-layered sampling-based planner in the belief space.
        In-depth empirical analyses
        % via
        % thorough experimentation conducted in simulation
        % simulation and experimentation
        illustrate some important properties of this approach, namely, (a)~the multi-layered planning strategy enables rapid exploration of the high-dimensional belief space while preserving asymptotic optimality and completeness guarantees, and
        % (b)~the significance of the proposed tighter bound on the computation of the probability of collision in contrast to other uncertainty-aware planners in the literature.
        (b)~the proposed routine for probabilistic collision checking results in tighter probability bounds in comparison to other uncertainty-aware planners in the literature.
        Furthermore, real-world in-water experimental evaluation on a non-holonomic torpedo-shaped \acl{AUV} and simulated trials in the Stairwell scenario of the DARPA Subterranean Challenge 2019 on a quadrotor \acl{UAV} demonstrate the efficacy of the method as well as its suitability for systems with limited on-board computational power.
    \end{abstract}


    \begin{acronym}
    \acro{2D}{\textit{two-dimensional}}
    \acro{3D}{\textit{three-dimensional}}
    \acro{AUV}{\textit{autonomous underwater vehicle}}
    \acro{BFS}{\textit{breadth-first search}}
    \acro{B-Space}{\textit{belief space}}
    \acro{BRM}{\textit{belief roadmap}}
    \acro{C-Space}{\textit{configuration space}}
    \acro{CIRS}{\textit{Underwater Vision and Robotics Research Centre}}
    \acro{CC-RRT}{\textit{chance constrained RRT}}
    \acro{COLA2}{\textit{component oriented layer-based architecture for autonomy}}
    \acro{CL-RRT}{\textit{closed-loop RRT}}
    \acro{DoF}{\textit{degree of freedom}}
    \acro{DVL}{\textit{doppler velocity logger}}
    \acro{EST}{\textit{expansive-spaces tree}}
    \acro{FaSTrack}{\textit{fast and safe tracking}}
    \acro{FIRM}{\textit{feedback-based information roadmap}}
    \acro{FoV}{\textit{field of view}}
    \acro{FSF}{\textit{full state feedback}}
    \acro{GPS}{\textit{global positioning system}}
    \acro{HFR}{\textit{high-frequency replanning}}
    \acro{iMDP}{\textit{incremental Markov decision process}}
    \acro{IMU}{\textit{inertial measurement unit}}
    \acro{ITOMP}{\textit{incremental trajectory optimization for motion planning}}
    \acro{KF}{\textit{Kalman filter}}
    \acro{LIDAR}{\textit{Light Detection and Ranging}}
    \acro{LoS}{\textit{line-of-sight}}
    \acro{LQG}{\textit{linear quadratic Gaussian}}
    \acro{LQR}{\textit{linear quadratic regulator}}
    \acro{LTI}{\textit{linear time-invariant}}
    \acro{NOAA}{\textit{National Oceanic and Atmospheric Administration}}
    \acro{MDP}{\textit{Markov decision process}}
    \acro{MLP}{\textit{multi-layered planner}}
    \acro{MSIS}{\textit{mechanical scanned imaging sonar}}
    \acro{OMPL}{\textit{open motion planning library}}
    \acro{PD}{\textit{proportional derivative}}
    \acro{POMDP}{\textit{partially observable Markov decision process}}
    \acro{PRM}{\textit{probabilistic roadmap}}
    \acro{RHC}{\textit{receding horizon control}}
    \acro{ROS}{\textit{robot operating system}}
    \acro{ROV}{\textit{remotely operated vehicle}}
    \acro{RRT}{\textit{rapidly-exploring random tree}}
    \acro{RRT*}{\textit{asymptotic optimal RRT}}
    \acro{RViz}{\textit{ROS visualizer}}
    \acro{SLAM}{\textit{simultaneous localisation and mapping}}
    \acro{SLP}{\textit{single-layered planner}}
    \acro{SMR}{\textit{stochastic motion roadmap}}
    \acro{SST}{\textit{stable sparse RRT}}
    \acro{U-Space}{\textit{control space}}
    \acro{UAV}{\textit{unmanned aerial vehicle}}
    \acro{UdG}{Universitat de Girona}
    \acro{USBL}{\textit{ultra-short base line}}
    \acro{UWSim}{\textit{underwater simulator}}
    \acro{VICOROB}{VIsi\'{o} per COmputador i ROB\`{o}tica}
    \acro{X-Space}{\textit{state space}}
\end{acronym}

    \maketitle

    % this is only for the pre-print version
    %\blfootnote{\\This manuscript is the original version submitted for review to the International Journal of Robotics Research (IJRR).}

    \section{Introduction} \label{sec:introduction}
    % ======================================
    % FIRST ACT: motivation and challenge
    % ======================================
    Autonomous robots have been increasingly employed to assist humans notably in hazardous or inaccessible environments in recent years. Examples include rescue missions in disaster response scenarios~\cite{murphy2008search}, in-water ship hull~\cite{hover2012advanced} and wind turbine quality assessment inspections~\cite{morgenthal2014quality}, underwater archaeology~\cite{bingham2010robotic}, and deep underwater and space exploration~\cite{whitcomb2000advances,galceran2015coverage,katz2003nasa}, among many others. A fundamental requirement for a robot engaged in any of these applications is to be adept at navigating autonomously through highly unstructured and hostile environments. However, this is not a trivial task due to limited or complete lack of prior knowledge about the environment in which the robot has to operate. This implies that the robot has to base its decision making on on-board sensors despite their limited accuracy. In addition, the robot itself might suffer from poor localisation, as well as restricted and uncertain manoeuvrability. Therefore, even though challenging, it is essential to jointly consider all these motion and sensory constraints as well as their associated uncertainties, 
    when planning for navigation actions. 
    % in autonomous navigation.
    This problem becomes particularly more challenging in safety-critical missions where the safety of the robot must be ensured at all times.

    % ======================================
    % SECOND ACT: related work and drama 
    % ======================================
	Although there exist alternative methodologies addressing each of the above-mentioned issues individually, limited attention has been devoted to the autonomous navigation problem in unknown environments as a whole. The classical algorithms known as \ac{SLAM} enable a mobile robot to concurrently build and use a map 
% 	while , at the same time, use the map
	to estimate its location~\cite{durrant2006simultaneous}. These algorithms rely on identifying distinctive landmarks which can bound the uncertainty of both the environment representation and the robot localisation. Nonetheless, even for scenarios rich in features, there are always some residual uncertainties. More recently, online motion planning frameworks have been developed to empower a mobile robot to compute navigation actions in unexplored environments while accounting for the system's motion capabilities, e.g.~\cite{hernandez2016planning,ho2018virtual,hernandez2019planning,vidal2019online,youakim2020multirepresentation}. These approaches, however, do not cope with any source of uncertainty and employ ad-hoc heuristics which lack quantified safety guarantees. The few attempts to ensure safety through probabilistic methods, such as~\cite{strawser2018approximate,janson2018monte,da2019collision}, are generally computationally expensive, are built on strong assumptions, and commonly suppose a complete prior knowledge of the surroundings.  Therefore, they are unsuitable for applications requiring online computations or dealing with unknown environments.
    
	In this context, our previous framework guaranteed (in compliance with a user-defined minimum probability of safety) the robot's safety when navigating through unexplored environments~\cite{pairet2018uncertainty}. The underlying strategy consisted of an iterative mapping-planning scheme capable of continuously modifying the vehicle's motion plan towards a desired goal according to the incremental environmental awareness. At any time, the resulting motion plan was guaranteed to be feasible and safe in face of localisation, mapping and motion uncertainties. This was achieved by incrementally encoding the vehicle's surroundings as an uncertainty-aware map, and by planning feasible trajectories (according to the system's kinodynamic constraints) over this representation, which provided probabilistic safety guarantees by taking into account the uncertainty on the system's localisation and motion, as well as the uncertainty on the environment awareness.
    
	Despite the satisfactory results achieved in our previous work, the framework had some limitations. Foremost, the framework was exclusively tailored to cope with \ac{2D} workspaces. The performance of the mapping-planning scheme and its constituent components would scale poorly when dealing with scenarios and systems of higher dimensionality. Furthermore, in-depth discussion, empirical analyses, and thorough experimental evaluation of the framework's key components were lacking. Finally, experimental validations were limited to scenarios with symmetric underwater structures, thus not illustrating the framework's capability to navigate through more challenging environments. Bearing the previous framework limitations in mind, this manuscript presents an extended and improved version of our framework that attempts to overcome these limitations.
    
	% ======================================
    % THIRD ACT: what do we propose? -- summary of contributions
    % ======================================
    While preserving the general mapping-planning scheme of our previous work~\cite{pairet2018uncertainty}, the main contributions of this manuscript are: 1)~the extension of the planning strategy to a multi-layered architecture, allowing for rapid search of a solution in high-dimensional belief spaces while preserving asymptotically optimal and completeness guarantees, 2)~the reformulation of the probabilistic collision checking routine, enabling the planner to efficiently evaluate the validity of a state subject to uncertainties and 
    %to account for tail events,
    to trade the tightness of the safety bound for computational efficiency, while accounting for the tail events,
    and 3)~the rigorous theoretical development and thorough experimental evaluations of the key constituent components of the framework, as well as the framework as a whole. All in all, these novel advancements allow for faster online motion planning and more efficient evaluation of uncertainties. Consequently, the framework enables the computation of navigation actions online for high-dimensional systems and more challenging unknown environments while still providing safety guarantees. To the best of the authors' knowledge, this is the first generic architecture capable of jointly dealing with kinodynamic and probabilistic constraints 
    % when planning navigation actions 
    in unknown environments online. Both the precedent and new framework are analysed and compared in multiple scenarios with different interesting real-world\footnote{A mission through a real breakwater structure with an \ac{AUV} can be seen in: \newline \url{https://youtu.be/dTejsNqNC00}.} and simulated\footnote{A mission through the DARPA Subterranean Challenge 2019 scenario with an \ac{UAV} can be seen in: \newline %\url{www.dropbox.com/s/y0jpx2a5azi41v2/ijrr.mp4?dl=0}.}}
    %\url{www.dropbox.com/s/vuqpyo5dy5tzdmw/IJRR20_v2_rs.mp4?dl=0}.} 
    \url{https://youtu.be/I5X_QFKDpeI}.}
    physical systems. The experimental results demonstrate the suitability of the proposed method to address the challenge of probabilistically-safe autonomous navigation in unknown environments while being suitable for systems with limited on-board computational power. 
    
	% ======================================
    % paper overview
    % ======================================
    The remainder of this manuscript is organised as follows. \sref{sec:background} provides a comprehensive review of the literature and the corresponding contribution of this paper. Then, \sref{sec:formulation} formally defines the considered problem.
    % and the set of assumptions under which the properties of the proposed framework hold. 
    % In \sref{sec:framework}, we present an overview of the framework itself and then detail the mapping and planning components in \sref{sec:mapping} and \sref{sec:planning}, respectively. 
    In \sref{sec:framework}, an overview of the framework is presented, and then the mapping and planning components are detailed in Sections \ref{sec:mapping} and \ref{sec:planning}, respectively. 
    The description of the framework is followed by a thorough analysis of its key constituent features and its performance and capabilities as a whole in \sref{sec:evaluation}. Finally, the paper concludes with a discussion in \sref{sec:conclusion}.
    \section{Related Work} \label{sec:background}
    This section gives a brief overview of prior work on planning under kinodynamic constraints and planning under uncertainty, as well as frameworks for online mapping-planning. Finally, this section discusses all contributions of this work with respect to the latest related literature.
    
    % ===============================
    % ===============================
    % ===============================
    \subsection{Planning under Kinodynamic Constraints}
        %A Review of Motion Planning Techniques for Automated Vehicles
        %https://ieeexplore.ieee.org/stamp/stamp.jsp?tp=&arnumber=7339478
        
        Planning under kinodynamic constraints deals with the challenge of computing trajectories which are feasible according to the vehicle's motion capabilities. This problem is commonly formulated as finding a trajectory between two points through the system's state space. The robotics literature offers various approaches to tackle this problem.

        One strategy is to represent the continuous state space as a lattice space, i.e., a graph where edges correspond to a reduced set of precomputed motion primitives. Then, the motion planning problem can be efficiently solved using graph search algorithms. For the particular case of a car-like system, the motion primitives can be defined as a set of lines and arcs to build a geometric state lattice \cite{dubins1957curves,reeds1990optimal}. These approaches can find the shortest path, but the transition between segments presents abrupt changes in angular velocity, which could only be achieved by a system capable of infinite angular acceleration. More complex lattice space definitions allow the consideration of more restrictive concatenation rules and richer sets of primitive motions, e.g. \cite{frazzoli2005maneuver,pivtoraiko2011kinodynamic}, at the cost of more memory usage and more computationally expensive queries. Even though planning in lattice spaces has proven to be suitable for many  applications, it requires the crafting of a set of motions such that the resulting lattice offers, at least, one suitable solution to the planning problem. Some works in the learning community have addressed this difficult and time-consuming task with data-driven techniques \cite{de2019learning}. However, the resulting set of motions still represents 
        % uniquely 
        a very limited range of the real dynamic capabilities of the robot. This is undesirable in applications where the environment is not known in advance, and where having the entire dynamic range of motions available for planning can be critical to finding a suitable solution. Moreover, accurate lattice-based methods struggle with high-dimensional state spaces due to the required grid-like discretization.
        
        To deal with kinodynamic constraints, 
        % without restricting the state space, 
        sampling-based motion planners offer great opportunities, e.g., \cite{kavraki1996probabilistic,hsu1997path,lavalle2001randomized}. Most sampling-based planners, however, lose their asymptotic optimality guarantees when a steering function does not exist in the system's kinodynamically constrained state space. To cope with this limitation, there are different assumptions and heuristics that can be applied at the expense of longer computational times. For example, \citeauthor{dustin2012} proposed a version of the \ac{RRT*} which can deal with kinodynamic constraints of systems with linearisable dynamics \cite{dustin2012}. If the system's dynamics are not linearisable, asymptotic optimality can be obtained in any planner by augmenting the dimensionality of the state space to account for the search cost \cite{hauser2016asymptotically}. However, this strategy implies solving the planning problem repeatedly to improve the cost of the solution at each iteration, consequently being unsuitable for applications with online requirements. Finally, the \ac{SST} planner offers asymptotically near-optimal guarantees by means of a shooting approach, which consists of expanding the tree from the node with the lowest cost within a neighbourhood of pre-defined \mbox{$\delta$-radius}~\cite{li2016asymptotically}.
        
        % Planning problems in high-dimensional spaces or with multiple constraints can endanger any planners' ability to find paths quickly. 
        Planning in high-dimensional spaces with multiple constraints poses a challenge for classical planners and typically result in long computation times if a solution can be found at all. In such problems, 
        a common approach to boost performance is via a multi-layered planning scheme. The key idea is to leverage from a lead to guide (warm-start) the search. In this regard, an interesting approach is the \ac{ITOMP} algorithm, which interleaves planning and optimisation; the planner is given a fixed time budget to find a solution, which is then used as a warm-start for the optimiser \cite{park2012itomp}. Work \cite{plaku2010motion,plaku2015region} introduced a synergistic three-layered planner: the high-level planner uses discrete search to initially determine those candidate regions (from a decomposed representation of the environment), which might contain part of the final solution; a low-level planner employs a sampling-based motion planner to find a solution; a middle layer updates the candidate regions according to the considered constraints. However, the proposed combination of planners does not guarantee asymptotic optimality, and the discrete planner becomes slow for high dimensional problems. \citeauthor{palmieri2016rrt} presented the \mbox{Theta*-\ac{RRT}} scheme, which first uses the Theta* path planner to compute a lead path, which is then employed to bias the search of the \ac{RRT} planner \cite{palmieri2016rrt}. This approach, however, lacks asymptotic optimality guarantees given that the second planner is an \ac{RRT}. More recently, a multi-layered approach based on the \ac{RRT*} as a lead planner and the \ac{SST} as the final planner has been proposed in \cite{vidal2019online}. The final planner's search space is strictly constrained around the lead path, raising concerns about the completeness guarantees of the overall architecture. 
 
    % ===============================
    % ===============================
    % ===============================
    %\subsection{Planning with Probabilistic Guarantees}
    \subsection{Planning under Uncertainty}
        % Survey of Motion Planning Literature in the Presence of Uncertainty: Considerations for UAV Guidance
        % https://www.academia.edu/26257664/Survey_of_Motion_Planning_Literature_in_the_Presence_of_Uncertainty_Considerations_for_UAV_Guidance?email_work_card=title

        An essential capability for any autonomous robot is to operate in the presence of uncertainty \cite{dadkhah2012survey}. Sources of uncertainty relevant to autonomous systems fall into four types \cite{lavalle1995framework}:
        \begin{itemize}
            % \item \textbf{Uncertainty in the pose information:}
            \item \textbf{Uncertainty in localisation:} 
            the robot's location is uncertain with respect to the environment. This issue is particularly critical in robots operating in GPS-denied environments, or for systems suffering from low-accuracy state estimation.
            % \item \textbf{Uncertainty in vehicle dynamics:} 
            \item \textbf{Uncertainty in motion (dynamics):} 
            the future robot state cannot be predicted accurately, either because of discrepancies between the considered and the real system's dynamic behaviour, or due to limited precision in the system's command tracking.% performance.
            \item \textbf{Uncertainty in the environmental awareness:} the robot has inexact or incomplete information about its surroundings (e.g. obstacle location). This issue can arise from inaccuracies in the a priori map, or imperfect and noisy exteroceptive sensory capabilities.
            \item \textbf{Disturbances in the operational environment:} the robot is subject to external factors, such as wind, atmospheric turbulences or water currents, which make the robot deviate from the planned trajectory, thus compromising the reliability of deterministic path planning techniques. %\ml{how's this different from motion uncertainty?} \erpaar{motion uncertainty comes from inaccurate "self-knowledge", while environment uncertainties come from unknown external factors. I agree that at the end everything can be model as motion uncertainty, but that was the classification in the paper mentioned above}\ml{well... motion uncertainty can be also be unknown and caused by unknown external factors.  But, since you're following LaValle's paepr, who am I to disagree! ;)}
        \end{itemize}
        This section scrutinises relevant planning strategies dealing with any of the three first sources of uncertainty.

        One approach that is popular among existing planners is based on discrete Markov processes. This strategy models the evolution of the system in the environment and generates a policy over the approximated Markov states. Examples of such motion planners include \ac{SMR}~\cite{alterovitz2007stochastic} and \ac{iMDP}~\cite{huynh2012incremental}. These methods have shown to be effective and provide optimality guarantees in terms of probability of reaching a desired goal; however, they assume perfect knowledge about the environment. Works such as~\cite{Luna:AAAI:2014} have extended these techniques to partially unknown environments. Nonetheless, their large computational times remains the main hurdle in applications with fully-unknown environments or requiring online planning.
        
        Another approach to deal with uncertainties in planning is by means of feedback controllers and sampling-based planners. \citeauthor{van2011lqg} proposed the \ac{LQG} motion planning method, which finds the best path simulating the performance of \ac{LQG} on all extensions of an \ac{RRT} \cite{van2011lqg}. This idea was later applied in roadmaps to propose the \ac{FIRM} \cite{agha2014firm}. This method, though, relies on full a priori awareness of the environment to explore the belief space offline, to then quickly perform queries online. Consequently, this strategy is not suitable for planning applications where the a priori information about the environment, if available, is not fully informative. Alternatively, \citeauthor{sun2015high} presented the \ac{HFR} architecture, a strategy which leverages from an \ac{LQG} and a multi-thread \ac{RRT}, allowing to continuously replan in the face of alterations in the robot or environment space, while accounting for uncertainties. However, the asymptotic optimality guarantees of such a method can only be assured if a multi-threaded implementation is realised.

        % Whereas most of the previous works aim at maximising the probability of success, some applications seek to find a motion path that satisfies a given (minimum) safety probability bound. In this regard, an efficient strategy is the chance-constraint method. 
        An alternative approach to dealing with uncertainty is the chance-constraint strategy.
        % have been developed to deal with uncertainty.  
        In these methods, 
        instead of maximising the probability of success, the objective is to find a path that satisfies a minimum safety probability constraint.   
        The challenge in incorporating this method in planners lies in the computation of the safety probability over plans. In~\cite{blackmore2011chance}, linear chance constraints are combined with disjunctive linear programming to perform probabilistic convex obstacle avoidance. This concept was extended and integrated into a sampling-based planner, leading to the \ac{CC-RRT}~\cite{luders2010chance} and the \mbox{CC-RRT*}~\cite{luders2013robust}. These approaches evolve the system's dynamics in an open-loop fashion, hence growing the uncertainty unboundedly forward in time. To improve accuracy, linear chance-constraints was applied after propagation of the system's state conditioned on the precedent states being collision-free \cite{patil2012estimating}. Such a strategy is commonly referred to as truncating the distribution estimating the system's state, and its usage in planning led to the \mbox{CC-RRT*-D} planner \cite{liu2014incremental}. The advantage of chance-constraint-based methods is that satisfying plans can be computed quickly, making them desirable for online applications. They are, however, built on strong assumptions which result in overly conservative calculations, and rely on the prior knowledge of a convex environment. Nonetheless, chance-constraint methods are still one of the most widely used strategies in the planning community to deal with localisation, motion and environmental uncertainties, e.g.,~\cite{strawser2018approximate,da2019collision}. 
        
        % Discrete supports are preferable when dealing with limited computational power or when meeting online requirements is a must. 
        In recent years, planners based on various discretization methods have been developed to deal with limited computational power or online planning requirements in face of uncertainty. 
        \citeauthor{majumdar2017funnel} proposed a precomputed library of funnels to efficiently estimate the system's kinodynamic and uncertainty propagation in \ac{3D} environments \cite{majumdar2017funnel}. However, library-based approaches consider a reduced set of the real system's capabilities which can endanger the efficacy of the planner. Another approach in favour of performance consists in approximating the computation of the probability of collision to a discrete support~\cite{strawser2018approximate,pairet2018uncertainty}. This strategy truncates the infinite expand of the belief in a bounded patch considered to contain a large portion of the belief's probability mass. In our previous work \cite{pairet2018uncertainty}, all uncertainties were projected onto a discrete support, referred to as kernel, whose resolution resembled the optimal one for online mapping applications. Although considering a discrete support for the computation of the probability of collision allows for quick calculations, none of the works using such technique actually normalises the calculations for the probability mass laying outside the patch, i.e., tail events. Therefore, they cannot offer guarantees on the compliance of the probabilistic safety constraints. 
        
    % ===============================
    % ===============================
    % ===============================
    \subsection{Frameworks for Online Mapping-Planning}
        Limited attention has been devoted to the online mapping and planning problem as a whole, especially in the face of uncertainties. Current frameworks in the robotics literature are built on strong assumptions which could endanger (or completely neglect) some of the essential requirements for safe navigation in undiscovered environments. Some of the prerequisites are the ability to create an uncertainty-aware representation of the environment, such that uncertainties about the environment can be considered at the planning stage. It is also crucial to ensure completeness guarantees, i.e. the ability of finding a solution if one exists, and among many others, being capable of guaranteeing the vehicle's safety at any time during the mission. Ideally, an online mapping-planning framework should be able to find paths quickly while offering asymptotic optimality guarantees.
        
        The common strategy for online mapping and planning is to continuously replan in the face of changes in the robot's pose or the environment awareness. \citeauthor{scherer2008flying} endowed an \ac{UAV} with the capability to map online with an occupancy probabilistic grid, to then guide itself towards the goal with a combination of global and local potential field-based planners \cite{scherer2008flying}. Along this line, navigation in 3D environments by mapping from stereo vision and planning with the \ac{RRT} was considered in \cite{andert2011mapping}. The resulting paths of these approaches do not account for kinodynamic constraints nor safety guarantees. Alternatively, in \cite{lin2014path}, the local planner of the \ac{RRT} approximated an \ac{UAV} capabilities by 3D Dubins paths. Nevertheless, none of these approaches considers any of the multiple sources of uncertainty in the mapping nor the planning stage, thus not providing any theoretical performance or safety guarantees.
        
        More recently, \cite{ho2018virtual} proposed an online framework to build an uncertainty-aware map and plan over it using the \ac{RRT}. However, the resulting paths do not meet kinodynamic nor safety constraints. Instead, proposals in \cite{hernandez2016planning, hernandez2019planning} presented an online framework to plan paths under motion constraints for \acp{AUV}, but their approach assumes zero uncertainty. Whilst their framework succeeded in solving start-to-goal queries in unexplored real-world environments, their planner used ad-hoc heuristics to estimate the risk associated with the solution path, and approximated the system's dynamics with Dubins curves. A similar framework is employed in \cite{vidal2019online}, where the \ac{SST} planner is employed to propagate an approximated dynamical model of the system. To counteract the computational expenses, the \ac{SST} is provided with a subregion of the state space drawn from a lead path. Finally, \cite{youakim2020multirepresentation} presented a multirepresentation, multi-heuristic A* planner capable of jointly dealing with the requirements of mobile-base and manipulation planning in unknown environments while accounting for localisation uncertainty via heuristics. Despite all methods have been tested in real-world environments, the underlying frameworks lack of theoretical analysis and do not provide a measure of robustness or quantified safety guarantees.

    % ===============================
    % ===============================
    % ===============================
    \subsection{Closely Related Contributions}
        An early version of the work presented in this manuscript has appeared before \cite{pairet2018uncertainty}. This consisted of a simpler framework that proved to be suitable for real-world motion planning problems, but its applicability was strictly limited to underwater robots operating at constant depth, i.e., 2D workspaces. This motivated the development of this follow-up work to extend the framework's capabilities to suit the requirements of a larger group of robotic systems and environments. Given the precedent efforts by the authors, this manuscript provides the following contributions:
        \begin{itemize}
            \item An online mapping-planning framework that guarantees the robot safety during navigation tasks in unknown environments (see \sref{sec:framework}). The framework now decouples the calculation of environmental uncertainties from the probabilistic collision checking routine, thus boosting the overall planning speed.
            \item A rigorous explanation of the mapping strategy using local submaps and the process of efficiently retrieving the environmental uncertainties in form of an uncertainty-aware map (\sref{sec:mapping}). These calculations now consider probabilistic map fusion to deal with the overlapping local submaps.
            \item An extension of the previous single-layered planner to a multi-layered scheme to guide the search, thus improving the planner's performance (\sref{sec:planning}). Kinodynamic constraints and probabilistic safety guarantees are still met. The probabilistic completeness and asymptotic optimality guarantees are also preserved. %\ml{completeness?} \erpaar{the planner in IROS was complete, wasn't it?} \ml{OK this one is very specific about the planner module, it's good. I missed it in the first round.}
            \item The formulation of a rapid probabilistic collision checking routine subject to a controllable confidence level~$\confidencelevel$ (\sref{sec:planning_pc}). 
            %The proposed routine can now adjust $\confidencelevel$ to trade accuracy 
            %\ml{tightness of the safety bound?} \erpaar{I agree that "accuracy" might not be the best description, but with "tightness of the safety bound" do you refer to psafe or confidence level? (see equation (21) and (22))} \ml{accuracy here kinda implies that the framework is giving up on safety for computational efficiency.  I would use a different word... like ``conservatism''?}
            %The proposed routine can now adjust $\confidencelevel$ to trade the tightness of the safety bound for computational efficiency, while correcting for the tail events (i.e. the probability mass excluded by the confidence level).
            Adjusting $\confidencelevel$ allows to trade the tightness of the safety bound for computational efficiency, while correcting for the tail events (i.e. the probability mass excluded by the confidence level).
            \item A thorough evaluation of the framework's key constituent components and the framework as a whole (see \sref{sec:evaluation}). This assessment considers extended analysis and additional experiments including simulations on different dynamical systems, and robot deployments on challenging real-world environments.
        \end{itemize}
    \section{Problem Formulation} \label{sec:formulation}

    In this work, the focus is on the challenging problem of safe autonomous navigation in unexplored environments for a mobile robot. To start with, the robotic system must be capable of perceiving and creating a consistent representation of the surroundings despite its potentially uncertain localisation. The perceived surroundings must be encoded efficiently such that the robot can exploit it online for planning purposes. Besides the mapping requirements, the process of planning navigation actions towards a desired goal is challenging by itself. The robot must not only account for its limited and uncertain manoeuvrability, but also for the evolving awareness and uncertainty of the surroundings as the robot moves.  This section provides formal definitions for these uncertainties and the problem of safe autonomous navigation in unexplored environments. The nomenclature used through this manuscript is summarised in \tref{tab:nomenclature}.

    \begin{table}[t]
        \centering
        %{\color{magenta}
        \begin{tabular}{ll}
            \toprule
            Symbol & Description \\
            \midrule
            \addlinespace[0.2cm]

            \multicolumn{2}{l}{\textbf{Generic definitions}} \\
            $\W$ & workspace \\
            $\X$ & state space \\
            $\B$ & belief space \\
            $\U$ & control space \\[0.2cm]
            
            $\belief_k = \N(\hat{\vx}_k, \; \cov_{\vx_k})$ & state belief at time k \\
            $\belief_A^B = \N(\hat{\vx}_A^B, \; \cov_A^B)$ & state belief of A as seen from B \\[0.2cm]
            
            \multicolumn{2}{l}{\textbf{Framework}} \\
            $\globalframe$ & global frame\\
            $\baseframe$ & robot base frame \\
            $\planningframe$ & planning frame \\[0.2cm]

            \multicolumn{2}{l}{\textbf{Mapping}} \\
            $\env$ & probabilistic map awareness \\
            $\locmap$ & local submap \\
            $F_O(\vx)$ & occupancy probability at $\vx$ \\
            $F_\X$ & cumulative map over $\X$ \\[0.2cm]

            \multicolumn{2}{l}{\textbf{Planning}} \\
            $b_\start$ & estimated system state at start \\
            $\X_\goal$ & goal region in state space \\
            $\Bgoal$ & goal region in belief space \\
            $p_\goal$ & minimum probability of goal \\
            $p_\safe$ & minimum probability of safety \\[0.2cm]
            
            $\planningtime$ & overall planning budget time \\
            $\planningtimescout$ & lead planner budget time \\
            $\planningtimetough$ & constrained planner budget time \\[0.2cm]
            
            $\traj^\prime$ & lead geometric path \\
            $\Xlead$ & lead region in state space \\
            %$\kernel_\alpha$ & $\alpha$-confidence uncertainty kernel \\
            $\traj$ & feasible and safe trajectory \\[0.2cm]

            \bottomrule
        \end{tabular}
        %}
        \caption{Summary of the nomenclature used in this document. %\erpaar{to make Table 1 extensive, it is missing: lmin, lmax, lfree, locc, DeltaT of local maps. Is it already too detailed?} \ml{no need to add them here.}
        }
        \label{tab:nomenclature}
    \end{table}

    \subsection{Motion Uncertainty and Constraints} \label{sec:formulation_preliminaries}
        %section III in https://www.ncbi.nlm.nih.gov/pmc/articles/PMC4535731/

        % Let the state ${\vx_\timestamp \in \X \subset \real^{n_x}}$ of a mobile robotic system which operates in a workspace ${\W \in \real^{n_w}}$ be probabilistically described at any time by a Gaussian distribution $\belief_\timestamp$:
        % \begin{align}
        % 	\vx_\timestamp \sim \belief_\timestamp = \N(\hat{\vx}_\timestamp, \; \cov_{\vx_\timestamp}),
        % \end{align}
        % where subscript $\timestamp$ represents time ${t = \timestamp \Delta t}$, $\belief_\timestamp$ is referred to as the \textit{belief} of $\vx_\timestamp$, and ${\hat{\vx}_\timestamp \in \real^{n_x}}$ and ${\cov_{\vx_\timestamp} \in \real^{n_x \times n_x}}$ are the mean and covariance of $\belief_\timestamp$. The set of all beliefs is called the belief space and denoted by $\B$, which is an uncertain representation of the state space $\X$.
        
        Consider a mobile robot that operates in a workspace ${\W \subset \real^{n_w}}$, where $n_w \in \{2,3\}$, under motion uncertainty.  
        The uncertainty in the robot's motion can be due to many reasons, e.g., unmodelled dynamics or noise in actuation, and can be described in several ways.  In this work, inspired by  \cite{nguyen2009model,le2009trajectory,hemakumara2013learning,beckers2017stable}, the evolution of the uncertain robotic system is assumed to be given by a Gaussian process.  That is, the robot state ${\vx \in \X \subseteq \real^{n_x}}$ at every time step $\timestamp$ is described by a Gaussian distribution, i.e.:
        \begin{align}
            \label{eq:belief_gp}
        	\vx_\timestamp \sim \belief_\timestamp = \N(\hat{\vx}_\timestamp, \; \cov_{\vx_\timestamp}),
        \end{align}
        where 
        $\belief_\timestamp$ is referred to as the \textit{belief} of $\vx_\timestamp$ and is fully defined by mean ${\hat{\vx}_\timestamp}$ and covariance ${\cov_{\vx_\timestamp}}$.  The set of all beliefs is called the belief space and denoted by $\B$.  Intuitively, $\B$ is an uncertain representation of the state space $\X$. 

        Mean $\hat{\vx} \in \X \subseteq \real^{n_x}$ is the nominal state of the robot and evolves according to:
        \begin{align}
            \label{eq:generic_km}
            \hat{\vx}_{k+1} = f(\hat{\vx}_k,\; \vu_k),
        \end{align}
        where ${\vu \in \U \subset \real^{n_u}}$ is the control input, and ${f: \X \times \U \to \X}$ captures the nominal (known) dynamics of the robot.
        Covariance ${\cov_\vx \in \real^{n_x\times n_x}_{>0}}$ describes the uncertainty around the nominal robot state and evolves according to:
        \begin{align}
            \label{eq:generic_cov}
            \cov_{\vx_{k+1}} &= g(\cov_{\vx_{k}},\; \vu_k),
        \end{align}
        where $g: \real^{n_x \times n_x} \times \U \to \real^{n_x \times n_x}_{>0}$ is the covariance function.  Examples of Gaussian process representations for robots with linear, unicycle, and fixed-wing dynamics are provided in \xref{sec:kinematic_models}.  Methods for modelling robots with (partially) unknown dynamics as Gaussian processes are discussed in \cite{nguyen2009model,jackson2020safety}.
        
    % ===============================
    % ===============================
    % ===============================
    \subsection{Environment Uncertainty} \label{sec:formulation_env}
        Some applications in robotics lack a complete awareness of the environment, either because there is no information of the surroundings or due to the presence of dynamic elements in the workspace. This work scopes the mapping requirements to undiscovered static environments. In order to reveal the obstacles in the environment, the robot is equipped with exteroceptive sensors, such that it can autonomously explore the surroundings as it moves, i.e., to integrate into the map the obstacles when they are inside the sensor's detection range. Importantly, most sensors uniquely detect points on the boundary of a nearby obstacle.

        This work assumes no uncertainty in the robot local observations denoted by $\reading_k$.  To transform this local observation from the robot frame to the global frame, let ${\reading_\timestamp \sim \N(\hat{\reading}_\timestamp, \; 0)}$.
        % Assuming no uncertainty on the robot local observations ${\reading_\timestamp \sim \N(\hat{\reading}_\timestamp, \; 0)}$, yet 
        Bearing in mind that the robot's location might be uncertain with respect to the global frame ${\belief_\timestamp \sim \N(\hat{\vx}_\timestamp, \; \cov_{\vx_\timestamp})}$, the observed point is represented in the global frame as:
        \begin{align}
            \belief_O &= \belief_\timestamp \oplus \reading_\timestamp, \\
            &= \N(\hat{\vx}_\timestamp \oplus \hat{\reading}_\timestamp, \; \jacobian_{1\oplus} \cov_{\vx_\timestamp} \jacobian_{1\oplus}^T),
        \end{align}
        where ${\belief_O \sim \N(\hat{\vx}_O, \; \cov_{\vx_O})}$ is the result of the Gaussian relationships via a compounding operator $\oplus$ explained in \xref{sec:operators}. From these uncertain points $\vx_O$, the robot constructs a probabilistic map $\env$. Then, the obstacle occupancy probability for point $\vx \in \X$ denoted by ${F_\X(\vx)}$ is the sum of the normally distributed densities in $\env$. The cumulative sum over all space $\X$ is called cumulative map and denoted by $F_\X$. 

    \subsection{Probabilistic Safety Guarantees} \label{sec:formulation_safety}
        The system's uncertainty and the environment's uncertainty are jointly considered to guarantee the vehicle's safety. More specifically, the probability of the system being in collision with an obstacle in the environment at time $k$ is characterised by:
        \begin{align}
            p_\col(b_\timestamp, \, \env) &= \int_{\X} b_\timestamp(\vx) \, F_\X(\vx) \, d\vx \nonumber \\
            &= \int_{\X} \N(\vx \, | \, \hat{\vx}_\timestamp, \, \cov_{\vx_\timestamp}) \, F_\X(\vx) \, d\vx,
            \label{eq:prob_collision}
        \end{align}
        where ${F_\X(\vx)}$ is the cumulative obstacle occupancy probability, as introduced in \sref{sec:formulation_env}. Then, given a minimum probability of safety $p_\safe$, we require ${1 - p_\col(b,~\env) \geq p_\safe}$ for every belief $b$ on the trajectory in order to probabilistically guarantee the robot's safety.

    % ===============================
    % ===============================
    % ===============================
    \subsection{Planning Problem} \label{sec:formulation_pp}
        Therefore, the planning problem considered in this work seeks a dynamically feasible trajectory in the belief space~$\B$ which is probabilistically safe. Formally, let ${\B_\goal \subset \B}$ denote the set of all belief states that correspond to the desired goal region $\X_\goal$ in the environment as:
        \begin{align}
            \Bgoal = \{b \in \B \; | \;  p_\region(b) \geq p_\goal\},
        \end{align}
        where,
        \begin{align}
            p_\region(b) = \int_{\X_\goal} \!\!\!\! b(\vx) \, d\vx,
        \end{align}
        and $p_\goal$ is the minimum probability that a belief must satisfy for being considered to be in the goal region. Then, the constrained planning problem is to compute a sequence of controls $\vu_0,\vu_1,\ldots,\vu_{T-1} \in \U$ that result in a dynamically-feasible trajectory ${\traj: [0,~T] \rightarrow \B}$ for the robotic system described by \eqref{eq:belief_gp}, \eqref{eq:generic_km}, and \eqref{eq:generic_cov} such that ${\traj(0) = b_\start \in \B}$, i.e. the system estimated state at the beginning of the mission, ${\traj(T) \in \Bgoal}$, and ${1 - p_\col(\traj(t),~\env) \geq p_\safe}$ for all ${t\in[0,~T]}$.
        
        %\ml{$\belregion_\goal$ and $b_\start$ are vague and not thoroughly undefined.  I know that I have a goal region in the work space.  How does that translate to $\belregion_\goal$.  We need to define this relationship.}
        
        %\ml{Are we planning for a path or a sequence of controllers?} % path -> trajectory

        \epa{}     

    \section{Framework for Online Mapping and Motion Planning} \label{sec:framework}

\begin{figure}[t]
        \centering
        \includegraphics[width=8cm]{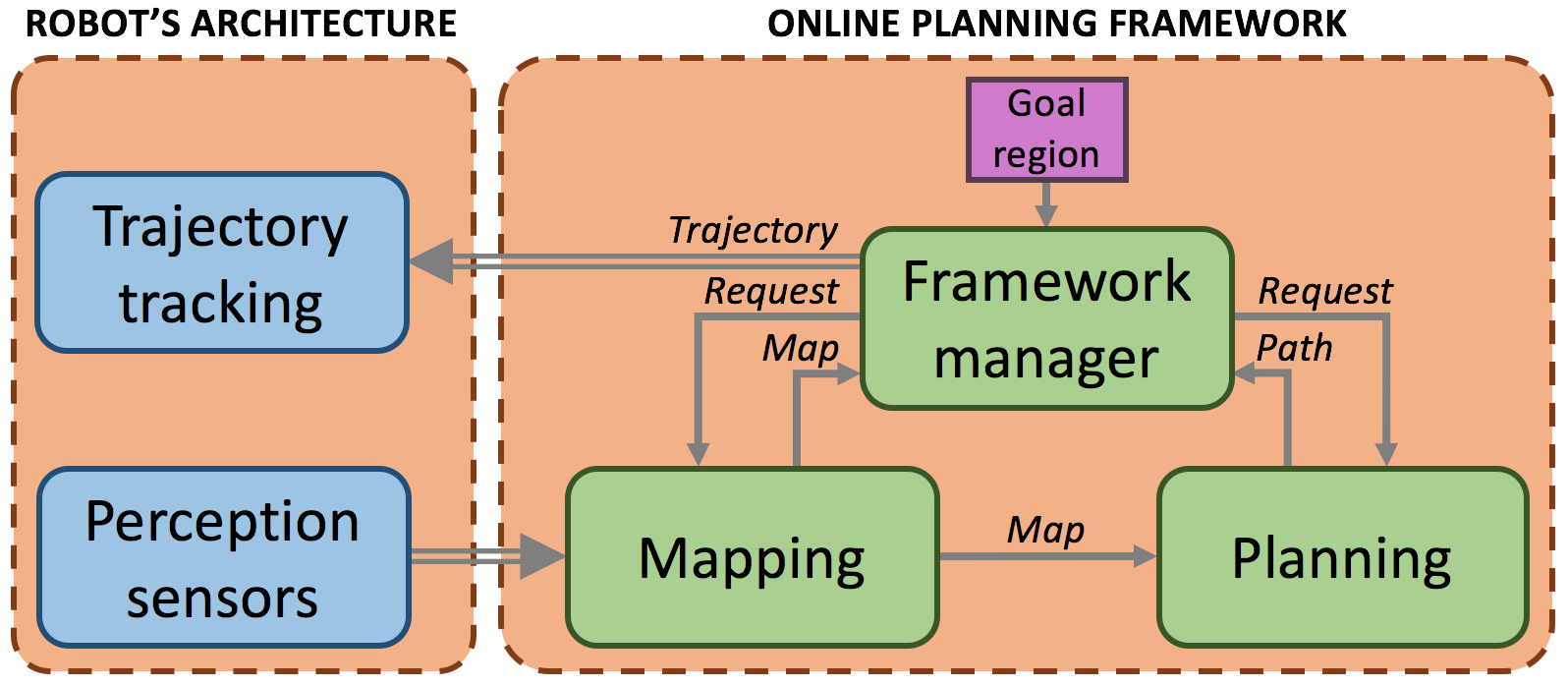}
        \caption{Framework for online mapping and motion planning under kinematic and uncertainty constraints.}
        \label{fig:framework_diagram}
    \end{figure}

    This manuscript presents a framework that endows a robotic system with the capability of safely navigating through unknown environments. This is achieved by means of online mapping and online motion planning of trajectories that meet motion and probabilistic constraints. The framework, depicted in \fref{fig:framework_diagram}, is threefold: (i)~a mapping module that incrementally builds an uncertainty-aware map, (ii)~a planning module that continuously computes a safe and feasible trajectory towards the goal, and (iii)~a framework manager that coordinates the overall framework's execution. The remainder of this section describes the manager's strategy to control the interaction between the two core modules of the framework, i.e. the mapping (see \sref{sec:mapping}) and the planning (see \sref{sec:planning}). Note that although the presentation of the framework focuses on the online mapping and planning challenge, the proposed online scheduling intrinsically solves the offline motion planning problem.

    % ===============================
    % ===============================
    % ===============================
    \subsection{Framework Pipeline}
        The framework manager coordinates the mapping (\textsc{mapper}) and planning (\textsc{planner}) modules according to the pipeline presented in Algorithm~\ref{alg:framework_manager}. This is, given the desired goal region $\Bgoal$ and the required probabilistic safety guarantees $p_\safe$, the manager conducts an iterative process until the system reaches the predefined goal region (line~\ref{alg_line:manager_stop}). An iteration consists of solving an updated version of the underlying motion planning problem which accounts for any alteration to the system's state and environment awareness.

        Each iteration starts with the manager predicting a suitable planning frame $\planningframe$ for the planning problem. The planning frame defines the state where the planner will start exploring the state space for a solution. As discussed in \sref{sec:framework_root}, the planning frame is determined according to the current plan in execution \textsl{ongoing\_traj} (line~\ref{alg_line:manager_root_1}). Then, the manager retrieves from the \textsc{mapper} the current environment awareness as a cumulative map $F_\X^{\planningframe}$ relative to $\planningframe$ (line~\ref{alg_line:manager_map}). Both the predicted planning frame $\planningframe$ and the updated cumulative map $F_\X^{\planningframe}$ are provided to the \textsc{planner} (line~\ref{alg_line:manager_update_1}~and~\ref{alg_line:manager_update_2}). %At this stage, the planning problem for this iteration is completely defined.
        
        Before proceeding to solve the updated planning problem, the current plan in execution \textsl{ongoing\_traj}, if any, is probabilistically checked for collision according to the current uncertainty-aware map $F_\X^{\planningframe}$. In the event of \textsl{ongoing\_traj} not being any longer valid, the framework manager dispatches to the robot the segment $\overline{\textsl{ongoing\_traj}}$ of \textsl{ongoing\_traj} that is still safe (line~\ref{alg_line:manager_old_path_1}~and~\ref{alg_line:manager_old_path_2}). 
        %\epa{The segment must satisfy a minimum length to avoid driving the vehicle towards a dead-end.}
        This approach prevents the vehicle from stopping every time that a trajectory gets partially invalidated while ensuring its safety.

        Finally, the \textsc{planner} attempts to solve the planning problem by growing a new tree in $\B$ for a specific amount of time $\planningtime$ (line~\ref{alg_line:manager_planner}). The \textsc{planner} tries to find a near-optimal trajectory which meets kinematic and probabilistic constraints within the allocated time budget $\planningtime$, and returns a \textsl{new\_traj} if one is found (line~\ref{alg_line:manager_solution}). 
        %\ml{what if it is not found?}. 
        The newly found \textsl{new\_traj} is uniquely dispatched to the robot when it fulfils the selection criteria defined in \texttt{satisfiesCriteria}() (line~\ref{alg_line:manager_path_update_1}~to~\ref{alg_line:manager_path_update_3}). This work bases the selection criteria \texttt{satisfiesCriteria}() on the length of the trajectory; \textsl{new\_traj} is dispatched if ${\texttt{length}(\textsl{new\_traj}) \leq \texttt{length}(\textsl{ongoing\_traj})}$, where ${\texttt{length}(\textsl{ongoing\_traj}) = \infty}$ if \textsl{ongoing\_traj} is partially invalidated, i.e. it does not reach the goal region $\Bgoal$.
        
        Note that the computations in line~\ref{alg_line:manager_root_1}~and~\ref{alg_line:manager_map} are low demanding and they can be scheduled in parallel to the main execution of the framework's pipeline. Therefore, the overall iteration rate of the framework is
        %approximately ${1/\planningtime}$. 
        ${1/\planningtime}$, as solving the planning problem (line~\ref{alg_line:manager_planner}) is the unique process of the framework that requires a non-negligible amount of time.
        %\ml{this is unclear.  What do you mean by "overall" iteration rate?  Each iteration of Alg 1 takes exactly $\planningtime$, doesn't it? Can the rate be larger than ${1/\planningtime}$?  Can the planner return a path in less than $\planningtime$?}
        
        Given the nature of the problem of navigation in unknown environments, it may be possible that a feasible and probabilistically safe trajectory towards the goal region does not exist.  Therefore,
        % As a feasible and probabilistically safe trajectory towards the goal region may not exist, 
        the framework is endowed with a contingency plan that attempts to return the vehicle nearby the deployment location $b_\start$. This contingency plan gets activated when the planner has not been able to find a solution in the last $n_{cp}$ consecutive iterations, where $n_{cp}$ is a user-defined safety value. In the event of the contingency plan getting activated, the \textsc{manager} is re-initialised with the new planning problem. Note that if the environment awareness is highly uncertain, there might not exist a trajectory towards the new goal region. In this situation, not considering the previous map information for planning would allow the vehicle moving safely towards the deployment location. In case no feasible motion plan is found to return to the deployment location, an emergency manoeuvre should be performed, e.g., coming to complete stop for ground vehicles, going to the water surface for AUVs, and immediate landing for UAVs.

        \begin{algorithm}[tb]
            \SetInd{0.7em}{0.7em}
        
            \caption{MANAGER($\Bgoal$, $p_\safe$) \label{alg:framework_manager}}
            \textbf{Input:} \\
            $\Bgoal$: Goal region \\
            $p_\safe$: Required probabilistic safety guarantees \\
            \Begin{
                $\textsl{ongoing\_traj} \gets \varnothing$ \\
                \textsc{planner}.\texttt{loadProblem}($\Bgoal$, $p_\safe$) \\
                \While{{\upshape \textbf{not} \texttt{isGoalAchieved}()} \label{alg_line:manager_stop}} {
                    \tcc{Predict planning frame}
                    $\planningframe \gets$ \texttt{pedictFrame}(\textsl{ongoing\_traj}) \\ \label{alg_line:manager_root_1}
                    \vspace{10pt}
                
                    \tcc{Retrieve cumulative map}
                    $F_\X^{\planningframe} \gets$ \textsc{mapper}.\texttt{getMap}($\planningframe$) \\ \label{alg_line:manager_map}
                    \vspace{10pt}
                    
                    \tcc{Update planning problem}
                    \textsc{planner}.\texttt{setNewFrame}($\planningframe$) \\ \label{alg_line:manager_update_1}
                    \textsc{planner}.\texttt{updateMap}($F_\X^{\planningframe}$) \\ \label{alg_line:manager_update_2}
                    \vspace{10pt}
                    
                    \tcc{Check ongoing plan}%validity of the path in execution}
                    \If{{\upshape \textbf{not} \textsc{planner}.\texttt{isValid}(\textsl{ongoing\_traj})} \label{alg_line:manager_old_path_1}} {
                        \texttt{dispatchPath}($\overline{\textsl{ongoing\_traj}}$) \label{alg_line:manager_old_path_2}
                    }
                    \vspace{10pt}

                    \tcc{Solve planning problem}
                    \textsc{planner}.\texttt{solve}($\planningtime$) \\ \label{alg_line:manager_planner}
                    \vspace{10pt}
                    
                    \tcc{Dispatch best valid plan}% to the robot}
                    \textsl{new\_traj} $\gets$ planner.\texttt{getSolution}() \\ \label{alg_line:manager_solution}
                    \If{\upshape \texttt{satisfiesCriteria}(\textsl{new\_traj}) \label{alg_line:manager_path_update_1}} {
                        \textsl{ongoing\_traj} $\gets$ \textsl{new\_traj} \\ \label{alg_line:manager_path_update_2}
                        \texttt{dispatchPath}(\textsl{ongoing\_traj}) \\ \label{alg_line:manager_path_update_3}
                    }
                }
            }
        \end{algorithm}
    
    % ===============================
    % ===============================
    % ===============================
    \subsection{Prediction of the Planning Frame \label{sec:framework_root}}
        % \ml{this sub-section looks out of place. We may not need such an elaborate discussion on Line 8.  It is pretty intuitive and can be summerized in 2-3 sentences.  Hence, can be inserted where line 8 is discussed above.} \erpaar{I agree that it is quite intuitive, particularly the calculations of predicting the frame. However, it seems that online planning is not that obvious to people used to manipulation/offline planning (have been asked about about the frame prediction via email (relating to our IROS 2018), and in the group meetings in Rice and Edinburgh.}
    
        Optimising motion planning strategies, such as the one presented in this work, employ a time budget $\Delta T$ to find a trajectory which solves a predefined planning problem. Consequently, there is, at least, a time lapse $\Delta T$ between the definition of the planning problem and the usage of the resulting trajectory. In applications where the robot state does not change during $\Delta T$, it is reasonable to define the planning frame $\planningframe \sim \vx_{\planningframe}^\globalframe$ as the current system's state ${\vx_{\baseframe}^\globalframe \sim \N(\hat{\vx}_{\baseframe}^\globalframe, \; \cov_{\baseframe}^\globalframe)}$, which is represented with respect to the global frame $\globalframe$.
        %\ml{the notations are confusing.  $B$ has been used in eqn (5) and also as regions in the belief space.  Now it is used as the predictFrame.  Also, what's $\vx_{\planningframe}^\globalframe$?  Is it a distribution?}
        However, in applications where the robot might be in movement in the course of $\Delta T$, such as the online mapping and planning application targeted in this work, a suitable planning frame must be specified ahead of time. Ideally, the frame of the planning problem should be defined such that it corresponds to the robot state at the time the resulting trajectory will be utilised.
        
        Bearing in mind that the presented framework computes a candidate new trajectory periodically every $\planningtime$ and that it has full knowledge of the plan in execution \textsl{ongoing\_traj}, calculating a planning frame at time $t$ (line~\ref{alg_line:manager_root_1}) which is suitable at time $t+\planningtime$
        %\ml{$k$ is defined as the time step in the Prob Formulation section.  Use a different notation, e.g., $t_k$}
        (line~\ref{alg_line:manager_path_update_3}) can be formulated as a state prediction problem. This is, given the current robot state $\vx_{\baseframe}^\globalframe$ and the set of subsequent controls $\vu$ involved in the execution of \textsl{ongoing\_traj}, $\vx_{\planningframe}^\globalframe$ is computed by integrating \mbox{\eref{eq:generic_km}-\eref{eq:generic_cov}} for the time-horizon $\planningtime$. 
        
        Predicting the frame of the planning problem ahead on time (i)~guarantees the feasibility of reaching the initial state of any found solution from the current robot state, (ii)~prevents sudden changes in the vehicle's direction of motion when transitioning from the \textsl{ongoing\_traj} to the \textsl{new\_traj} (line~\ref{alg_line:manager_path_update_1}~to~\ref{alg_line:manager_path_update_3}), and (iii)~enables the planner to periodically set a planning problem that considers any update on the robot's localisation estimate.
    \section{Incrementally Mapping Unknown Environments via Local Maps} \label{sec:mapping}
    % some relevant references:
    % http://robots.stanford.edu/papers/thrun.mapping-tr.pdf
    % https://ieeexplore.ieee.org/stamp/stamp.jsp?tp=&arnumber=8392399
    % https://ieeexplore.ieee.org/stamp/stamp.jsp?tp=&arnumber=7487233&tag=1
    % https://ieeexplore.ieee.org/stamp/stamp.jsp?tp=&arnumber=6766467
    % http://www.cs.cmu.edu/~kaess/pub/Ho18iros.pdf
    % http://citeseerx.ist.psu.edu/viewdoc/download?doi=10.1.1.373.1169&rep=rep1&type=pdf

    Incrementally exploring the environment with a location-uncertain system leads to an uncertain representation of the surroundings. Under these conditions, obtaining a consistent and reliable representation of the entire environment is a challenging task commonly addressed with probabilistic inference approaches. These algorithms rely on gathering data from which distinctive features (landmarks) can be extracted and used to bound the uncertainty of the environment representation and system localisation. Nonetheless, even for scenarios rich in features, there are always some residual uncertainties. Moreover, onboard perception sensors usually suffer from noises, which compromise the accuracy of the environment representation. All these issues motivate the need of an environment representation that jointly explains captures the uncertainty on the true obstacle's localisation and the detection confidence according to the sensor model, while being suitable for motion planning. In this work, such a representation is referred to as probabilistic map. 
    
    This section details the undertaken mapping approach, which builds a set of local occupancy submaps whose base poses are uncertain with respect to a global frame (see \sref{sec:mapping_global}). Each submap is an occupancy grid map, which provides an efficient strategy to encode the incremental environment awareness (see \sref{sec:mapping_local}) and retrieve information about the environment occupancy (see \sref{sec:mapping_single_query} and \sref{sec:mapping_all_query}). This overall mapping strategy has proven to be suitable for real-time robotic mapping and planning applications in our previous work~\cite{pairet2018uncertainty}, and despite being out of the scope of this manuscript, has also shown to be effective for online mapping and localisation applications~\cite{ho2018virtual}. 
    %\ml{this paragraph seems unnecessary! can keep the first sentence but the rest of the paragraph is just an outline of the section, which I don't think is necessary.}

    % ===============================
    % ===============================
    % ===============================
    \subsection{Global Map as a Set of Local Submaps \label{sec:mapping_global}}
        There are different alternatives to represent the incremental knowledge of an environment. The framework presented in this manuscript encodes the environment $\env$ via a set of $n$ local stochastic submaps~\cite{pinies2007scalable,pinies2008large} due to its demonstrated efficiency on dealing with applications requiring real-time robotic localisation, mapping and planning~\cite{pairet2018uncertainty,ho2018virtual}. Formally, the local submaps method is defined as:
        %\ml{the word strategy means a very different thing to me.  Is it necessary to keep?}
        \begin{gather}
            \env = \{ \locmap_1, \; \dots, \; \locmap_n \}, \\
            \locmap_i = \left\{ \{\voxel_1, \; \dots, \; \voxel_m\}, \; \hat{\vx}_{\locmap_i}^{\globalframe}, \; \cov_{\locmap_i}^{\globalframe} \right\},
        \end{gather}
        where each local submap $\locmap_i$ contains a set of sequential sensor scans over a finite horizon time $\Delta T_{\locmap}$. Within this time period, all point coordinates $\voxel$ of the sensed environment are registered into the active submap $\locmap_n$. The coordinate frame of $\locmap_n$ is defined in a global frame $\globalframe$ by its estimated state ${\vx_{\locmap_n}^{\globalframe} \sim \N\left(\hat{\vx}_{\locmap_n}^{\globalframe}, \cov_{\locmap_n}^{\globalframe}\right)}$. 
        % \ml{Again, the notation is a bit confusing.  The meaning of the subscripts keep changing.  We've used the subscripts for time step $k$, for belief $B$ or prediction, and now local maps!} \erpaar{still after adding Table 1?}
        Importantly, such local registration assumes null uncertainty on observations, i.e. ${\cov_\voxel^{\locmap_n} = \mathbf{0} \; \forall \; \voxel \in \locmap_n}$.
        
        A new local submap $\locmap_{n+1}$ is initiated every $\Delta T_{\locmap}$ such that the accumulated localisation error within the active local submap $\locmap_n$ is low. In other words, the local mapping time horizon $\Delta T_{\locmap}$ must be defined such that it always 
        %maintain the relative robot pose covariance $\cov_{\baseframe}^{\locmap_n}$ negligible.
        maintains the robot pose uncertainty $\cov_{\baseframe}^{\locmap_n}$ within the active local map $\locmap_n$ negligible.
        %\epa{\st{within admissible uncertainty levels.} below a pre-defined threshold such that the uncertatnites withinthe local map are negligible.}
        %\ml{I'm not sure what it means. By admissible uncertainty level, do you mean negligible?}
        
        The coordinate system of a new local submap $\locmap_{n+1}$ is defined at the robot state estimate when $\locmap_{n+1}$ is initiated, i.e. ${\vx_{\locmap_{n+1}}^{\globalframe} = \vx_{\baseframe}^{\globalframe}}$. 
        % \ml{notation!} \erpaar{still after adding Table 1?}
        It is assumed that the robot starts building $\locmap_{n+1}$ as soon as it finishes the $\locmap_{n}$. Therefore, the robot state at the end of $\locmap_{n}$ (defined as the last global robot state when building $\locmap_{n}$) is the same as the global robot start state of $\locmap_{n+1}$. For simplicity, the origin of the global map $\globalframe$ is chosen to be the same as the coordinate frame of the first local submap $\locmap_1$, i.e. the robot's initial state. 
        
        \begin{figure}[t]
            \centering
            \subfloat[Environment]{\includegraphics[width=1.0\columnwidth]{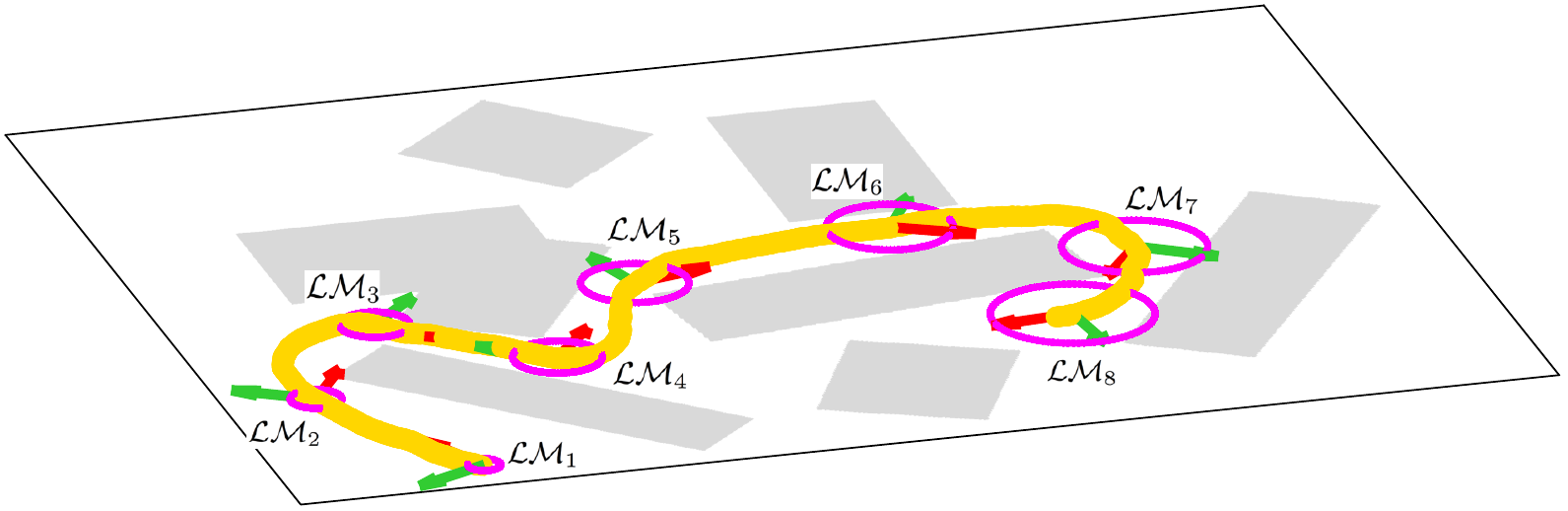}}
            \\
            \subfloat[Probabilistic map $\env$]{\includegraphics[width=1.0\columnwidth]{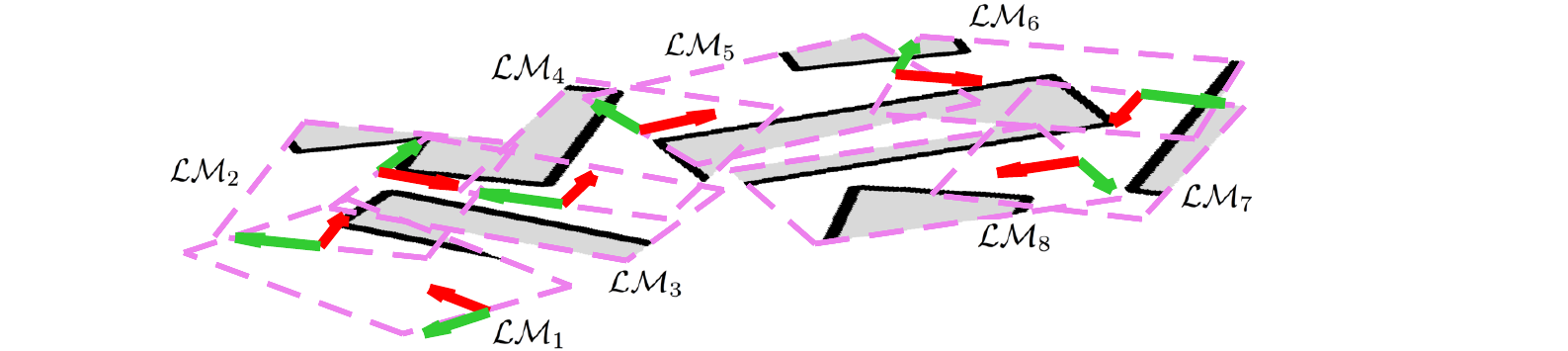}}
            \caption{As the robot navigates through an unknown environment (top image), it builds a probabilistic map $\env$ which represents the surroundings as a set of local maps (bottom image). Each local map is uncertain (magenta circle on top image) with respect to the global frame.}
            \label{fig:mapping_local_maps}
        \end{figure}
        
        \fref{fig:mapping_local_maps} illustrates the concept of using local submaps to map the incremental knowledge about the environment. Particularly, the figure depicts a robot which has been navigating in an unknown environment, while in the meantime, it has been encoding the perceived surrounding environment in a total of eight local submaps. Noteworthy, the example assumes an open-loop navigation, i.e., without localisation updates. Therefore, the first defined submaps are less uncertain with respect to the global frame $\globalframe$ than those built at a later stage. This fact corresponds to an unbounded growth of the uncertainty on the system localisation estimate.
    
    % ===============================
    % ===============================
    % ===============================
    \subsection{Local Submap as Occupancy Grid Map \label{sec:mapping_local}} % https://tel.archives-ouvertes.fr/tel-01680375/document
    
        The assumption of null uncertainty on the robot pose within each local submap, also referred to as known robot poses, enables the representation of each local submap as an occupancy grid map. The chosen alternative to efficiently encode an occupancy grip map is via Octomaps~\cite{hornung13auro}. Octomaps permits fusing range-based data into a probabilistic voxel representation, which generates an occupancy grid map with adjustable resolution. Octomaps store the information in an octree data structure, which provides fast access time while, at the same time, optimises the memory usage. All these desirable features make the undertaken mapping strategy ideal for online mapping and planning.

        The probabilistic sensor fusion within an occupancy grid map is performed as an Octomap~\cite{moravec1985high,hornung13auro}. This is, the probability $P(\voxel|\reading_{1:\timestamp})$ of a cell $\voxel$ to be occupied given a set of sensor measurements $\reading_{1:\timestamp}$ is estimated as:
        \begin{gather}
            P(\voxel|\reading_{1:\timestamp}) =
            \resizebox{0.78\hsize}{!}{$
            \left[1+\frac{1-P(\voxel|\reading_\timestamp)}{P(\voxel|\reading_\timestamp)}\frac{1-P(\voxel|\reading_{1:\timestamp-1})}{P(\voxel|\reading_{1:\timestamp-1})}\frac{1-P(\voxel)}{P(\voxel)}\right]^{-1}\!\!,$}
            \label{eq:voxel_update_p}
        \end{gather}
        where $P(\voxel|\reading_\timestamp)$ is the inverse sensor model characterising the sensor used for mapping 
        %\ml{I don't understand this.  Isn't it just a probability of $\voxel$ being occupied given the measurement at time $k$?},
        and $P(\voxel|\reading_{1:\timestamp-1})$ is the preceding estimate given all historical measurements. Using log-odds notation:
        \begin{align}
            L(\cdot) = \log\left[\frac{P(\cdot)}{1-P(\cdot)}\right],
        \end{align}
        %${L(\cdot) = \log\left[\frac{P(\cdot)}{1-P(\cdot)}\right]}$ \ml{this definition should have its own line} notation 
        and under the common assumption of a uniform (non-informative) prior, i.e., ${P(\voxel) = 0.5}$, \eref{eq:voxel_update_p} is simplified to:
        \begin{gather}
            L(\voxel|\reading_{1:\timestamp}) = L(\voxel|\reading_{1:\timestamp-1}) + L(\voxel|\reading_\timestamp).
            \label{eq:logodd_update}
        \end{gather}
        
        To change the state of a node $\voxel$, \eref{eq:logodd_update} requires as many observations as the ones used to define its current state. This overconfidence in the map is addressed as in~\cite{yguel2008update} by using a clamping policy to ensure that the confidence in the map remains bounded: 
        \begin{align}
            L(\voxel|\reading_{1:\timestamp})
            & = \left[L(\voxel|\reading_{1:\timestamp})\right]_{l_{min}}^{l_{max}} \nonumber \\
            & = \max\left(\min\left(L(\voxel|\reading_{1:\timestamp}), l_{max}\right), l_{min}\right),
            \label{eq:logodd_clamping}
        \end{align}
        where $l_{min}$ and $l_{max}$ denote lower and upper bound on log-odds values. As a consequence, the model of the environment remains updatable~\cite{hornung13auro}.
        
        \begin{figure}[b]
            \centering
            \subfloat[]{\includegraphics[width=0.48\columnwidth]{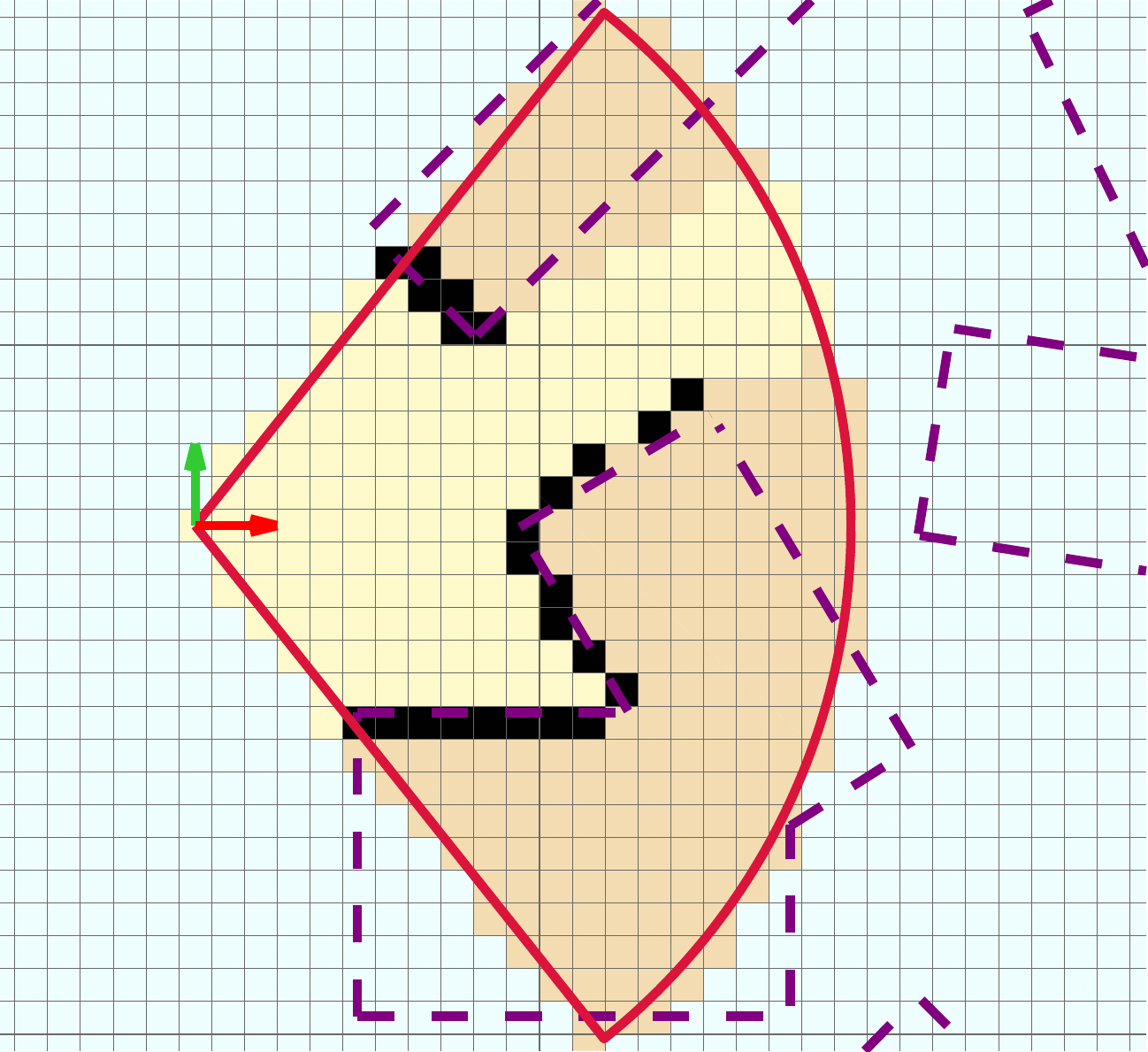}}
            \;
            \subfloat[]{\includegraphics[width=0.48\columnwidth]{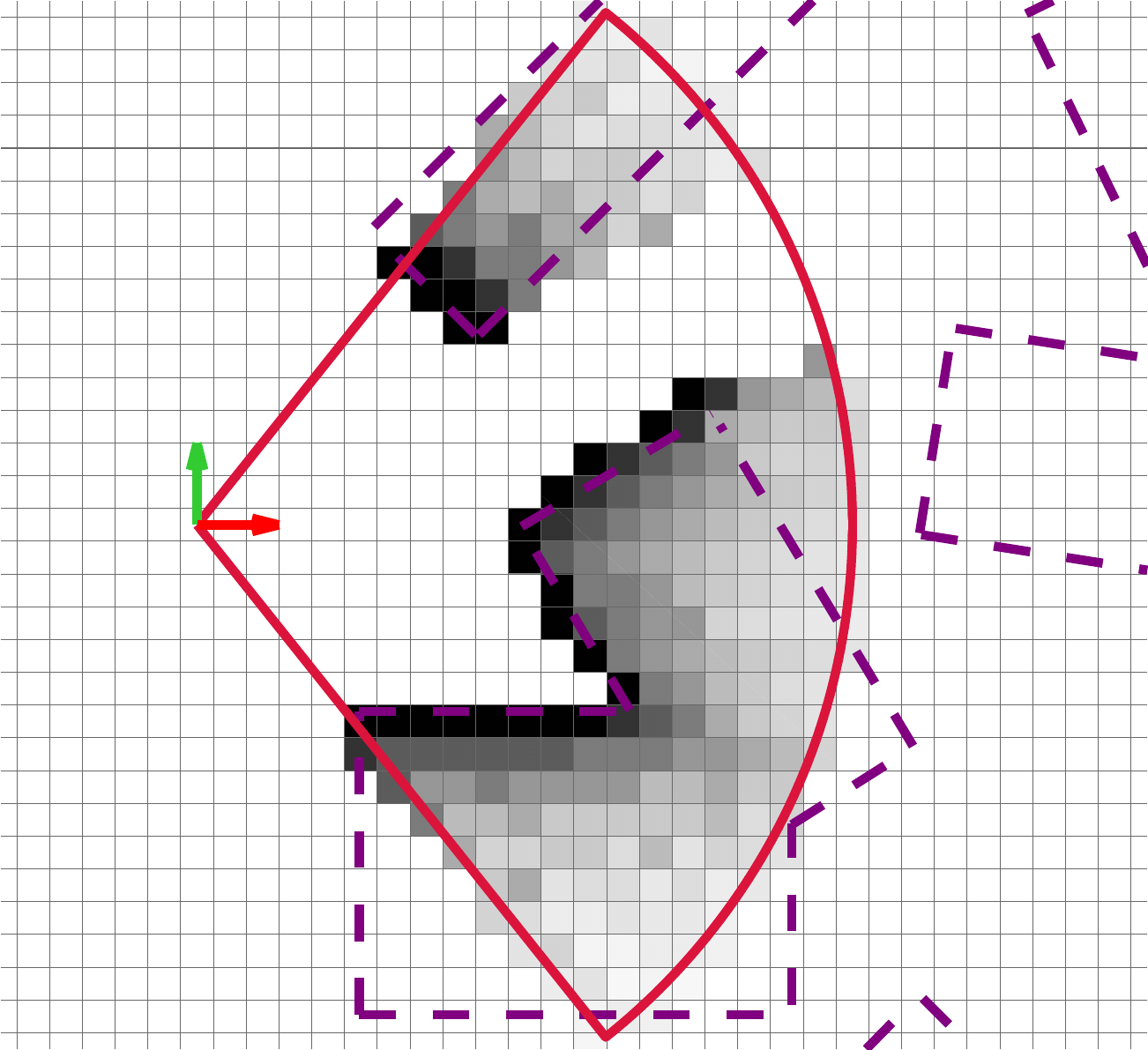}}
            \\
            \includegraphics[width=\columnwidth]{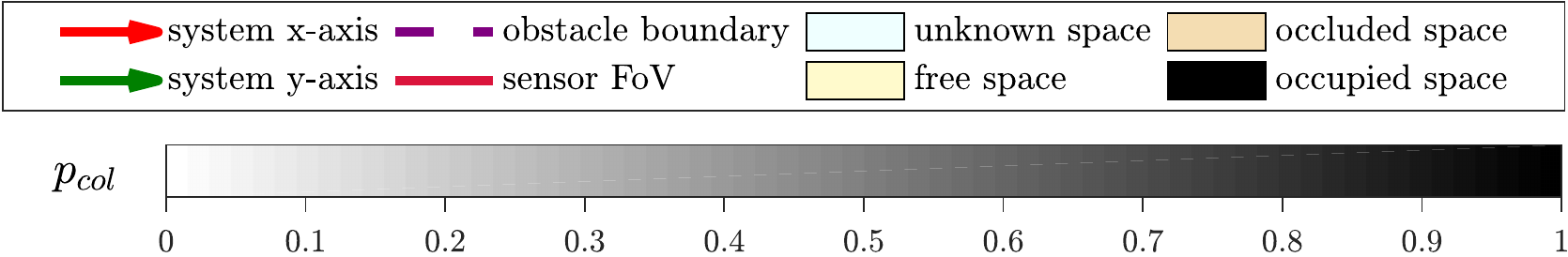}
            \caption{Environment understanding according the considered inverse sensor model when scanning from a locally known state. }
            \label{fig:inverse_sensor_model}
        \end{figure}
        
        The measurement update rules in~\eref{eq:logodd_update}-\eref{eq:logodd_clamping} can be used with any kind of distance sensor, as long as the inverse sensor model is available. Our framework employs the extended beam-based inverse sensor model depicted in \fref{fig:inverse_sensor_model}. This model assumes: (i)~that the line of sight between the sensor origin and the endpoint of a measurement does not contain any obstacle (free space), (ii)~that endpoints correspond to obstacle surfaces (occupied space), and (iii)~that the line continuing beyond the endpoint until the maximum sensor range is likely to be occupied by the observed obstacle (occluded space). Then, the extended ray-casting operation to update each voxel $\voxel$ from the sensor origin to the maximum sensor range is performed using the following log-odds inverse sensor model:
        \begin{gather}
            L(\voxel|\reading_t) =
            \begin{cases}
                l_{free} & \!\!\text{if $\voxel$ is traversed by the beam}, \\
                l_{occ} & \!\!\text{if $\voxel$ is hit by the beam}, \\
                l_{ocl} & \!\!\text{if $\voxel$ is between the hit and sensor range},
            \end{cases}
        \end{gather}
        where $l_{free}$ and $l_{occ}$ are constants determined according to the sensor model, and $l_{ocl}$ penalises occluded zones according to the decaying function:
        \begin{gather}
            l_{ocl} = \gamma^d \; l_{occ},
            \label{eq:decaying_occlusion}
        \end{gather}
        where for a decay rate $\gamma \in (0, 1)$, $l_{ocl}$ decreases $\gamma$ times for each unit of $d$, which is the distance from the measurement endpoint. This corresponds to ${l_{ocl} = l_{occ}}$ for $d=0$, i.e., in the hit point, and to ${l_{ocl} \rightarrow 0}$, i.e., to a non-informative ${P(\voxel) = 0.5}$, as ${d \rightarrow \infty}$. The maximum expand of the occluded region is as far as the sensor range.
        
        %In our experimentation, we use the default parameters in~\cite{hornung13auro} of ${l_{min}=-2}$ and ${l_{max}=3.5}$ corresponding to the occupancy probabilities ${P(\voxel) = 0.12}$ and ${P(\voxel) = 0.97}$, respectively, and ${l_{free} = -0.4}$ and ${l_{occ} = 0.85}$ which corresponds to ${P(\voxel) = 0.4}$ and ${P(\voxel) = 0.7}$, respectively. The decay rate is set at $\gamma=0.8$.
        %\ml{should this paragraph be here or in the experiments section?}

    % ===============================
    % ===============================
    % ===============================
    \subsection{Map Fusion and Single Point Query \label{sec:mapping_single_query}}
        An occupancy query to the current probabilistic map $\env$ is done by converting the given query into multiple local queries. The occupancy probability values at each local submap can be fused together by means of the log-odds update rule in~\eref{eq:logodd_update} with the corresponding clamping operation in~\eref{eq:logodd_clamping}. These operations apply because combining measurements from multiple local submaps is a similar operation as combining multiple measurement updates in a single global map~\cite{ho2018virtual}.

        Without loss of generality, assume that an occupancy query at position $\hat{\vx}^\randomframe$ 
        %\ml{is it the expected value of the position?} \erpaar{it is to highlight that the query is not a distribution but a configuration.} \ml{in that case just use $x$ rather than $\vx$.  To me $\vx$ is a process and evolves with time.  $x$ is just a point in $\X$.} 
        is performed from an uncertain coordinate frame $\randomframe$ with known pose estimate ${\vx_{\randomframe}^{\globalframe} \sim \N\left(\hat{\vx}_{\randomframe}^{\globalframe}, \cov_{\randomframe}^{\globalframe}\right)}$. 
        % \ml{why is $\randomframe$ a subscript here?  Notations need to be consistent.} \erpaar{state belief of Y as seen with respect to the global frame W}
        This global query corresponds to the multiple local log-odds occupancy queries:
        \begin{gather}
            L(\hat{\vx}^\randomframe) = \sum_{i = 1}^{n}
            \left[
            L_{1:i-1}(\hat{\vx}^\randomframe) + L_i(\vx^{\locmap_i})
            \right]_{l_{min}}^{l_{max}},
            \label{eq:pocc_single_map}
        \end{gather}
        where $L_{1:i-1}(\hat{\vx}^\randomframe)$ is the accumulative log-odd estimate from the precedent ${i-1}$ local submaps with ${L_{1:i-1}(\hat{\vx}^\randomframe) = 0}$ for ${i = 1}$, $L_i(\cdot)$ implies that the log-odds lookup is done in the local submap $\locmap_i$, and ${\vx^{\locmap_i} \sim \N(\hat{\vx}^{\locmap_i}, \cov_{\randomframe}^{\locmap_i})}$ corresponds to $\hat{\vx}^{\randomframe}$ in local coordinates. $\vx^{\locmap_i}$ is calculated via the linear estimation of known spatial relationships:
        \begin{gather}
            \vx^{\locmap_i} = \ominus \vx_{\locmap_i}^{\globalframe} \oplus 
            \left(
            \vx_{\randomframe}^{\globalframe} \oplus \vx^{\randomframe}
            \right),
        \end{gather}
        where $\oplus$ denotes the compounding operation and $\ominus$ corresponds to its inverse relation, as commonly used to simplify notation when calculating spatial transformations (see \xref{sec:operators} for a brief introduction and \cite{smith1990estimating} for a full review). 
        
        Given that $\hat{\vx}^{\randomframe}$ in local coordinates follows a probabilistic distribution, the local occupancy query $L_i(\vx^{\locmap_i})$ is:
        \begin{gather}
            P_i(\vx^{\locmap_i}) = \!\!\! \sum_{\voxel \in \locmap_i} \!\!\! P(\voxel) \; \N(\voxel \; | \; \hat{\vx}^{\locmap_i}, \cov_\randomframe^{\locmap_i}),
            \label{eq:pocc_single_locmap}
        \end{gather}
        where $\voxel$ represents the set of voxels in submap $\locmap_i$ and $P_i(\vx^{\locmap_i})$ can be described in log-odds $L_i(\vx^{\locmap_i})$ notation via the log-odds transform.

    % ===============================
    % ===============================
    % ===============================
    \subsection{Computation of the Cumulative Map $F_\X^{\planningframe}$ \label{sec:mapping_all_query}}
        The previous section provides a strategy to query the occupancy probability $P(\vx)$ of a single point coordinate~${\vx \in \X}$. Our previous work~\cite{pairet2018uncertainty} 
        %proved \ml{did we prove it or did we show it?} \erpaar{not theoretically, but we demonstrated that it is a feasible approach for online planning via experiments} 
        demonstrated that this approach is suitable for the requirements of an online planner under probabilistic constraints. However, bearing in mind that each planning cycle requires numerous queries of $P(\vx)$ involving different $\vx$, the overall planner performance can be enhanced by computing the map fusion before the planning time budget starts.
        
\begin{figure}[t]
            \centering
            \subfloat[Probabilistic map $\env$]{\includegraphics[width=1.0\columnwidth]{Figures/mapping_local_maps.pdf}} \\
            \subfloat[Cumulative map $F_\X^{\planningframe}$]{\includegraphics[width=1.0\columnwidth]{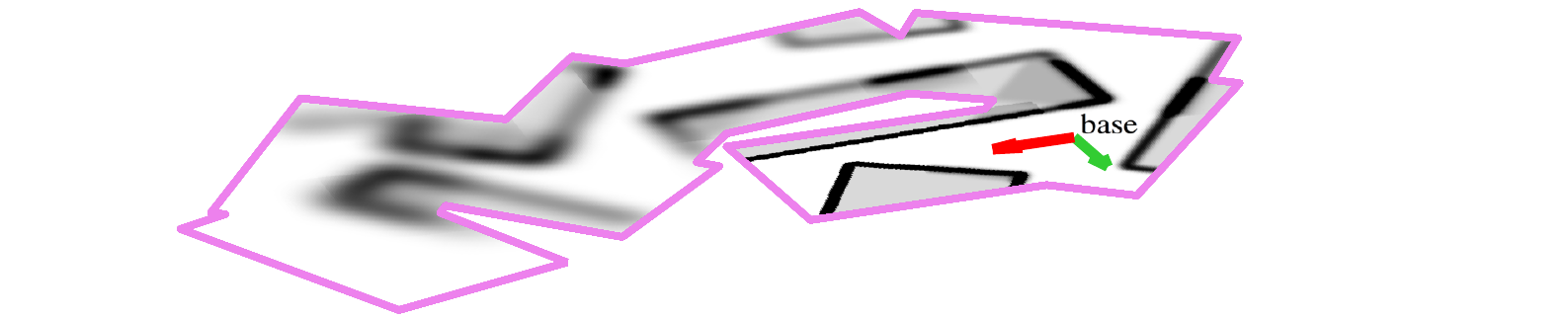}}
            \caption{The set of local maps (encoded as occupancy grid maps) is fused into a cumulative map representation $F_\X^{\planningframe}$ taking into account the relative uncertainty of the current probabilistic map with respect to the predicted planning frame $\planningframe$.}
            \label{fig:mapping_query_all}
        \end{figure}

        The probabilistic map fusion over all state space $\X$ is described by the cumulative distribution $F_\X$ over the local density distributions of the sensed environment\footnote{Only those voxels describing the known environment, i.e. free, occluded and occupied space, are considered in the computation of~$F_\X$. Considering the unknown space with its ${P(\voxel) = 0.5}$ in the computations would lead to a cumulative map with misleadingly over-estimating occupancy probabilities.}. In particular for the online planning problem, it is of interest to fuse the map information with respect to the predicted planning frame $\planningframe$ (see \sref{sec:framework_root}), such that the cumulative map $F_\X^{\planningframe}$ reflects the relative uncertainty between the current environment awareness and the planning frame $\planningframe$. \fref{fig:mapping_query_all} illustrates the extraction of $F_\X^{\planningframe}$ from a set of local maps. Computing $F_\X^{\planningframe}$ implies that the computational requirements of retrieving $P(\vx^{\planningframe})$ during planning time are reduced to those of a look-up table in the cumulative map $F_\X^{\planningframe}$.
        
        Subject to the log-odds transformation, $F_\X^{\planningframe}$ is computed by rewriting~\eref{eq:pocc_single_map}-\eref{eq:pocc_single_locmap} as:
        \begin{gather}
            L(\hat{\X}^{\planningframe}) = \sum_{i = 1}^{n}
            \left[
            L_{1:i-1}(\hat{\X}^{\planningframe}) + L_i(\X^{\locmap_i})\right]_{l_{min}}^{l_{max}},
            \label{eq:pocc_multiple_map}
        \end{gather}
        where $L_{1:i-1}(\hat{\X}^{\planningframe})$ is the accumulative log-odd estimate from the precedent ${i-1}$ local submaps with ${L_{1:i-1}(\hat{\X}^{\planningframe}) = 0}$ for $i = 1$, $L_i(\cdot)$ implies that the log-odds lookup is done in the local occupancy submap $\locmap_i$, and ${\X^{\locmap_i} \sim \N(\hat{\X}^{\locmap_i}, \cov_{\planningframe}^{\locmap_i})}$ corresponds to the state space $\hat{\X}^{\planningframe}$ in local coordinates defined as:
        \begin{gather}
            \X^{\locmap_i} = \ominus \X_{\locmap_i}^{\globalframe} \oplus 
            \left(
            \X_{\planningframe}^{\globalframe} \oplus \X^{\planningframe}
            \right).
        \end{gather}

        Then, the occupancy probability $L_i(\X^{\locmap_i})$ at $\locmap_i$ for all $\vx \in X^{\locmap_i}$ is computed as:
        \begin{align}
            P_i(\X^{\locmap_i}) &= \!\!\! \sum_{\voxel \in \locmap_i} \!\!\! P(\voxel) \; \N(\voxel \; | \; \hat{\vx}, \cov_{\planningframe}^{\locmap_i}) \nonumber \; \forall \; \vx \in \X^{\locmap_i} \nonumber\\
            &= \locmap_i \otimes \kernel_\confidencelevel\left(\cov_{\planningframe}^{\locmap_i}\right)
            \label{eq:pocc_local_map}
        \end{align}
        where $\voxel$ represents the set of voxels in submap $\locmap_i$, $\kernel_\confidencelevel(\cdot)$ with confidence level ${\confidencelevel=1}$ is a kernel representing the discrete version of a Gaussian distribution over the entire span of the local submap $\locmap_i$ (see \xref{sec:kernel_construction}). $\otimes$ is the correlation operator, i.e. a sliding inner product, and $P_i(\X^{\locmap_i})$ can be described in log-odds $L_i(\X^{\locmap_i})$ via the log-odds transform.
        
        Interestingly, the underlying computation of $F_\X^{\planningframe}$ is the correlation operator~$\otimes$, a common technique for which there exist efficient implementations. On top of that, the independence between local submaps allows parallelising the computation of \eref{eq:pocc_local_map} for each $\locmap_i$ in different threads. Ideally, this process could be scheduled such that $F_\X^{\planningframe}$ is ready before the planning time budget starts.
    \section{Multi-layered Motion Planning under Environment and Motion Uncertainty} \label{sec:planning}

    \begin{figure*}[t]
        \centering
        \subfloat[Multi-layered planning scheme]{\includegraphics[width=6cm, clip]{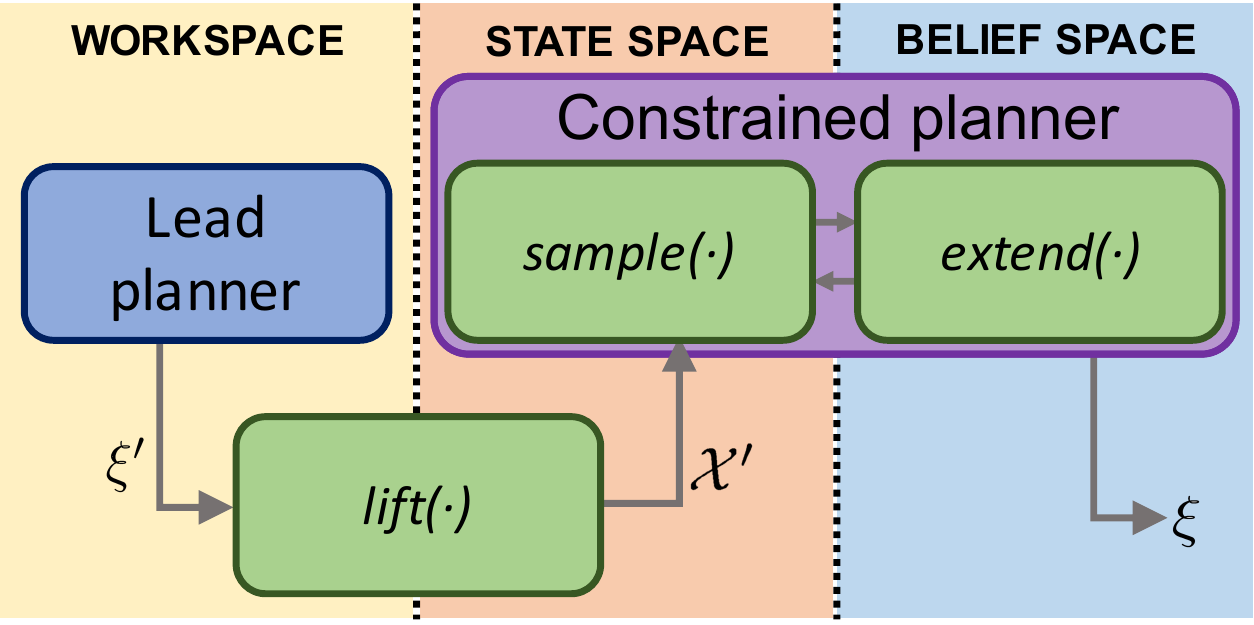}\label{fig:mlp_bigpicture_scheme}}
        \,
        \subfloat[Lead planner - \ac{RRT*}]{\includegraphics[width=0.65\columnwidth, trim=6cm 0cm 0cm 0cm, clip]{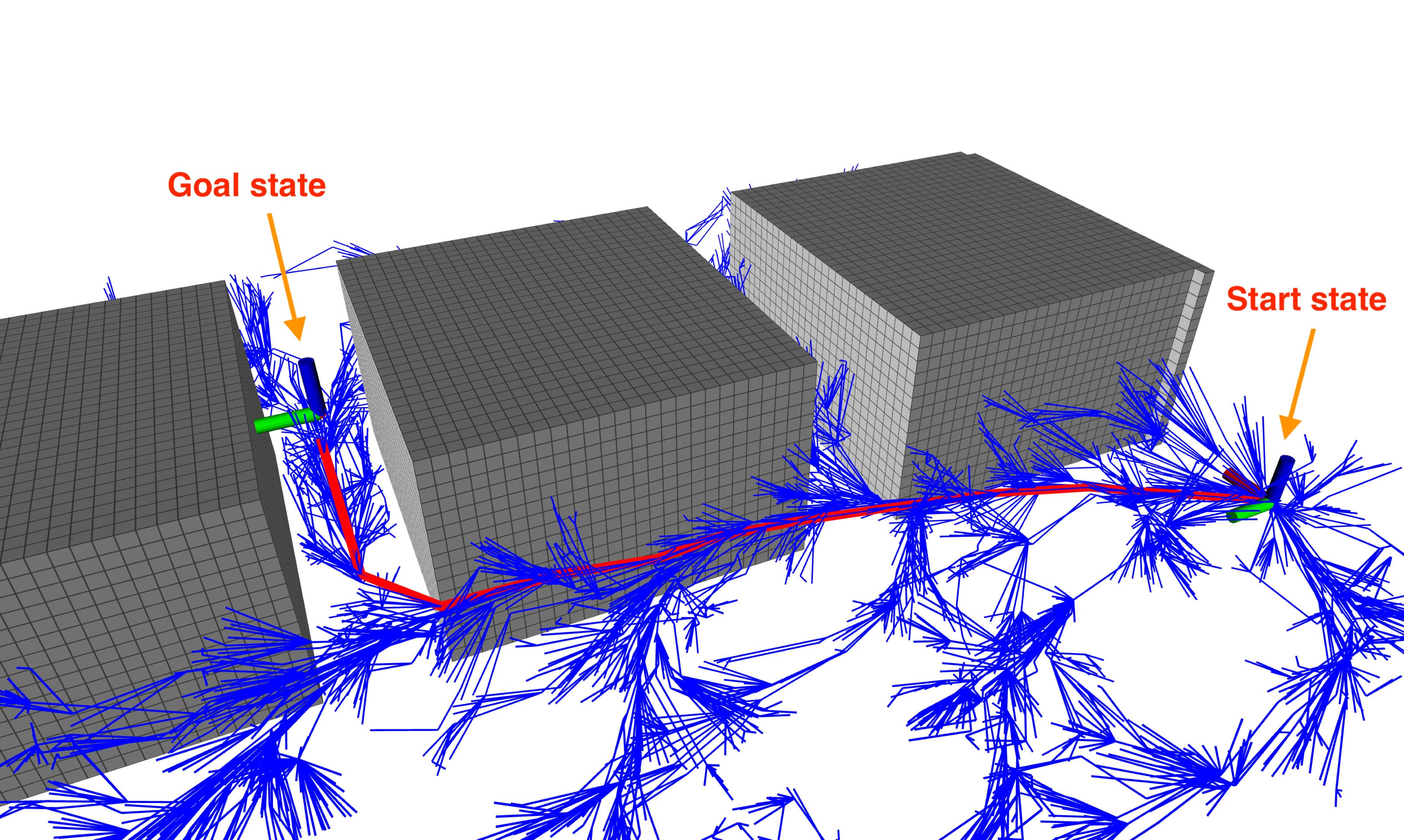}}
        \,
        \subfloat[Constrained planner - \ac{SST}]{\includegraphics[width=0.65\columnwidth, trim=6cm 0cm 0cm 0cm, clip]{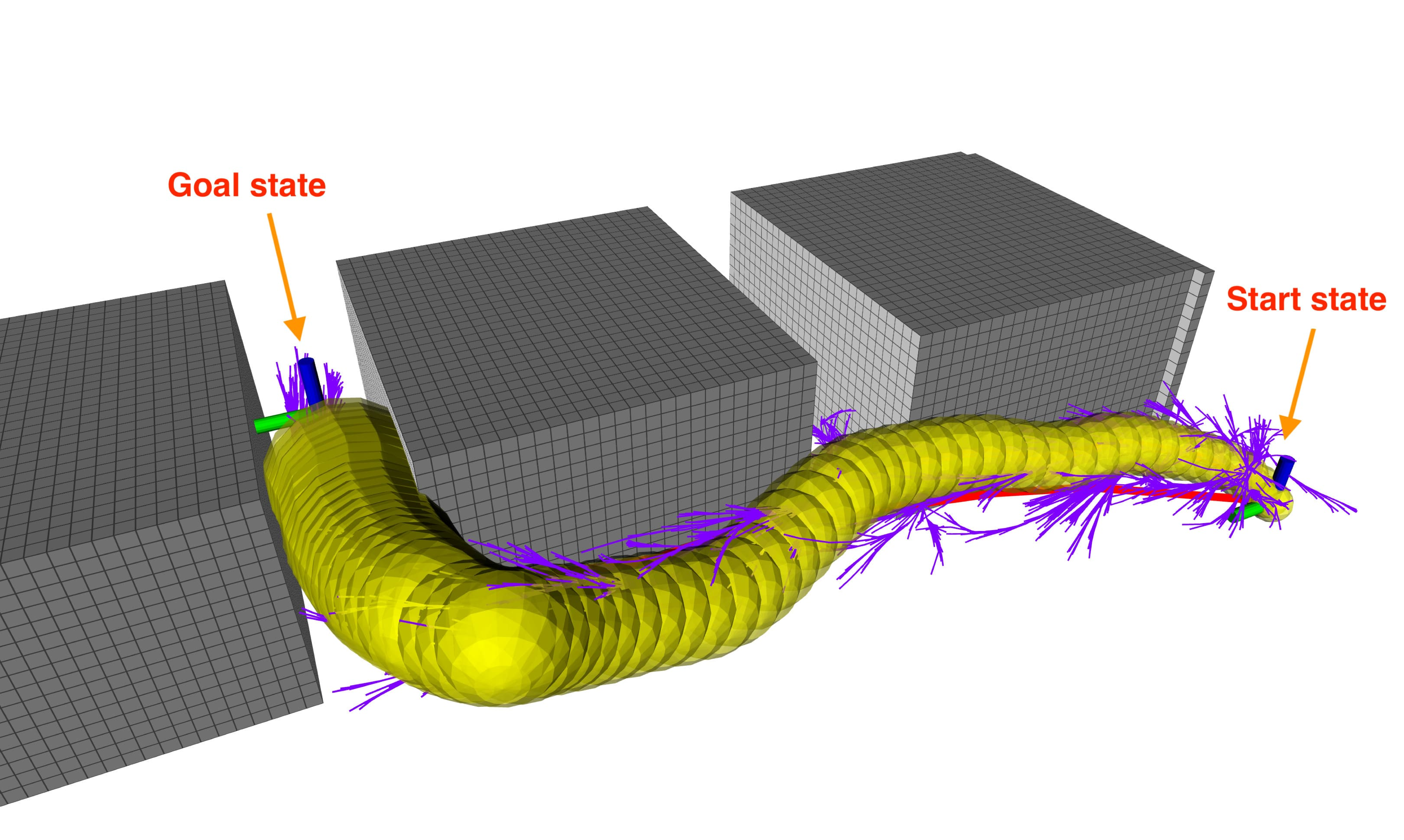}}
        \caption{
        Multi-layered motion planning framework: the lead planner shown in blue in (b) computes a geometric lead $\traj^\prime$ (red path) to guide the search space $\Xlead$ of the constrained planner shown in magenta in (c). The resulting trajectory $\traj$ with its uncertainty (yellow funnel) satisfies kinodynamic and probabilistic safety constraints.}
        \label{fig:mlp_bigpicture}
    \end{figure*}

    The planning problem defined in Section~\ref{sec:formulation_pp} has three main requirements: (a)~to consider the vehicle's motion constraints, (b)~to validate probabilistic constraints in face of uncertainties, and (c)~to meet online computation limitations. Our previous approach successfully addressed all these requirements formulating a single-layered sampling-based planning strategy in the belief space \cite{pairet2018uncertainty}. The planner in question (i)~samples feasible states in the system's state space, and (ii)~extends and validates the tree of motions in the belief space. This approach proved to be suitable for solving online motion planning problems in challenging real-world scenarios, but its applicability was limited to low-dimensional planning problems given the huge search space and the computational burden of all considered constraints.

    As discussed in \sref{sec:background}, multi-layered planning strategies enable online planning in high-dimensional spaces. This motivates the use of such an idea to extend our framework’s capabilities to suit the planning requirements of a larger group of robotic systems and environments. Principally, the extended planning strategy employs a multi-layered planning scheme (see \sref{sec:planning_multi}) to overcome the aforementioned scalability issues. Such a strategy allows deferring the computation of kinematic constraints (see \sref{sec:planning_kc}) and probabilistic constraints (see \sref{sec:planning_pc}) after identifying some subregions of the system's state space that potentially contain a solution to the planning problem.
    
    % ===============================
    % ===============================
    % ===============================
    \subsection{Multi-layered Motion Planning \label{sec:planning_multi}}
        The capabilities of our previous planner (hereinafter referred to as the constrained planner) are extended to deal with problems of higher dimensionality by means of a multi-layered planning strategy. As schematised in \fref{fig:mlp_bigpicture}, the proposed strategy adopts a sequential two layered planning scheme consisting of a \textbf{lead planner} and the \textbf{constrained planner}. The lead planner seeks to determine a subregion ${\Xlead \subset \X}$ of the entire state space that eases, and consequently speeds up, the search of the final trajectory $\traj$ which accounts for all considered constraints (see \sref{sec:formulation_pp}). To this aim, the multi-layered scheme is designed as following:
        \begin{itemize}
            \item \textbf{Lead planner:} employs the \ac{RRT*} algorithm to rapidly find a path in the workspace $\W$ while optimising a desired cost. The obtained lead path is a nearly optimal geometric solution ${\traj^\prime \in \W}$ used to determine $\Xlead$ via the lifting operator ${lift: \W \rightarrow \X}$ detailed below.
            \item \textbf{Constrained planner:} leverages the delimited search space $\Xlead$ and the \ac{SST} algorithm in \cite{pairet2018uncertainty} to rapidly find the final solution $\traj$ which meets kinodynamic constraints (see \sref{sec:planning_kc}) and probabilistic safety constraints (see \sref{sec:planning_pc}).
        \end{itemize}
        
        Although the planners within the multi-layered planning scheme could be different, the selection above suits the online requirements of our framework. This is, the framework's overall planning time $\planningtime$ is divided as ${\planningtime = \planningtimescout + \planningtimetough}$, where $\planningtimescout$ and $\planningtimetough$ are the time budgets allocated to the lead and constrained planners, respectively. Then, given our selection of planners, the assignment of time budgets allows ${\planningtimescout \ll \planningtimetough}$ as (i)~the lead planner is adept at providing quickly a suitable lead path, such that (ii)~the constrained planner has at its disposal most of the time budget ${\planningtimescout \approx \planningtime}$ to refine the final trajectory which accounts for all the considered constraints.
        
        Given the aforementioned selection of planners and their operational space, the designed multi-layered planning scheme requires the lifting ${lift: \W \rightarrow \X}$. A common $lift(\cdot)$ strategy is to define $\Xlead$ as a tube around $\traj^\prime$ with radius $d$ for the geometric components of the state space, whereas the non-geometric components are left unbounded~\cite{vidal2019online}. The performance of this approach, however, is susceptible to the parametrisation of $d$; tight search spaces, i.e. small radius $d$, promote final solutions with lower cost than those obtained with bigger radius $d$. On top of that, relying on a fixed $d$ requires hand-tuning such parameter to ensure that the final solution lies within $\Xlead$; if $\Xlead$ does not contain the final solution, the planner will lack probabilistic completeness. Adjusting $d$ to ensure probabilistic completeness would prove to be a cumbersome task since the type of environment and planning constraints, among many other factors, should be taken into account.

        Differently from other multi-layered planning schemes in the literature, ours uses a method of information interchange between planners that maintains the completeness and asymptotic optimality properties of the constrained planner when used in a standalone fashion~\cite{pairet2018uncertainty}. This work builds on the idea of sampling around a lead path to present alternative definitions of $\Xlead$ via the $lift(\cdot)$ operator.
        %, which maintain the probabilistic completeness of the constrained planner in our previous work \cite{pairet2018uncertainty}. 
        In particular, the designed multi-layered planner exploits a mixture of samplers to trade-off the low-cost trajectories found when sampling around a lead path and the probabilistic completeness of uniform sampling. This manuscript proposes two mixture of sampling techniques:
        \begin{itemize}
            \item \textbf{Bias to rigid $\Xlead$:} given a fixed radius $d$, the planner samples uniformly in $\Xlead$  with probability $p$ and uniformly over the space with probability ${1-p}$.
            \item \textbf{Adaptive $\Xlead$:} the planner adjusts $d$ within the range of a strictly guided sampling to a uniform search. Adjusting $d$ can be conducted via some heuristics or as an optimisation problem subject to a cost function.
        \end{itemize}
        
\begin{figure}[t]
            \centering
            \subfloat[Lead planner - \ac{RRT*}]{\includegraphics[width=0.49\columnwidth, trim=6cm 0cm 7cm 0cm, clip]{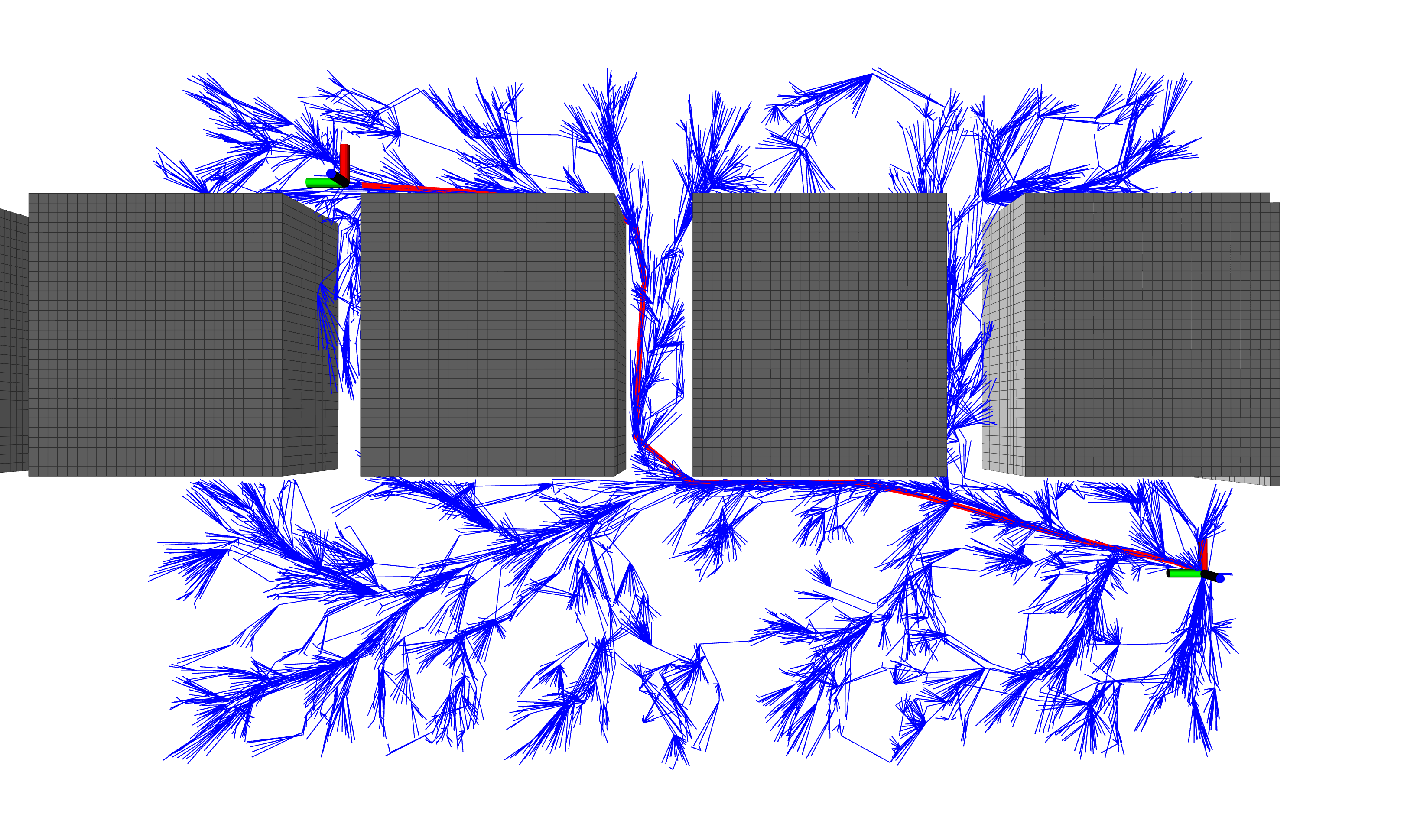}\label{fig:mlp_dyn_rad_1s}}
            \,
            \subfloat[Constrained planner]{\includegraphics[width=0.49\columnwidth, trim=6cm 0cm 7cm 0cm, clip]{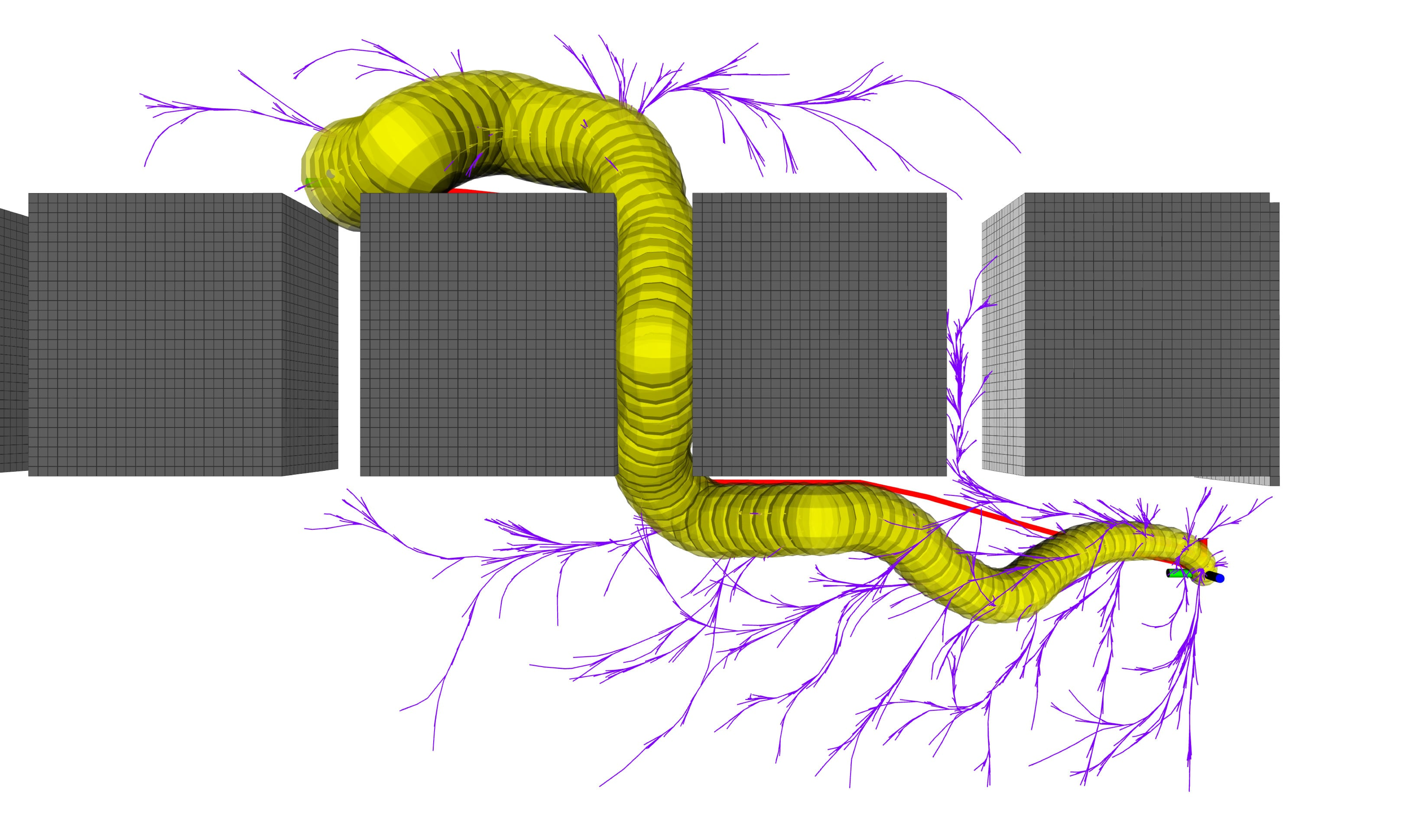}\label{fig:mlp_dyn_rad_1t}}
            \\
            \subfloat[Lead planner - \ac{RRT*}]{\includegraphics[width=0.49\columnwidth, trim=6cm 0cm 7cm 0cm, clip]{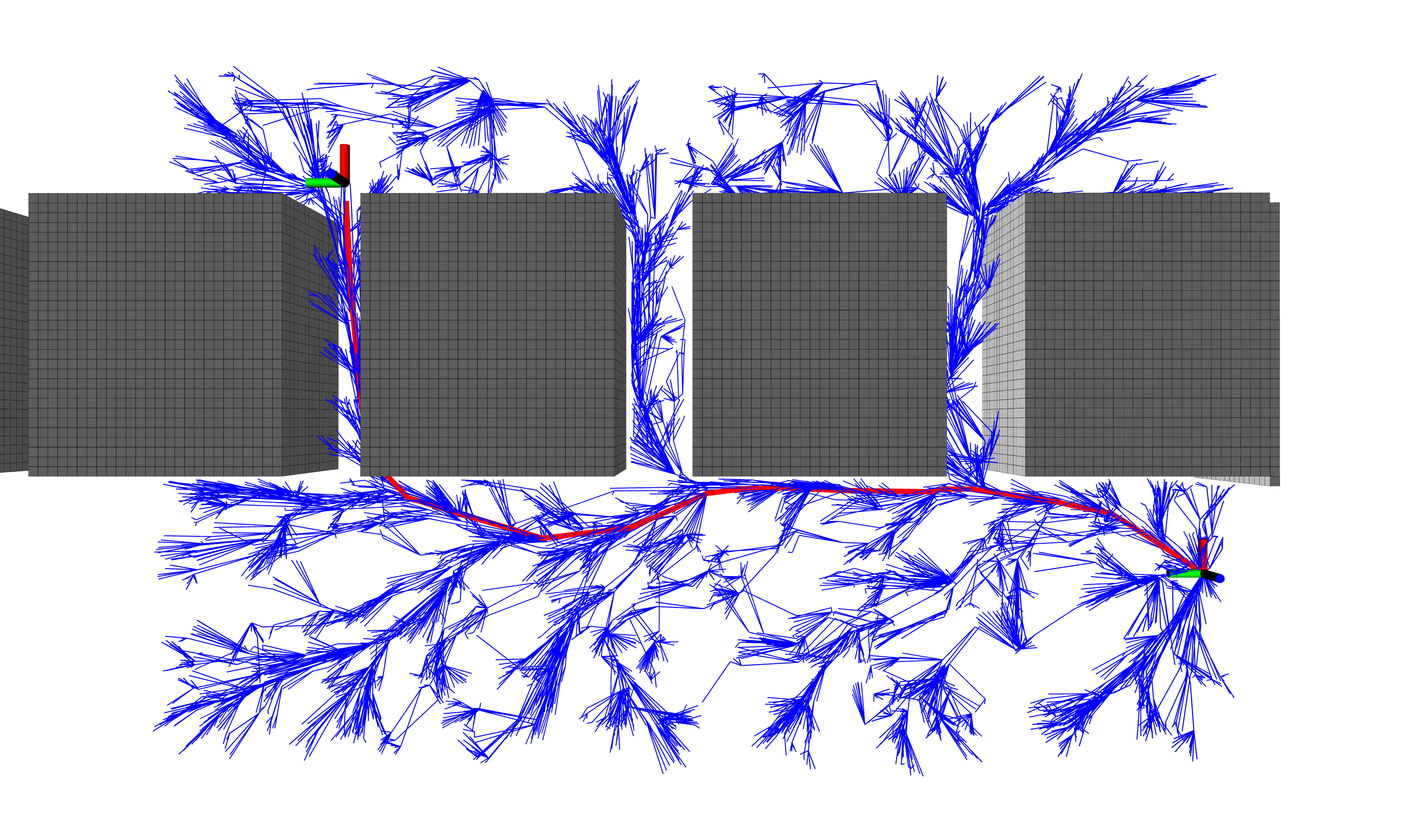}\label{fig:mlp_dyn_rad_2s}}
            \,
            \subfloat[Constrained planner]{\includegraphics[width=0.49\columnwidth, trim=6cm 0cm 7cm 0cm, clip]{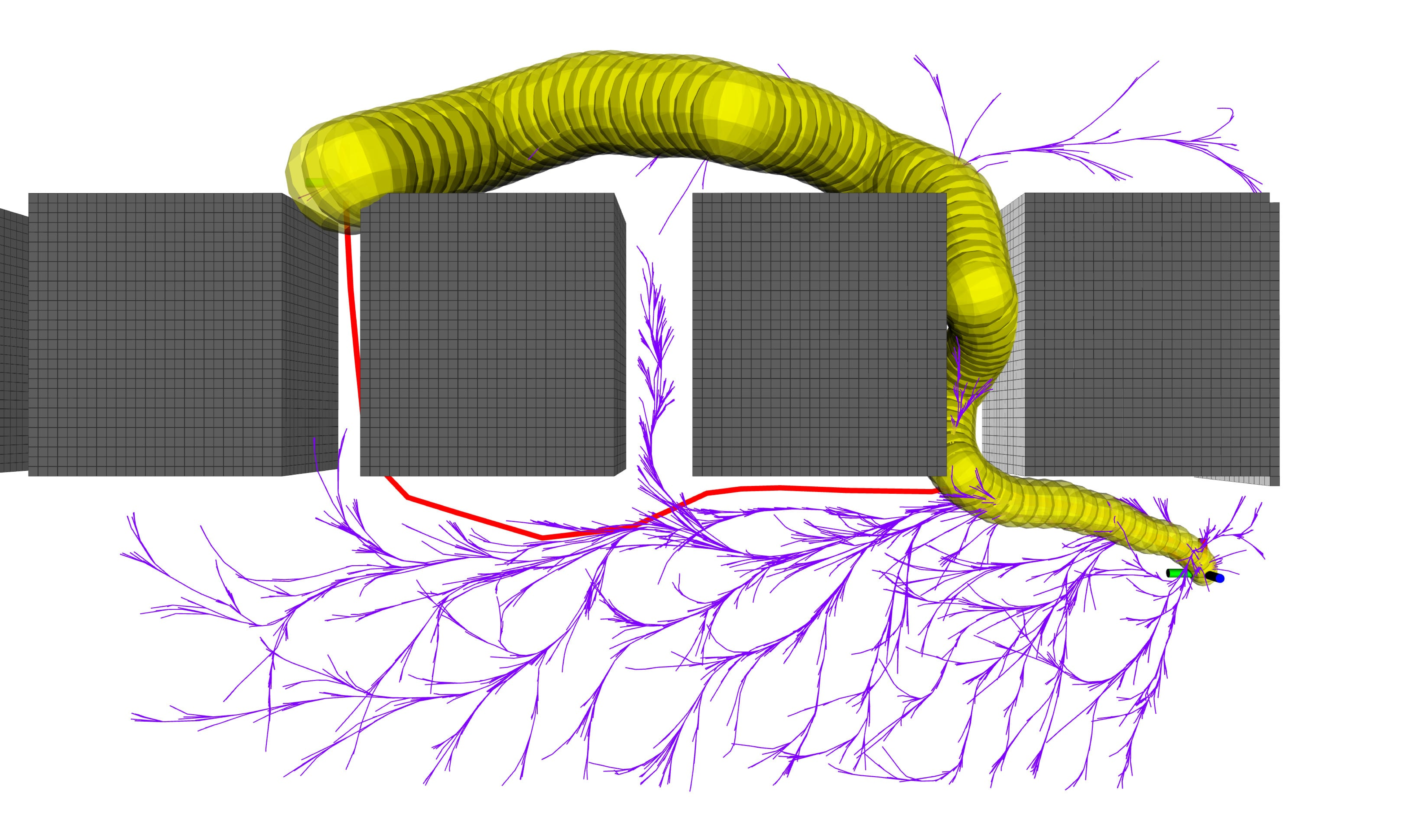}\label{fig:mlp_dyn_rad_2t}}
            \caption{Probabilistic completeness of the proposed multi-layered planning scheme with adaptive lead, which \mbox{(a-b)}~promotes finding the final solution in the neighbourhood of the asymptotically optimal lead $\traj^\prime$ (red), while it \mbox{(c-d)}~preserves completeness guarantees even when the lead $\traj^\prime$ transverses a corridor which does not offer a probabilistic safe passage.}
            \label{fig:mlp_completeness}
        \end{figure}

        The performance of both approaches in comparison to a rigid $\Xlead$ strategy is discussed in \sref{sec:evaluation_mlp}. Noteworthy, any of the two presented mixture of sampling strategies ensures probabilistic completeness of the overall multi-layered scheme. As an extreme example, let us consider the scenario depicted in \mbox{\fref{fig:mlp_dyn_rad_2s}-\ref{fig:mlp_dyn_rad_2t}}, where the lead planner finds an asymptotically optimal solution through the farthest (most left) corridor. However, according to the probabilistic safety constraints defined in \sref{sec:planning_pc}, such a corridor does not offer any safe passage. Despite the initial bias towards this unsuitable $\Xlead$, a mixture of sampling strategies, as the ones introduced in this section, permits finding a solution if one exists provided enough time, thus ensuring probabilistic completeness guarantees.
        
    % ===============================
    % ===============================
    % ===============================
    \subsection{Planning Under Motion Constraints \label{sec:planning_kc}}

\begin{figure}[b]
            \centering
            \includegraphics[width=6cm]{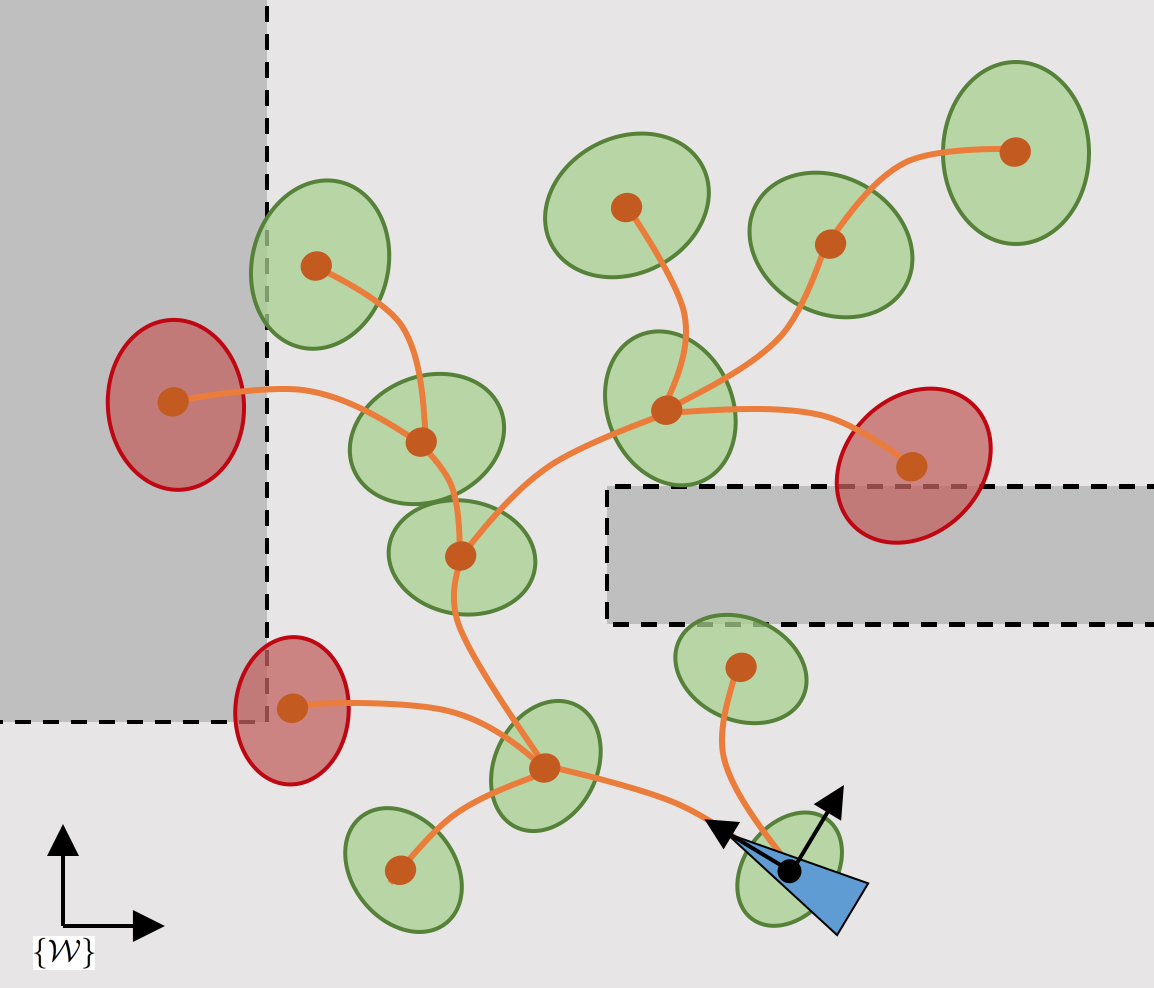}
            \caption{Tree expansion under motion and probabilistic constraints. The state beliefs (nodes) of the tree are obtained by considering the motion capabilities. The ellipses surrounding the states represent their uncertainty, where green corresponds to those states satisfying the probabilistic safety constraints, and red those that do not.}
            \label{fig:tree_expansion}
        \end{figure}

	    The system's motion capabilities are considered in the constrained planner by expanding a tree with the system's motion model~\mbox{\eqref{eq:generic_km}-\eqref{eq:generic_cov}}. 
	    %Examples of motion models for unicycle and fixed-wing dynamics are provided in \xref{sec:kinematic_models}. 
	    In particular, the constrained planner employs the \ac{SST} algorithm~\cite{li2016asymptotically} to build a tree in the belief space
	    %, whose nodes are described as state beliefs ${x \sim b = \N(\hat{x}, \; \cov_{x})}$.
	    with state beliefs ${x \sim b = \N(\hat{x}, \; \cov_{x})}$ as nodes.
	    The tree expansion is based on two procedures: $\textit{sample}(\cdot)$ and $\textit{extend}(\cdot)$, which are conducted in the state and belief space, respectively (see \fref{fig:mlp_bigpicture_scheme}). That is, $\textit{sample}(\cdot)$ draws a random state $x_{rand} \in \Xlead$, where $\Xlead$ is a subregion of $\X$ as defined in \sref{sec:planning_multi}. The planner then selects a node from the tree to attempt connecting to the randomly sampled state $x_{rand}$. Such a selection is conducted via nearest neighbour in the state space using Euclidean metric. The selected node $x_{near}$ has a probabilistic representation in the belief space, %according to previous tree expansions, 
	    i.e. $x_{near}$ is better described as ${x_{near} \sim b_{near} = \N(\hat{x}_{near}, \; \cov_{x_{near}})}$. Then, from this belief, the $\textit{extend}(\cdot)$ procedure expands the tree in the belief space by evolving the system's motion model~\mbox{\eqref{eq:generic_km}-\eqref{eq:generic_cov}} with a randomly sampled control input ${\vu \in \U}$. This expansion is done for a random period of time $T_{\prop}$. Since the considered motion model includes the system's uncertainty, each obtained belief (tree node) corresponds to a vehicle's state with its associated uncertainty (see \fref{fig:tree_expansion}).
	    
    % ===============================
    % ===============================
    % ===============================
    \subsection{Planning Under Probabilistic Constraints \label{sec:planning_pc}}
        Given a minimum safety probability bound $p_\safe$, the imposed probabilistic constraint requires ${1 - p_\col(b,~\env) \geq p_\safe}$ for every belief $b$ on the trajectory in order to probabilistically guarantee the robot's safety. In our approach, the environment awareness and the relative uncertainties are represented by the cumulative distribution $F_\X^{\planningframe}$, which jointly encodes the density distribution of the perceived environment on a discrete support (see \sref{sec:mapping}). As discussed previously, this representation of the environment favours efficiency for online mapping and planning applications. In fact, such encoding allows to guarantee ${1 - p_\col(b,~\env) \geq p_\safe}$ for each belief $b$ of the tree as:
        \begin{align}
            1 - \left(p_{\col,\confidencelevel}(b,~F_\X^{\planningframe}) + (1 - \confidencelevel)\right) &\geq p_\safe \\
            \confidencelevel - p_{\col,\confidencelevel}(b,~F_\X^{\planningframe}) &\geq p_\safe
            \label{eq:pocc_guarantees}
        \end{align}
        where $\confidencelevel$ is the confidence level on the computation of $p_{\col,\confidencelevel}(\cdot) \in [0, \confidencelevel]$. In other words, $p_{\col,\confidencelevel}(\cdot)$ does not cover a $(1-\confidencelevel)$ span of the belief $b$ over the state space. Therefore, it is assumed that the remaining $(1-\confidencelevel)$ is in collision to ensure probabilistic guarantees on the collision checking decision. All in all, this method can be exploited to trade a constant conservatism $\alpha$ in favour of performance.
        
        The probability of collision of a robot centred belief ${b^{\planningframe} \sim \N(\hat{b}^{\planningframe}, \; \cov_{b}^{\planningframe})}$ with the environment is:
        \begin{align}
            p_{\col,\confidencelevel}(b^{\planningframe}, \, F_\X^{\planningframe}) &= \left\langle \kernel_\confidencelevel\left(\cov_{b}^{{\planningframe}}\right), \nonumber \; F_\X^{\planningframe} \right\rangle_F \\
            &= \vect\left(\kernel_\confidencelevel\left(\cov_{b}^{{\planningframe}}\right)\right)^T \vect\left(F_\X^{\planningframe}\right)
        \end{align}
        where ${\langle \cdot, \; \cdot \rangle_F}$ is the Frobenius inner product of the overlapping region between the $\hat{b}^{\planningframe}$-centred discrete state belief $\kernel_\confidencelevel\left(\cov_{b_k}^{{\planningframe}}\right)$ (see \xref{sec:kernel_construction}) and the cumulative environment awareness $F_\X^{\planningframe}$. The Frobenius inner product is an efficient operation via matrix vectorisation.
        
        %Note that the probability of collision of a belief completely lying on an unknown space would be $0.5$ according to the map initialisation described in see \sref{sec:mapping} \ml{this sentence is unclear!}. In a completely unknown environment, this prior would reject any state for ${p_\safe > 0.5}$, therefore impeding the search of any plan. To favour exploration and focus computational power on known regions, this framework preserves the opportunistic collision checking strategy of our previous work. This is, if $\hat{\belief}$ has not yet being observed, $\hat{\belief}$ is considered to be valid. Such a strategy was originally presented in~\cite{hernandez2016planning, hernandez2019planning} and already employed in our previous work~\cite{pairet2018uncertainty}.
        
        The overall proposed multi-layered planner leads to the exploration tree depicted in \fref{fig:tree_expansion}, whose edges account for the vehicle's kinodynamic capabilities and whose nodes are probabilistically safe subject to the system's localisation, motion and environment uncertainties. Additionally, the expansion of the tree is also subject to states not leading to an inevitable collision, i.e. a state must allow for the vehicle to make a full stop before colliding.
        
        % \ml{should add a pseudocode for the planner and/or a block diagram} \erpaar{is this addressed with Figure 5(a)?} \ml{The figure is OK.}

    \section{Experimental Evaluation} \label{sec:evaluation}
   
    \begin{figure}[b]
	    \centering
	    \includegraphics[width=55mm]{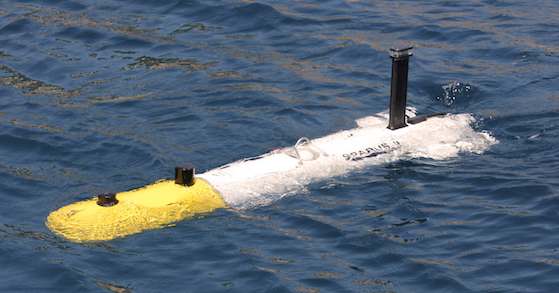}
	    \caption{Sparus~II~AUV, a nonholonomic vehicle.}
	    \label{fig:sparus2}
	\end{figure}

    The proposed framework has been implemented in \ac{ROS} and uses the facilities provided by the Octomap~\cite{hornung13auro} as the core building block of the adopted mapping strategy, and the OMPL~\cite{sucan2012the-open-motion-planning-library} to ease the motion planning requirements. This implementation has been used to evaluate thoroughly the different proposed features and the framework as a whole. This section presents the results of such analysis in an incremental fashion. First, the capabilities of the precedent version of the framework in simulated and real-world scenarios are analysed and discussed in \sref{sec:evaluation_precedent_framework}. Then, \sref{sec:evaluation_mlp} and \sref{sec:evaluation_pcc} report the performance of the key components of the newly proposed framework, i.e. the multi-layered scheme and the probabilistic collision checking. The potential of these components is individually evaluated against closely related state-of-the-art approaches. Finally, the capabilities of the new framework are demonstrated in \sref{sec:evaluation_current_framework} in different environments.
    
    % ===============================
    % ===============================
    % ===============================
    \subsection{Start-to-goal Queries in Undiscovered 2D Environments \label{sec:evaluation_precedent_framework}}
        %The proposed framework has been implemented in \ac{ROS} and uses the facilities provided by the OMPL~\cite{sucan2012the-open-motion-planning-library} to ease the motion planning requirements. To conduct the experiments reported in this section, the framework has been integrated with the \ac{COLA2}~\cite{palomeras2012cola2}, a control software architecture fully developed in \ac{ROS} that controls the \acp{AUV} from the \acf{CIRS}. One of such vehicles is the Sparus~II, a nonholonomic \ac{AUV} used in this work (Fig.~\ref{fig:sparus2}). The Sparus~II is a torpedo-shaped vehicle with hovering capabilities rated for depths up to $200m$~\cite{carreras2015testing}. For the trials presented in this paper, the \ac{AUV} was equipped with a \ac{MSIS} to perceive the surroundings.
    
        The experimentation reported next has been conducted using the precedent version of the online mapping-planning framework presented in this manuscript, which namely was limited to 2D environments and had a single-layered planning strategy~\cite{pairet2018uncertainty}. 
        %Such a framework has been integrated with the \ac{COLA2}~\cite{palomeras2012cola2}, a control software architecture fully developed in \ac{ROS} that controls the Sparus~II \ac{AUV} (see \fref{fig:sparus2}), i.e. the robot used for the experimentation reported below. The Sparus~II \ac{AUV} is a nonholonomic torpedo-shaped vehicle with hovering capabilities rated for depths up to $200m$~\cite{carreras2015testing}.
        Such a framework has been deployed on the Sparus~II \acf{AUV} (see \fref{fig:sparus2}), a nonholonomic torpedo-shaped vehicle with hovering capabilities rated for depths up to $200m$~\cite{carreras2015testing}. The \ac{AUV} is 
        %equipped with a \ac{MSIS} to perceive the surroundings and 
        limited to operate at a constant depth, i.e. in $\mathrm{SE}(2)$, to meet the limitations of the precedent framework. Under these conditions, the motion model of the Sparus~II can be approximated by a unicycle system as detailed in \xref{sec:km_auv_2d}.  
        
        The \ac{AUV} is equipped with a \ac{MSIS} to perceive the surroundings. The readings of such sensor are used to build a representation of the environment. We use the default parameters in~\cite{hornung13auro} of ${l_{min}=-2}$ and ${l_{max}=3.5}$ corresponding to the occupancy probabilities ${P(\voxel) = 0.12}$ and ${P(\voxel) = 0.97}$, respectively, and ${l_{free} = -0.4}$ and ${l_{occ} = 0.85}$ which corresponds to ${P(\voxel) = 0.4}$ and ${P(\voxel) = 0.7}$, respectively. The decay rate in \eqref{eq:decaying_occlusion} is set to $\gamma=0.8$. The framework's planning time is set to ${\planningtime = 1.5s}$.
    
        %The test-bed to evaluate the proposed approach consists of two environments located in Sant Feliu de Gu\'{i}xols (Spain): (a)~breakwater structure that is composed of a series of concrete blocks ($14.5m$ long by $12m$ width), which are separated by four-metre gaps (Fig.~\ref{fig:scenario_block_real} and Fig.~\ref{fig:scenario_block_simu}), and (b)~rocky formations that create an underwater canyon of $28m$ long (Fig.~\ref{fig:scenario_canyon_real} and Fig.~\ref{fig:scenario_canyon_simu}). Using these environments, three experiments are reported: (i)~comparison of the proposed probabilistic collision checking method with other approaches in the literature, (ii)~evaluation of the overall performance of the framework in the \ac{UWSim}~\cite{prats2012open} and (iii)~validation of the framework in real in-water trials. Experiments~(i)~(ii), and (iii)~are tested in the breakwater structure scenario, while experiment (ii)~is also evaluated in a simulated underwater canyon.
        
\begin{figure}[b]
            \centering
            \subfloat[Real breakwater]{\includegraphics[width=3.5cm]{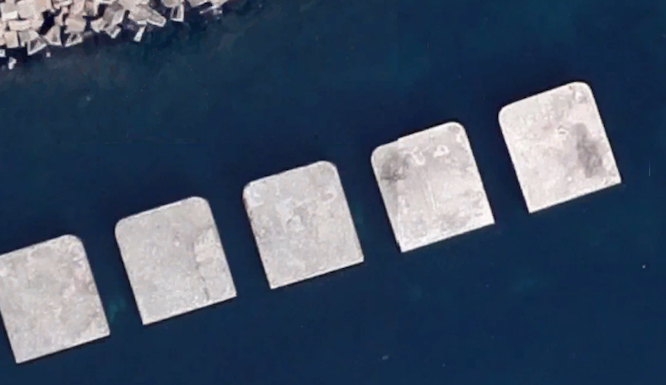} \label{fig:scenario_block_real}}
            \;
            \subfloat[Simulated breakwater]{\includegraphics[width=3.5cm]{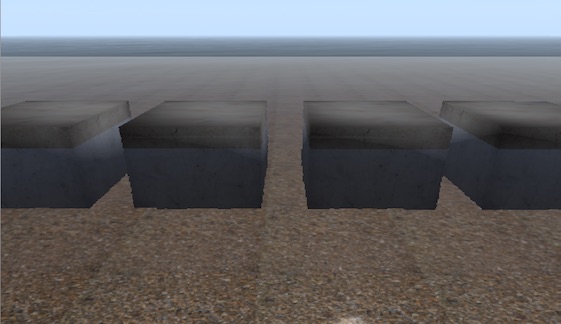} \label{fig:scenario_block_simu}}
            \;
            \subfloat[Real canyon]{\includegraphics[width=3.5cm]{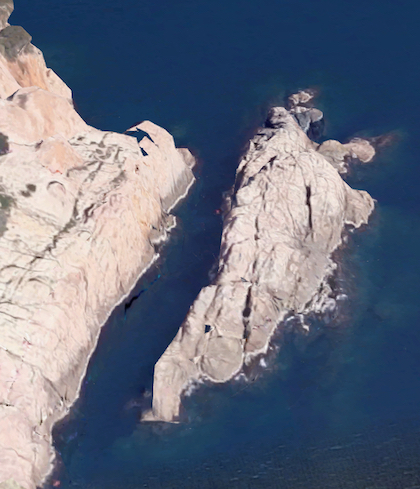} \label{fig:scenario_canyon_real}}
            \;
            \subfloat[Simulated canyon]{\includegraphics[width=3.5cm]{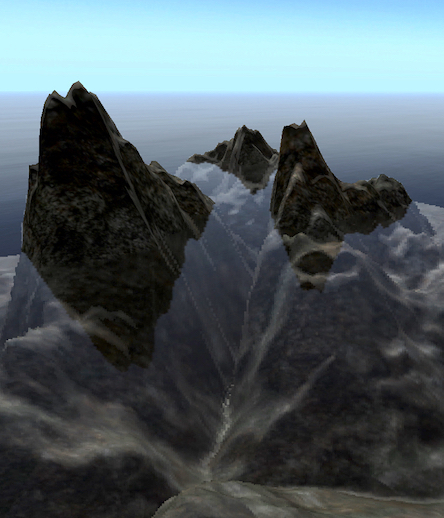} \label{fig:scenario_canyon_simu}}
            \caption{Evaluation scenarios.}
            \label{fig:scenarios}
        \end{figure}

        The test-bed to evaluate the precedent framework consists of two environments located in Sant Feliu de Gu\'{i}xols (Spain): (a)~breakwater structure that is composed of a series of concrete blocks ($14.5m$ long by $12m$ width), which are separated by four-metre gaps (\fref{fig:scenario_block_real} and \fref{fig:scenario_block_simu}), and (b)~rocky formations that create an underwater canyon of $28m$ long (\fref{fig:scenario_canyon_real} and \fref{fig:scenario_canyon_simu}). Using these environments, two experiments are reported: (i)~evaluation of the overall performance of the framework in the \ac{UWSim}~\cite{prats2012open} and (ii)~validation of the framework in real in-water trials. Experiment (i) is conducted in both environments, while experiment~(ii) is uniquely tested in the real breakwater structure scenario.
        
        \subsubsection{Simulated trials:}
\begin{figure}[t]
                \centering
                \subfloat[Initial empty map]{\includegraphics[width=3.4cm, trim=0cm 0cm 0cm 0cm, clip]{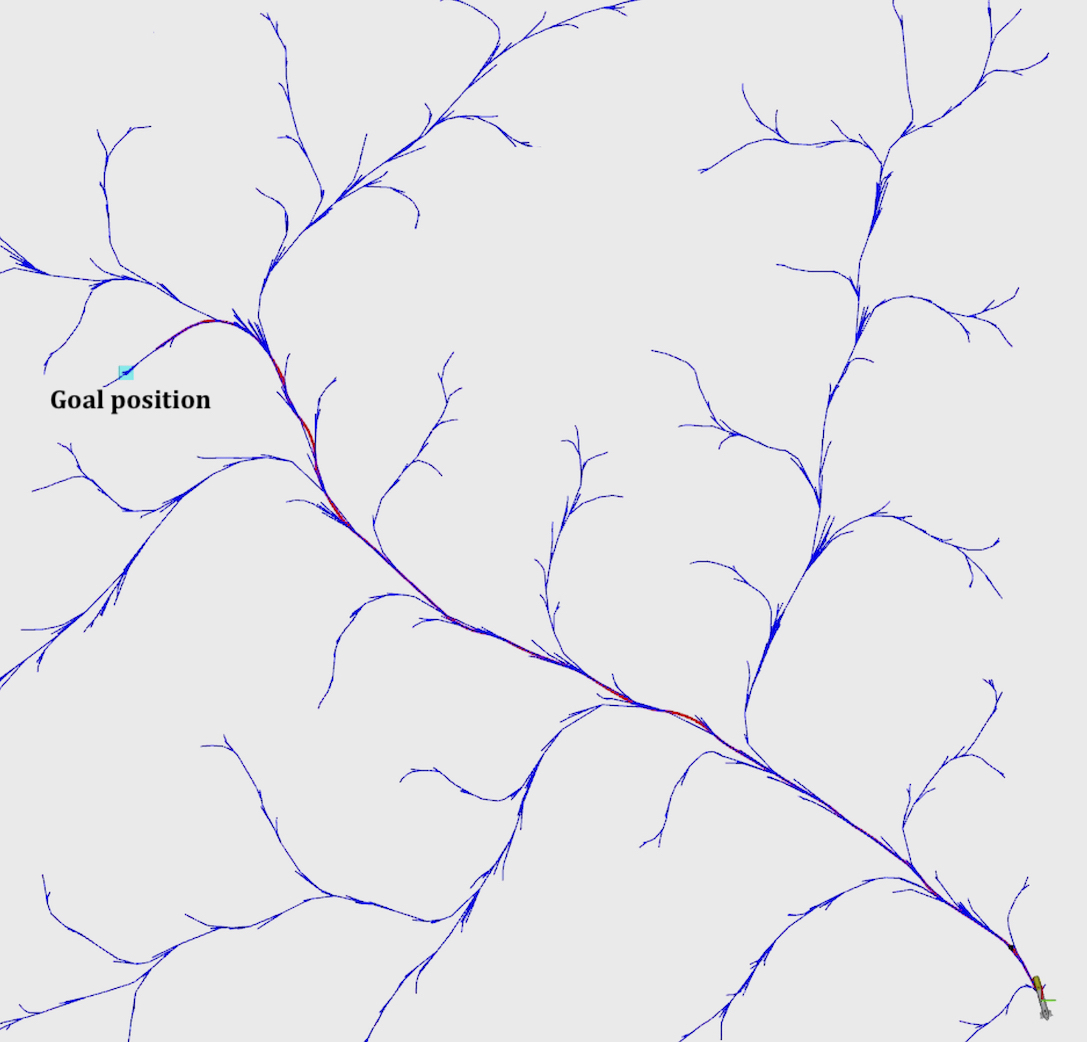}\label{fig:framework_illustration_a}}\quad
                \subfloat[Invalidated trajectory]{\includegraphics[width=3.4cm, trim=0cm 0cm 0cm 0cm, clip]{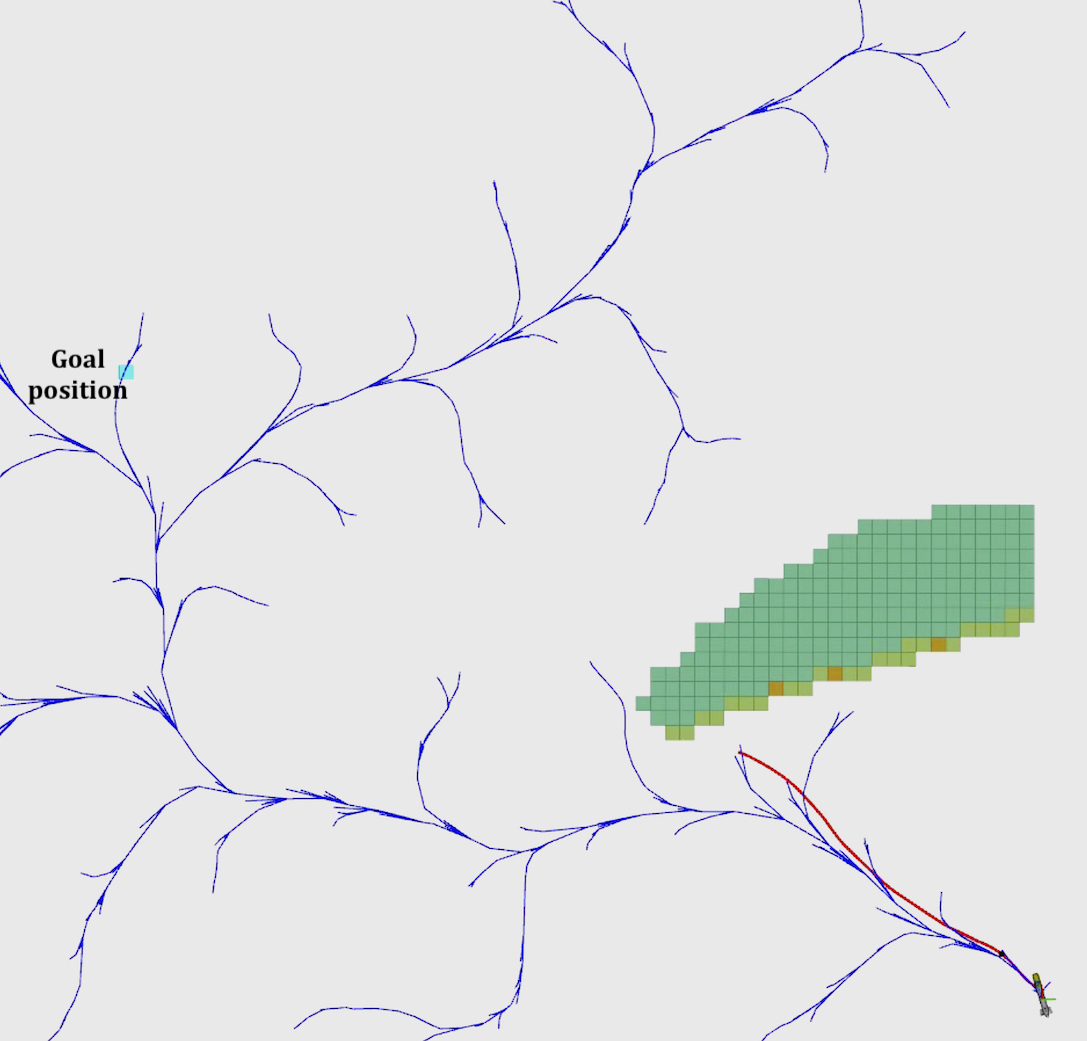}\label{fig:framework_illustration_b}}
                \\
                \subfloat[Replanning the trajectory]{\includegraphics[width=3.4cm, trim=0cm 0cm 0cm 0cm, clip]{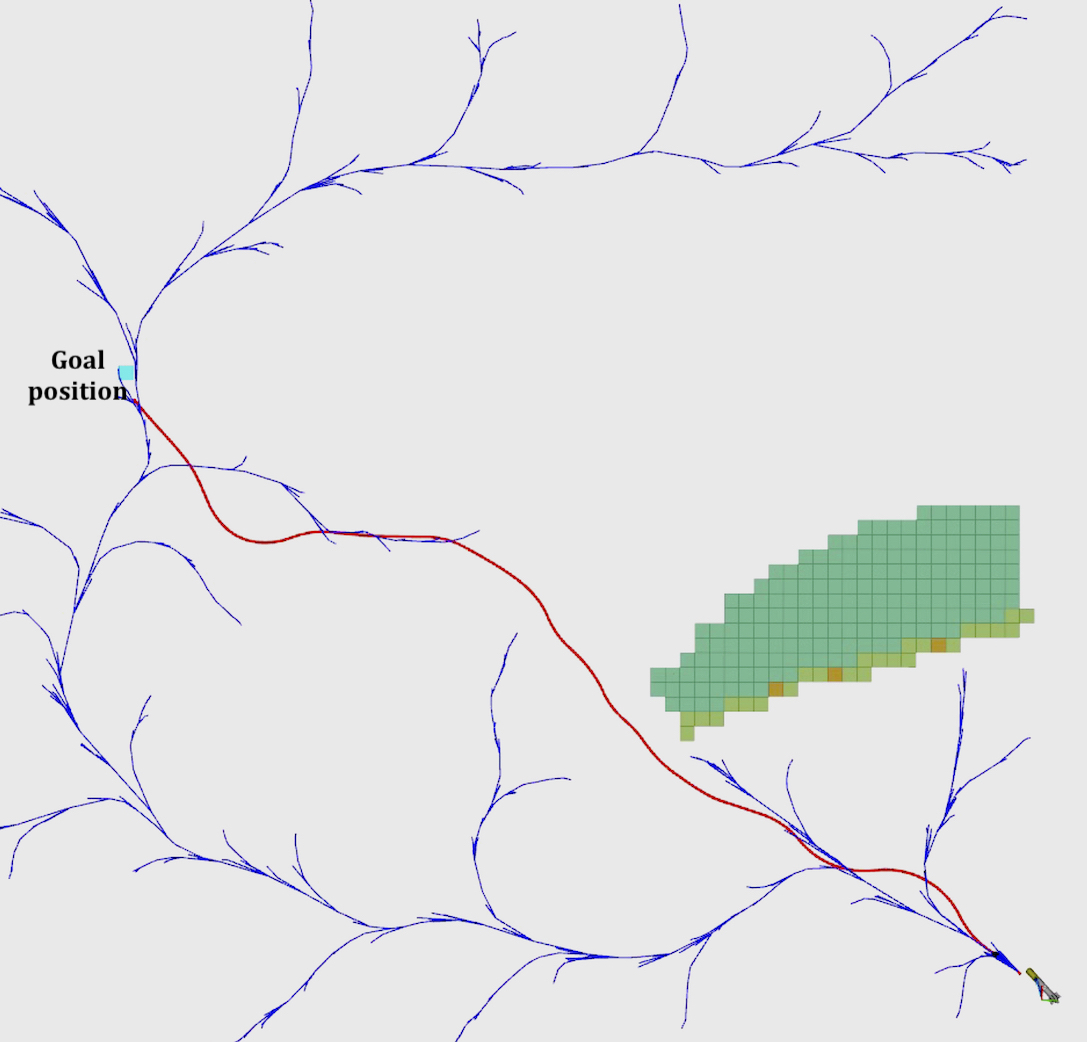}\label{fig:framework_illustration_c}}\quad
                \subfloat[Final part of the survey]{\includegraphics[width=3.4cm, trim=0cm 0cm 0cm 0cm, clip]{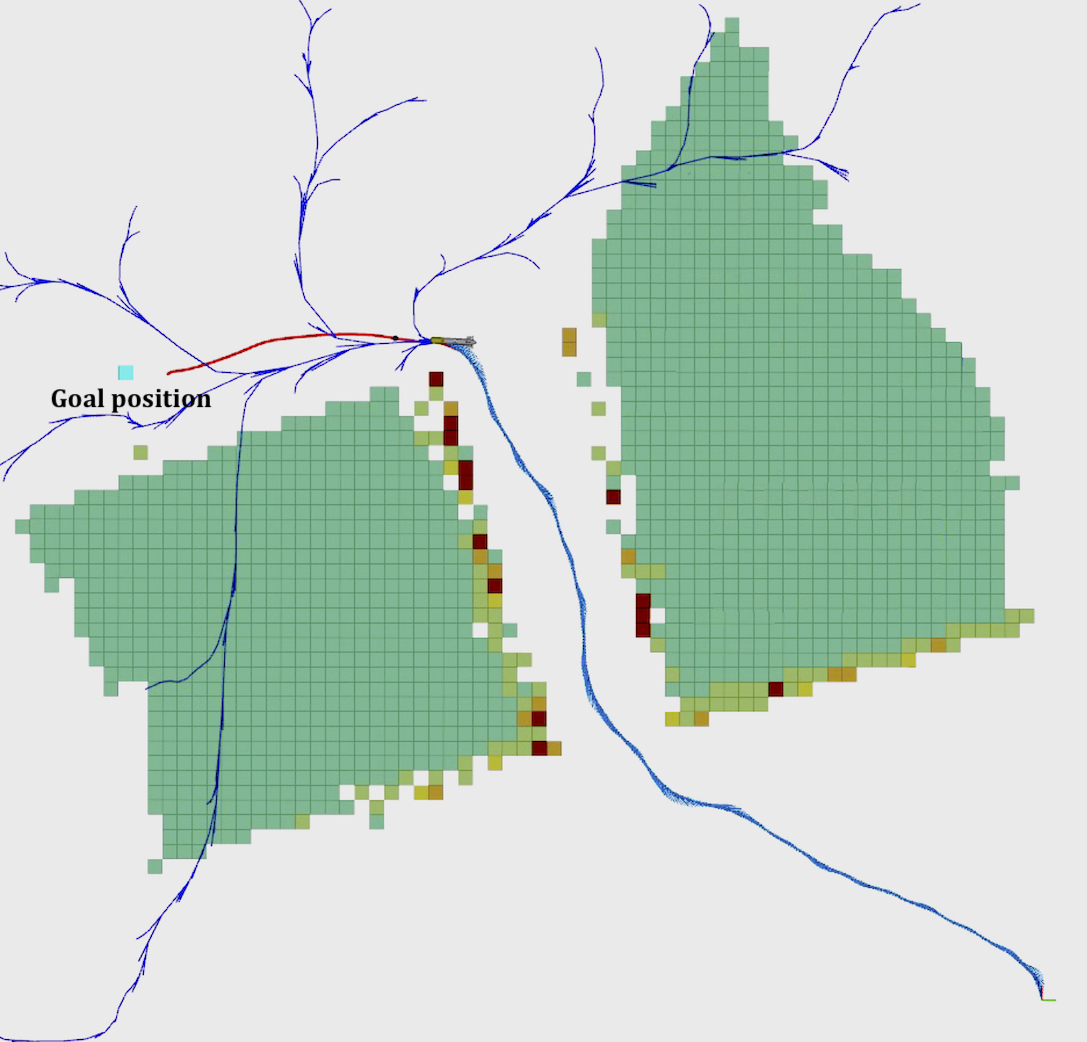}\label{fig:framework_illustration_d}}
                \caption{Incrementally mapping and planning in the undiscovered breakwater structure scenario.}
                
                \label{fig:framework_illustration}
            \end{figure}

            before conducting in-water experiments, the framework was exhaustively tested in the simulated breakwater structure and canyon scenarios. In the former environment, $19$ start-to-goal queries out of $20$ attempts were successfully solved. Among those $19$ successful experiments, the robot achieved the goal region $\Bgoal$ by crossing through the first four-metre gap in $17$ occasions, while in the remaining two trials, the planner found a less optimal trajectory through the second four-metre passage. \fref{fig:framework_illustration} depicts the mission execution in one of those trials. In the initial part of the mission, the environment is completely undiscovered, finding a solution trajectory that goes straight to the goal (\fref{fig:framework_illustration_a}). As soon as the trajectory gets invalidated (\fref{fig:framework_illustration_b}), a new collision-free trajectory is computed (\fref{fig:framework_illustration_c}). After some mapping-planning iterations, the robot gets out of the four-metre gap between two blocks (\fref{fig:framework_illustration_d}). In average, the calculated trajectory towards the goal has a length of approximately $45.2m$ and is completed within $2'21''$.
            
            \begin{figure}[b]
                \centering
                \subfloat[Sparus~II in the \ac{UWSim}]{\includegraphics[width=3.5cm, trim=0cm 0.6cm 0cm 0cm, clip]{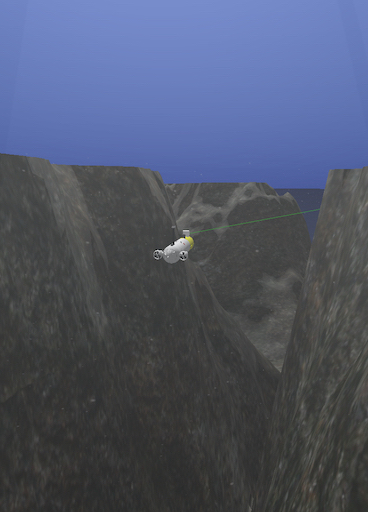}} \quad
                \subfloat[Trajectory calculated in the canyon]{\includegraphics[width=3.5cm, trim=0cm 0cm 0cm 1.5cm, clip]{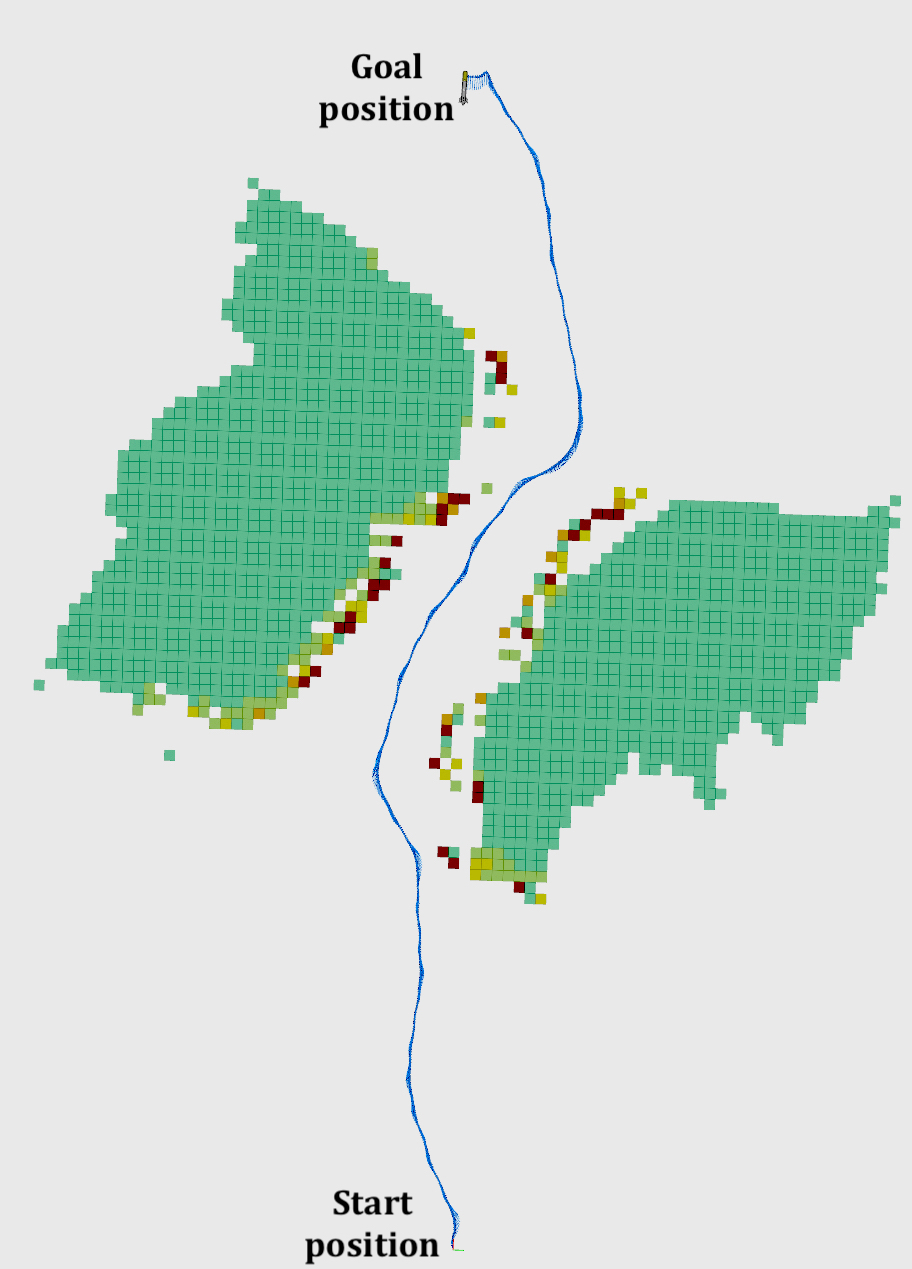}}
                \caption{Incrementally mapping and planning in the undiscovered canyon scenario.}
                \label{fig:experiments_simulated_canyon}
            \end{figure}
            
            The start-to-goal query in the simulated canyon scenario has been successfully solved $20$ times out of the performed $20$ trials. The higher success rate with respect to the previous experiment is given by the nature of the environment; this scenario involves less abrupt manoeuvres and the passage is wider, more than twice larger though. \fref{fig:experiments_simulated_canyon} depicts the trajectory calculated in one of those successful trials through the narrow passage in the middle of the canyon. In average, the calculated trajectories towards the goal have a length of approximately $58.4m$ and are completed within $2'59''$.

            %After those $40$ trials conducted in two different scenarios, the framework has demonstrated to have a high success rate when solving start-to-goal queries in undiscovered environments. This performance is because the planner exploits the entire system's dynamic range, thus allowing the vehicle to execute more complex manoeuvres by adjusting the surge (forward) velocity and prioritising the turning rate.
            After those $40$ trials conducted in two different scenarios, the framework has demonstrated a satisfactory performance to proceed with the deployment in real-world.

        \subsubsection{Real-world trials:}

        	\begin{figure}[t]
                \centering
                \begin{minipage}{0.2\textwidth}
                    \centering
                    \subfloat[Sparus~II during the survey]{\includegraphics[width=3.7cm]{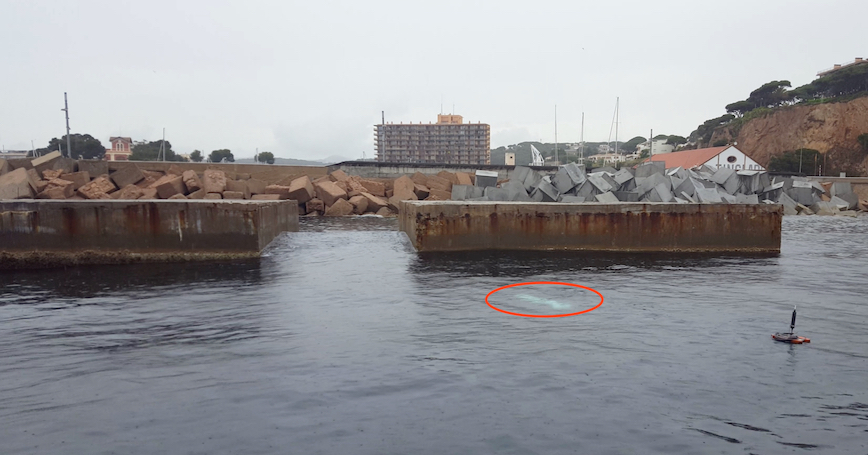} \label{fig:real_experiments_a}} \\
                    \subfloat[Trajectory towards the goal]{\includegraphics[width=3.7cm]{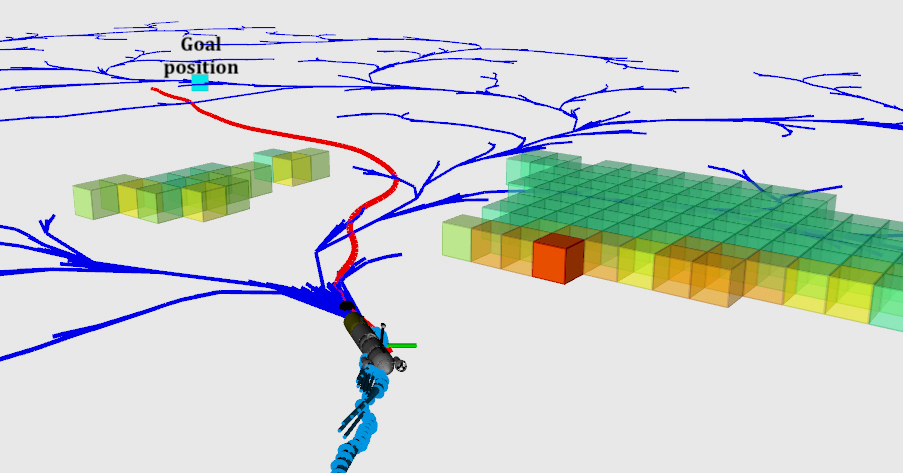} \label{fig:real_experiments_b}}
                \end{minipage}
                \;
                \begin{minipage}{0.25\textwidth}
                    \centering
                    \vspace{0.4cm}
                    \subfloat[Trajectory through the breakwater]{\includegraphics[width=4.3cm, trim=0cm 0cm 4cm 0cm, clip]{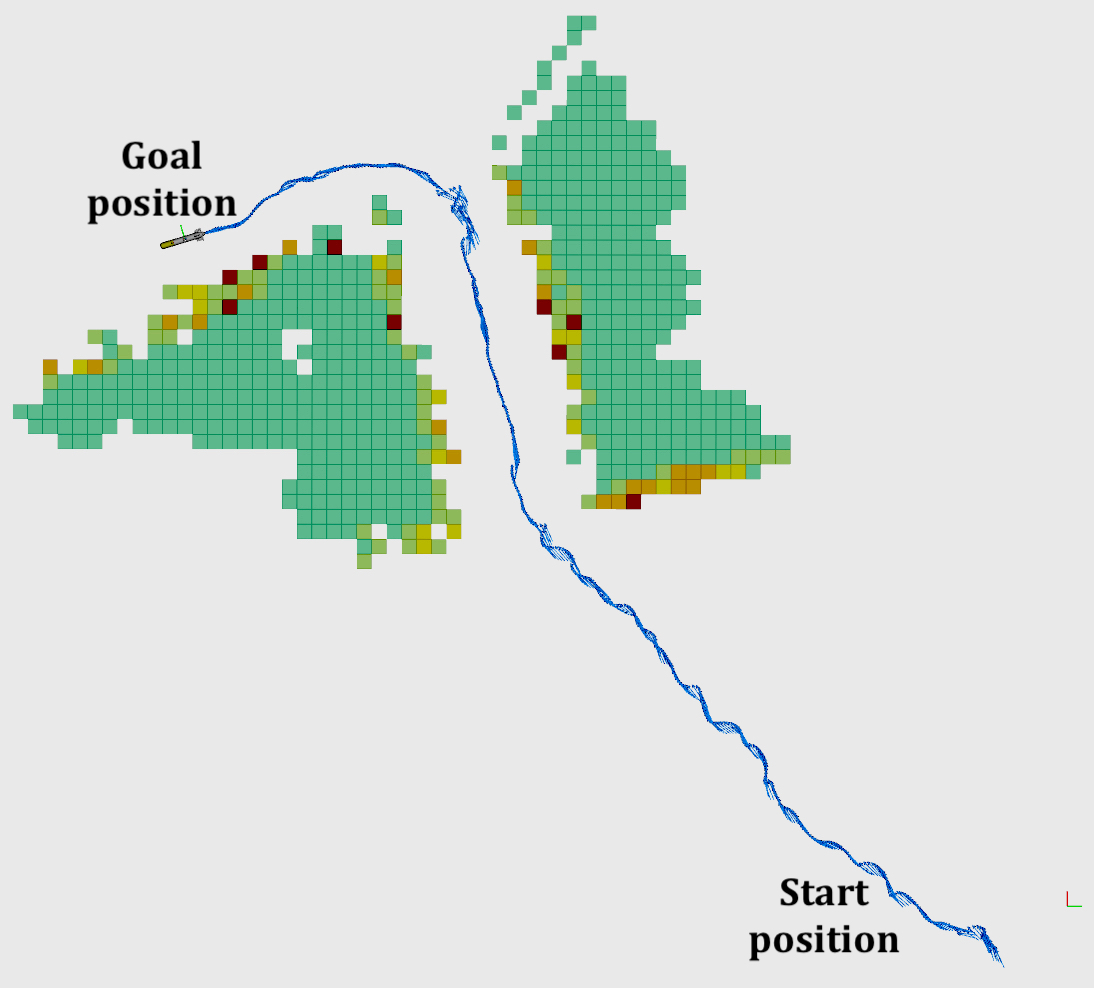} \label{fig:real_experiments_c}}
                \end{minipage}
                \caption{Sparus~II AUV guided by the proposed uncertainty-based framework to solve a start-to-goal query in an undiscovered environment.}
                \label{fig:real_experiments}
            \end{figure}
        
            after the framework had demonstrated a satisfactory performance and high success rate in the simulated trials, it was deployed on the Sparus~II~AUV to prove its suitability for real-world robots with limited on-board computation power. The in-water experiments were conducted in the real breakwater structure located in Sant Feliu de Gu\'{i}xols (Spain). The robot was required to solve a start-to-goal-query to reach a goal region $\Bgoal$ located on the opposite side of the structure, which can only be achieved by navigating through any of the narrow four-metre gaps. During those autonomous missions, the vehicle is connected to a wireless access point buoy for monitoring purposes.
            
            %To undertake the in-water trials, it was necessary to adjust the span view of the \acf{MSIS}. The wider the span view, the more visibility of the surroundings but more time is needed to complete a full scan cycle. A satisfactory trade-off between these two terms was experimentally found when configuring the span from $-60^{\circ}$ to $60^{\circ}$ with respect to the vehicle's direction of motion. Regarding the range of the beam, it has been set at 10 metres to avoid false-positive detections caused by the wide vertical aperture ($40^\circ$) of the sonar.
            
            Prior experiments before running the entire framework revealed that the robot's behaviour when dealing with real conditions diverged from simulations in (i)~the perception quality, being significantly noisier and (ii)~the trajectory tracking performance, being slightly degraded because of the waves and currents. Noise on the observations was reduced by experimentally adjusting the range of the \acf{MSIS} at $10$ metres.
            
            Despite these challenging conditions, the framework successfully accomplished finding and driving the Sparus~II~AUV towards the desired goal region $\Bgoal$ through one of the narrow gaps in the breakwater structure\footnote{A complete sea-trial through the real breakwater structure can be seen in: \url{https://youtu.be/dTejsNqNC00}.}. The trajectory was found through the first corridor four out of five times, while in the other trial the robot went through the second gap. \fref{fig:real_experiments} depicts Sparus~II in one of those in-water trials and the trajectory calculated towards the goal, which has a length of $57.9m$ and took $3'07''$.    

    % ===============================
    % ===============================
    % ===============================
    \subsection{Multi-layered Planning Scheme \label{sec:evaluation_mlp}}
\begin{figure}[b]
            \centering
            %\fbox{\includegraphics[width=0.9\columnwidth, clip]{Figures/mlp_pp.png}}
            \includegraphics[width=0.9\columnwidth, clip]{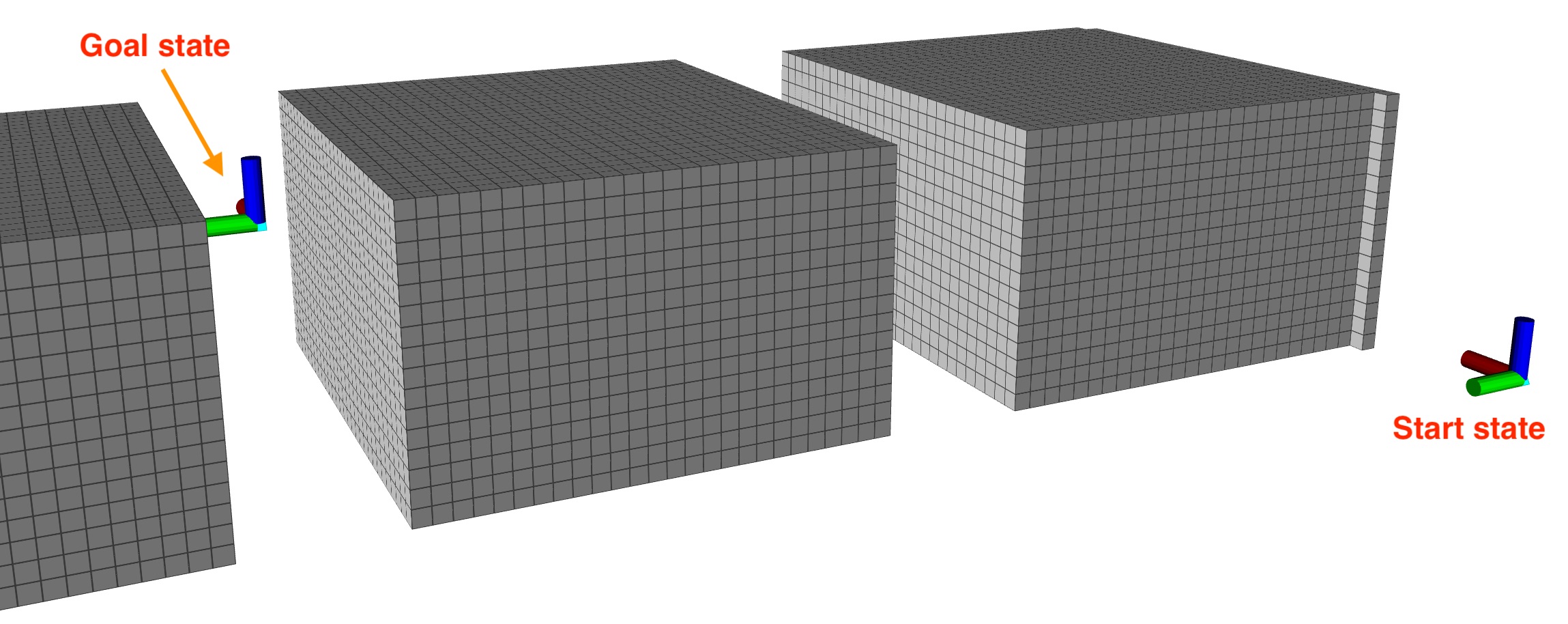}
            \caption{Planning problem to assess the performance of the proposed multi-layered scheme with adaptive $\Xlead$ in comparison to other state-of-the-art approaches. The problem is defined in the belief space for a $\mathrm{SE}(3)$ system operating in a \ac{3D} workspace. The minimum safety bound is set to $p_\safe=0.99$.}
            %${\vx_{start}=(7.0, \; 0.0, \; 1.0, \; 0.0)^T}$ ${\vx_{goal}=(20.0, \; -31.5, \; 6.0, \; 0.0)^T}$ ${p_{\safe}=0.99}$
            \label{fig:mlp_pp}
        \end{figure}

        The multi-layered planning scheme presented in \sref{sec:planning_multi} is one of the key features allowing to overcome the scalability issues of our previous single-layered planner \cite{pairet2018uncertainty}. Nonetheless, differently from current multi-layered approaches which rely on rigid definitions of the candidate search space $\Xlead$ (\mbox{rigid-$\Xlead$}), this manuscript explores two alternative definitions of $\Xlead$ (\mbox{biased-$\Xlead$} and \mbox{adaptive-$\Xlead$}) based on a mixture of sampling experts. This section reports the performance of these four strategies in the scenario depicted in \fref{fig:mlp_pp}, where the planning problem defined in the belief space consists in reaching the state between the blocks while satisfying kinodynamic and probabilistic safety constraints subject to a $p_\safe=0.99$ minimum safety probability bound. In this evaluation, the entire environment is considered to be known in advance and the system dynamics are approximated as described in \xref{sec:km_fixed_wing}.
        
        The four methods (single-layered planner (SLP) and multi-layered planner (MLP) with \mbox{rigid-$\Xlead$}, \mbox{biased-$\Xlead$} and \mbox{adaptive-$\Xlead$}) are evaluated for their ability to quickly find a solution and for the cost of the resulting trajectory. The given total planning time budget is set at ${\planningtime=1.5s}$ to emulate online planning requirements, which is distributed as ${\planningtimescout=0.3s}$ and ${\planningtimetough=1.2s}$ for the three multi-layered schemes. With this setup, each planner attempts to solve the defined planning problem for a total of $2{,}000$ times.
        
        \begin{figure}[t]
            \centering
            \subfloat[Number of successfully solved trials out of $2{,}000$ attempts]{\includegraphics[width=0.95\columnwidth, clip]{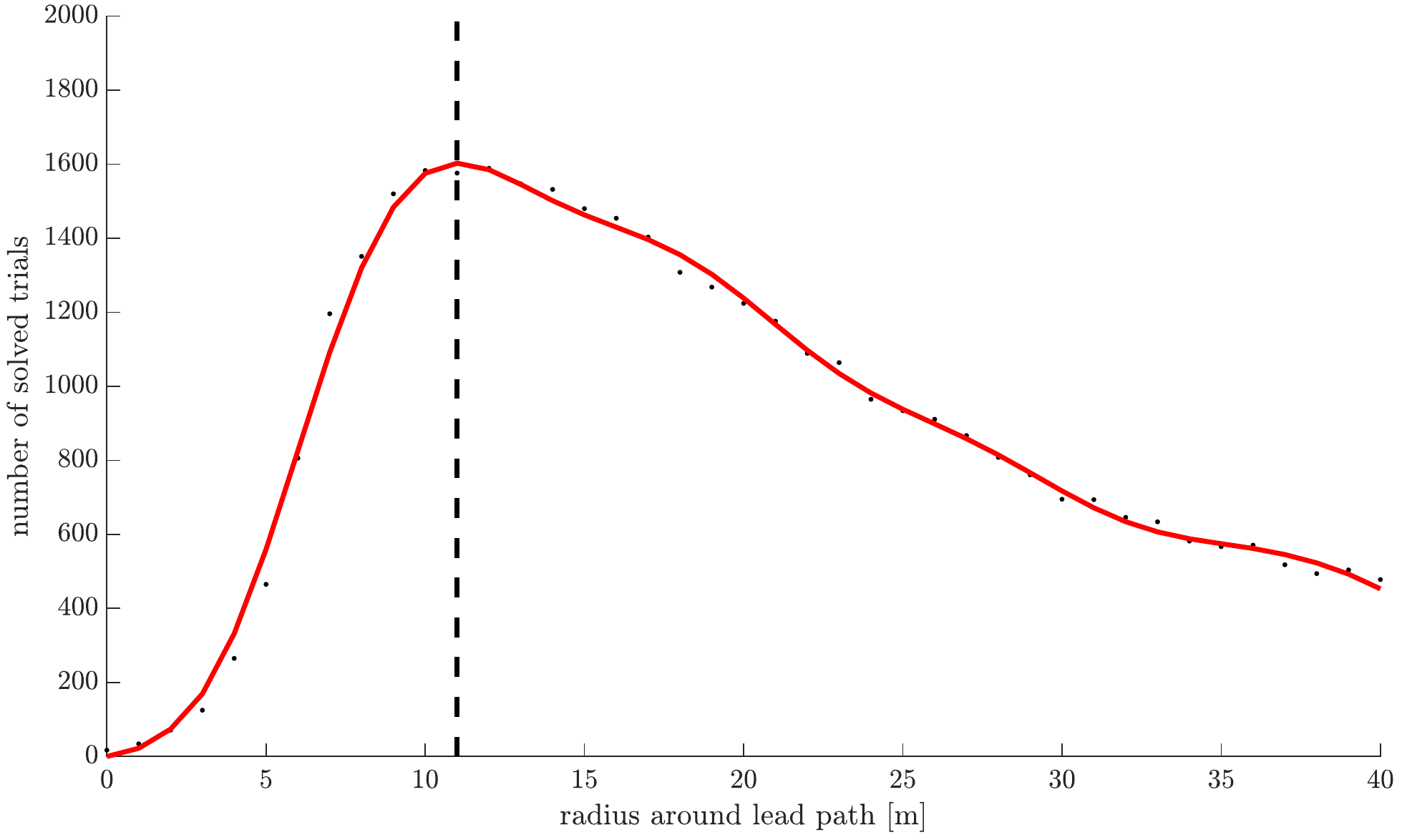}\label{fig:mlp_solved}}
            \\
            \subfloat[Trajectory length of the solved trials]{\includegraphics[width=0.95\columnwidth, clip]{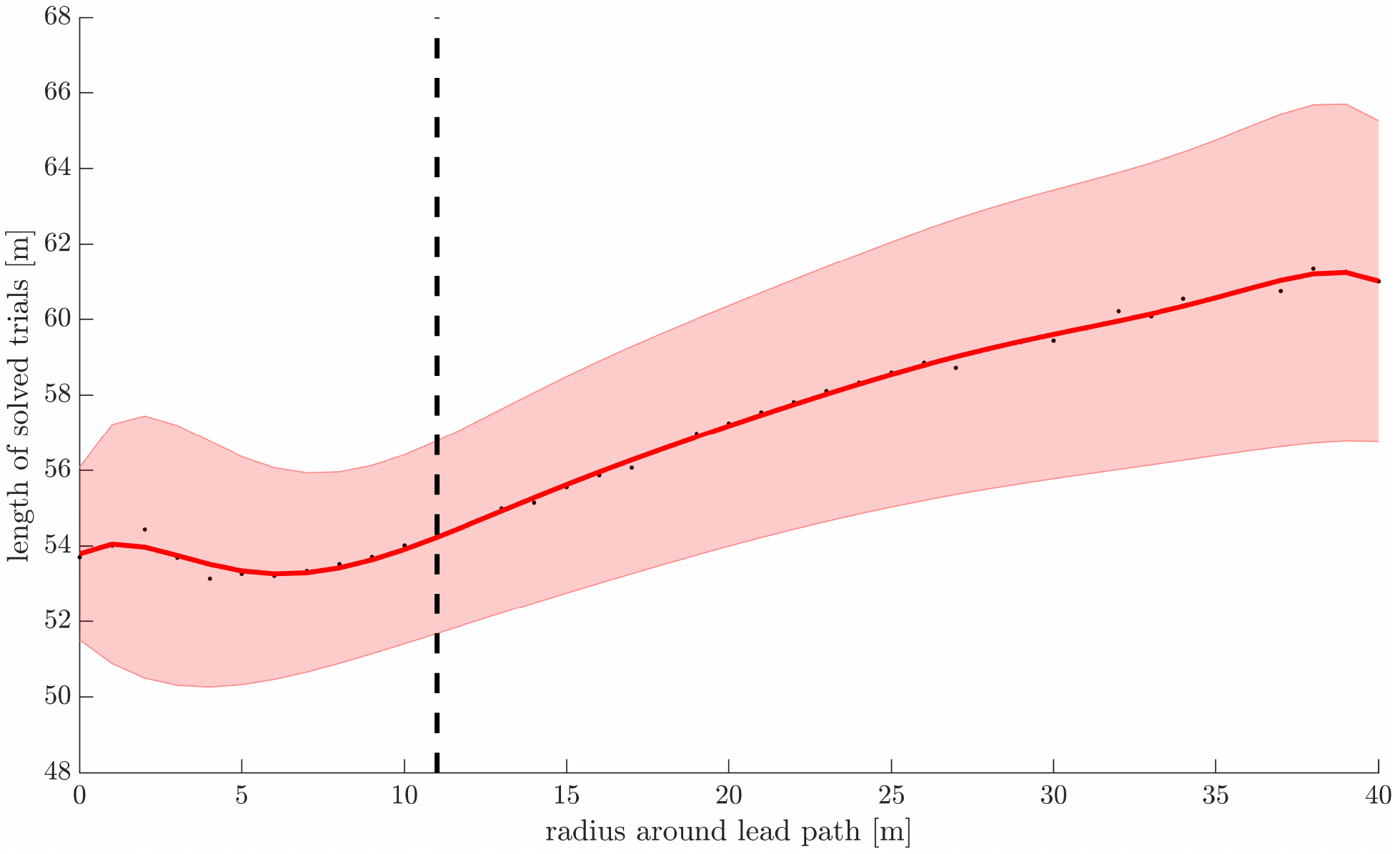}\label{fig:mlp_length}}
            \caption{Performance of the multi-layer planning scheme with a \mbox{rigid-$\Xlead$} lead, i.e. fixed radius around the geometric lead path.}
            \label{fig:mlp_fixed_radius}
        \end{figure}
        
        \begin{figure}[b]
            \centering
            \includegraphics[width=0.95\columnwidth, clip]{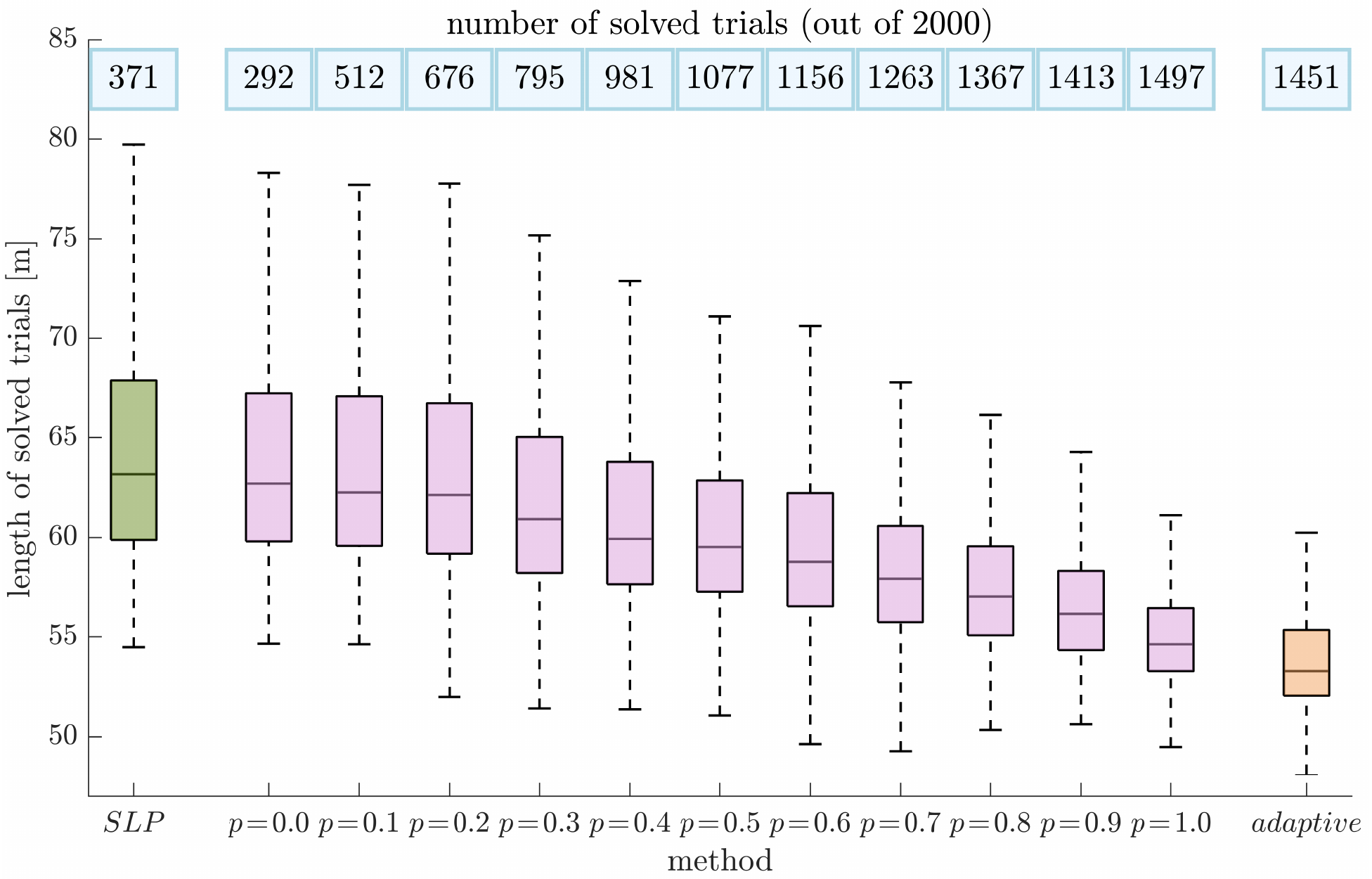}
            \caption{Performance of (i)~our precedent single-layered planning (SLP) scheme (green) and the newly proposed multi-layered scheme when considering (ii)~a fixed lead with different bias $p$ (magenta, with best radius as found in \fref{fig:mlp_fixed_radius}), or (iii)~an adaptive lead as defined in \fref{fig:mlp_scheme} (orange).}
            \label{fig:mlp_boxplot}
        \end{figure}
        
        \begin{figure}[t]
            \centering
            \includegraphics[width=0.9\columnwidth]{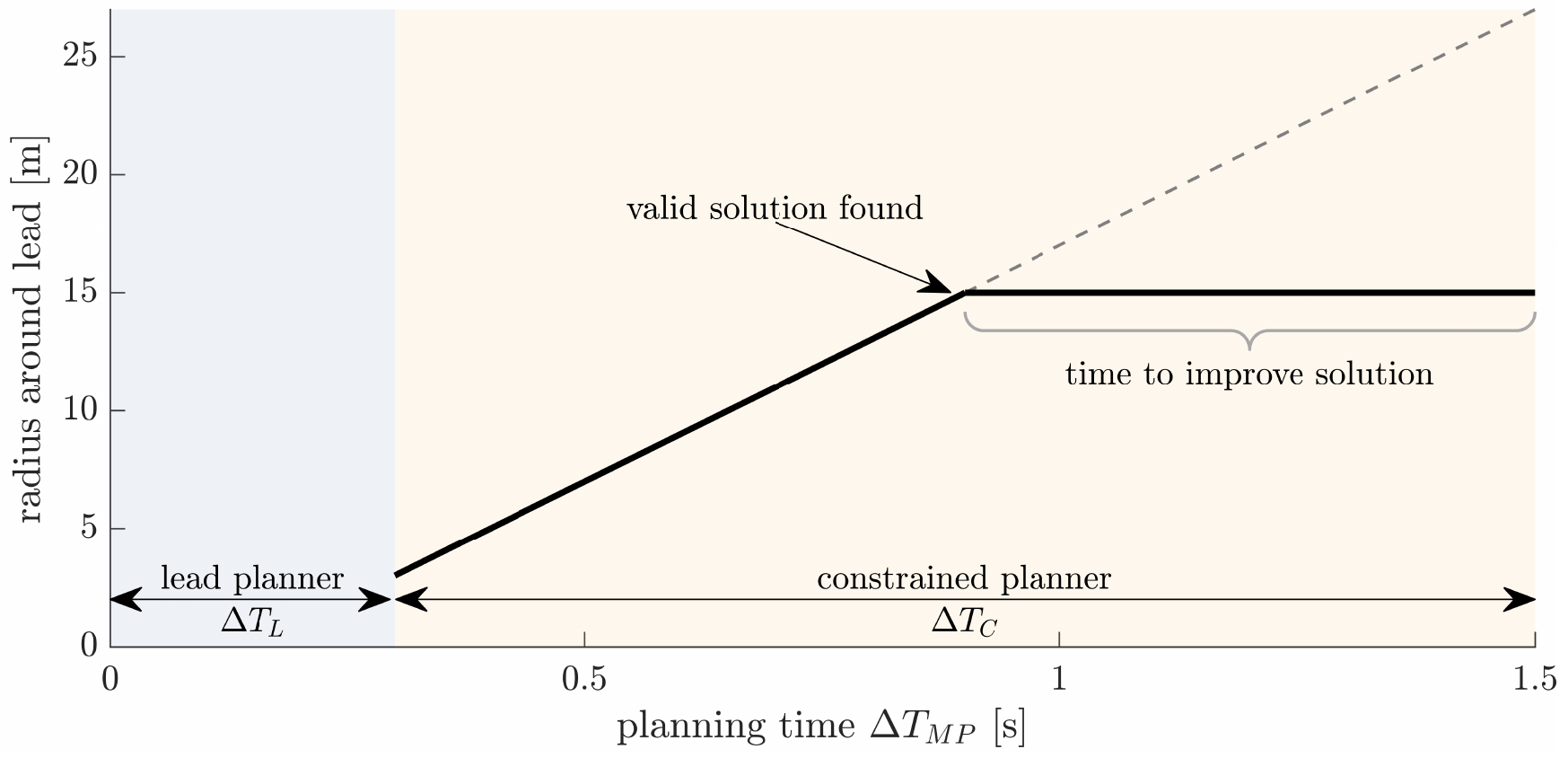}
            \caption{Two layered planning scheme proposed in this work. After computing a lead path, the constrained planner leverages an adaptive $\Xlead$ strategy to initially promote solutions with low cost (small $d$) before ensuring probabilistic completeness by sampling the entire space (${d \rightarrow \infty}$). Once a solution is found, $\Xlead$ is fixed to let the constrained planner refine the found solution until the completion of the planning time $\planningtime$.}
            \label{fig:mlp_scheme}
        \end{figure}
        
        \fref{fig:mlp_fixed_radius} depicts the number of successfully solved trials and the resulting trajectory length when considering a \mbox{rigid-$\Xlead$} lead with radius parameterisations ${d \in [0 \; 40]m}$. While ${d=0m}$ strictly limits the search space to those states forming the lead path, ${d=40m}$ spans the search over all state space of the defined planning problem, therefore resembling uniform sampling. As it can be observed, small search spaces (small $d$) endanger the planner's ability to find a solution with limited time. However, when a trajectory is found, the resulting cost is lower than those solutions found with wider $\Xlead$ leads. Instead, these wide search spaces (big $d$) make the planner struggle at solving most of the planning problems due to the search space extent. In between these two extremes, a suitable parameterisation with ${d=12m}$ (dashed lines) enables solving most of the trials to the planning problem while providing a trajectory with low length cost. Nevertheless, there are not efficient means of defining the optimal $d$ in advance since it is dependant on the planning problem and environment characteristics. Therefore, a \mbox{rigid-$\Xlead$} strategy is not suitable for applications which lack of a fully prior informative representation of the environment. Moreover, too restrictive guided searches can endanger the completeness guarantees of the planner.
        
        The performance of those approaches which guarantee completeness, i.e. the single-layered planner (SLP) (green) and multi-layered planner (MLP) with \mbox{biased-$\Xlead$} (magenta) and \mbox{adaptive-$\Xlead$} (orange) strategies, is depicted in \fref{fig:mlp_boxplot}. In particular, \mbox{biased-$\Xlead$} is parametrised with radius ${d=12m}$ (best lead definition according to experimentation in \fref{fig:mlp_fixed_radius}) and analysed for different ${p \in [0 \; 1]}$, whereas \mbox{adaptive-$\Xlead$} is defined as shown in \fref{fig:mlp_scheme}, i.e. with an initial radius ${d=3m}$ which increases at a rate of ${20m/s}$. This naive implementation of \mbox{adaptive-$\Xlead$} adjusts $d$ from a strictly guided sampling to a uniform search such that as ${t \rightarrow \infty}$, ${d \rightarrow \infty}$, i.e. ${\Xlead \rightarrow \X}$ (uniform sampling).
        
        As it can be observed in \fref{fig:mlp_boxplot}, our precedent single-layered planning scheme struggles at finding a solution on most of the trials. This is because sampling uniformly the entire high-dimensional belief space requires more time to find a solution than the affordable time budget in online applications. Slightly worse performance is obtained when using a multi-layered scheme with \mbox{biased-$\Xlead$} and ${p=0}$ because it still uses uniform sampling but with a portion of the total planning time budget. However, as ${p \rightarrow 1}$, i.e. the planner is more guided to the lead $\Xlead$ (whose optimal radius has been determined empirically in \fref{fig:mlp_fixed_radius}), the performance of the planner increases, in both number of solved trials and length of the final solution. Interestingly, the proposed adaptive sampling method endows the framework with a competitive success rate and solution length than when hand-defining the optimal radius. 

    % ===============================
    % ===============================
    % ===============================
    \subsection{Comparison of Probabilistic Collision Checking Methods \label{sec:evaluation_pcc}}
        \begin{figure*}[t]
            \centering
            \includegraphics[width=0.85\textwidth, clip]{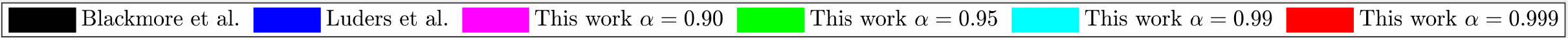}\label{fig:cc_legend}\\
            
            \subfloat[]{\includegraphics[height=5.5cm, clip]{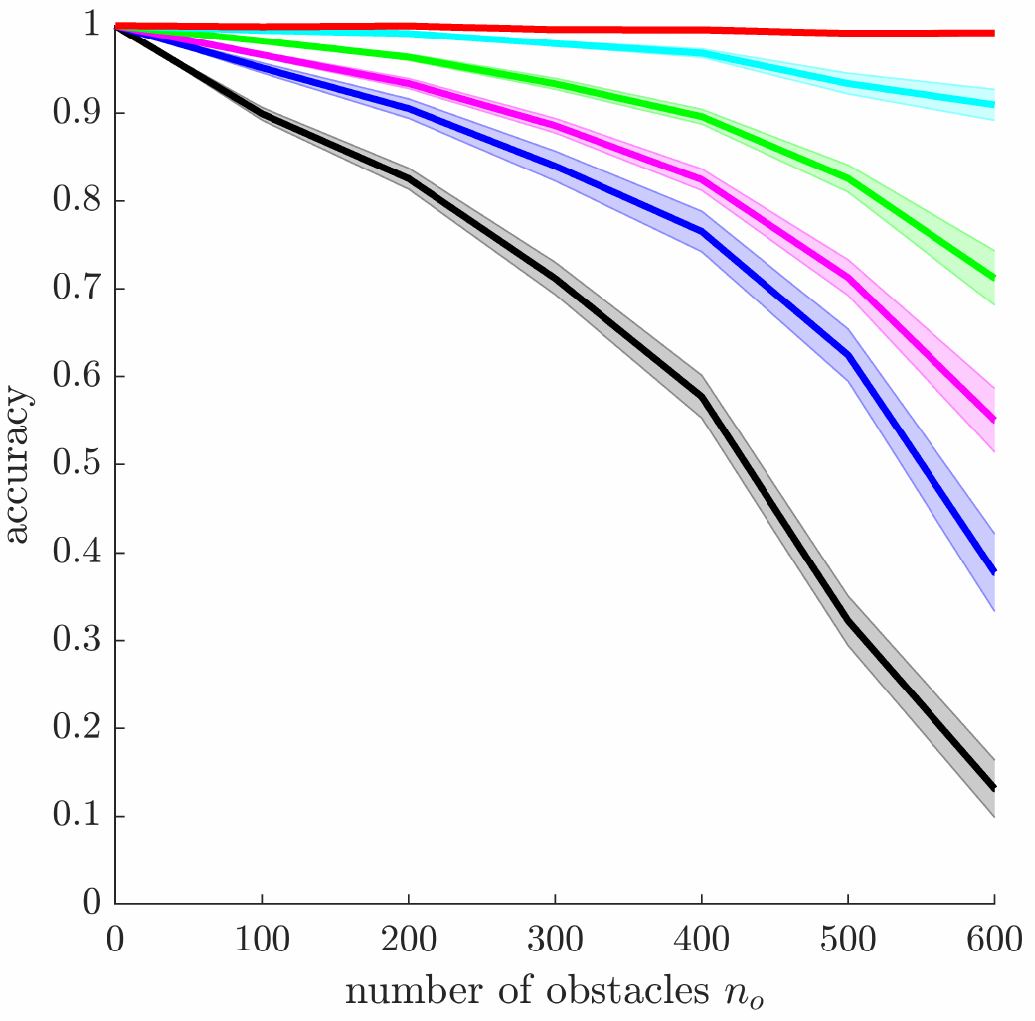}\label{fig:cc_obs_acc}}\;
            \subfloat[]{\includegraphics[height=5.5cm, clip]{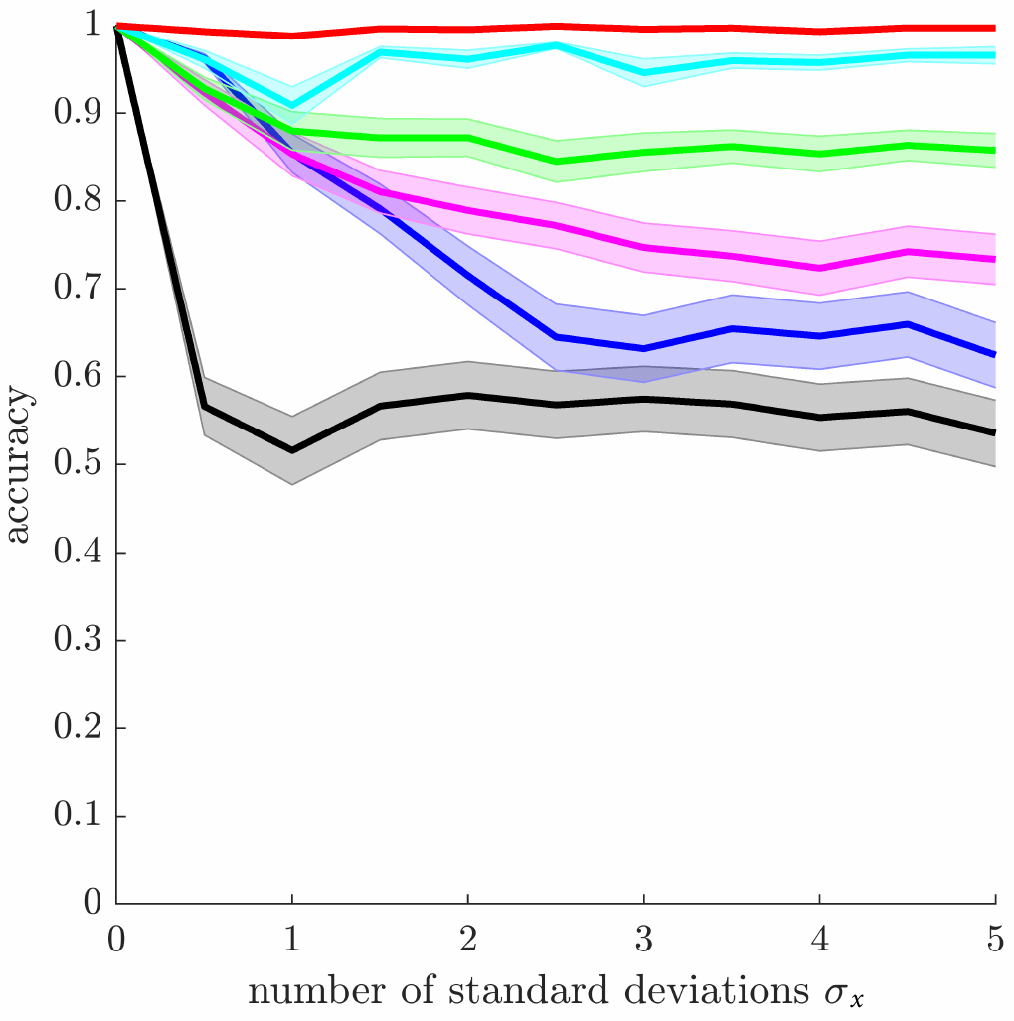}\label{fig:cc_std_acc}}\;
            \subfloat[]{\includegraphics[height=5.5cm, clip]{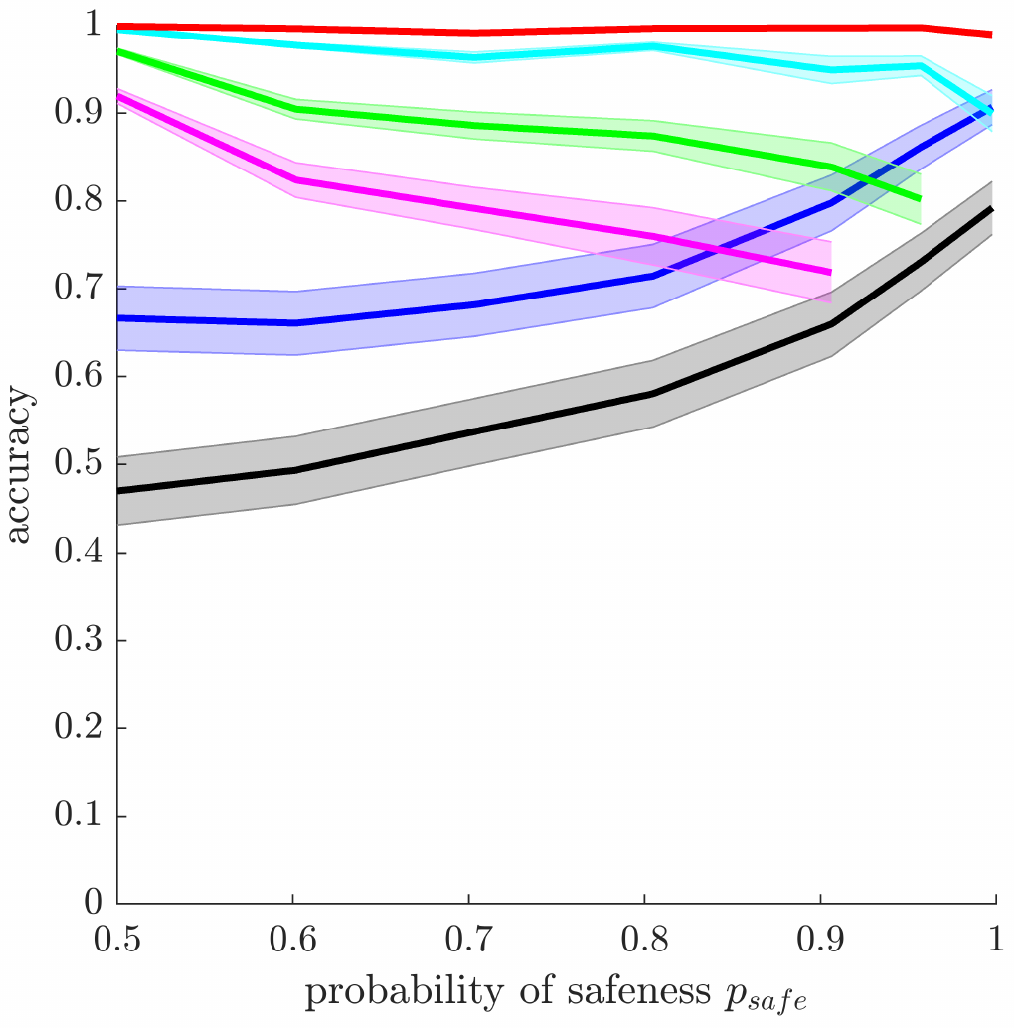}\label{fig:cc_ps_acc}}
            \\
            \subfloat[]{\includegraphics[height=5.5cm, clip]{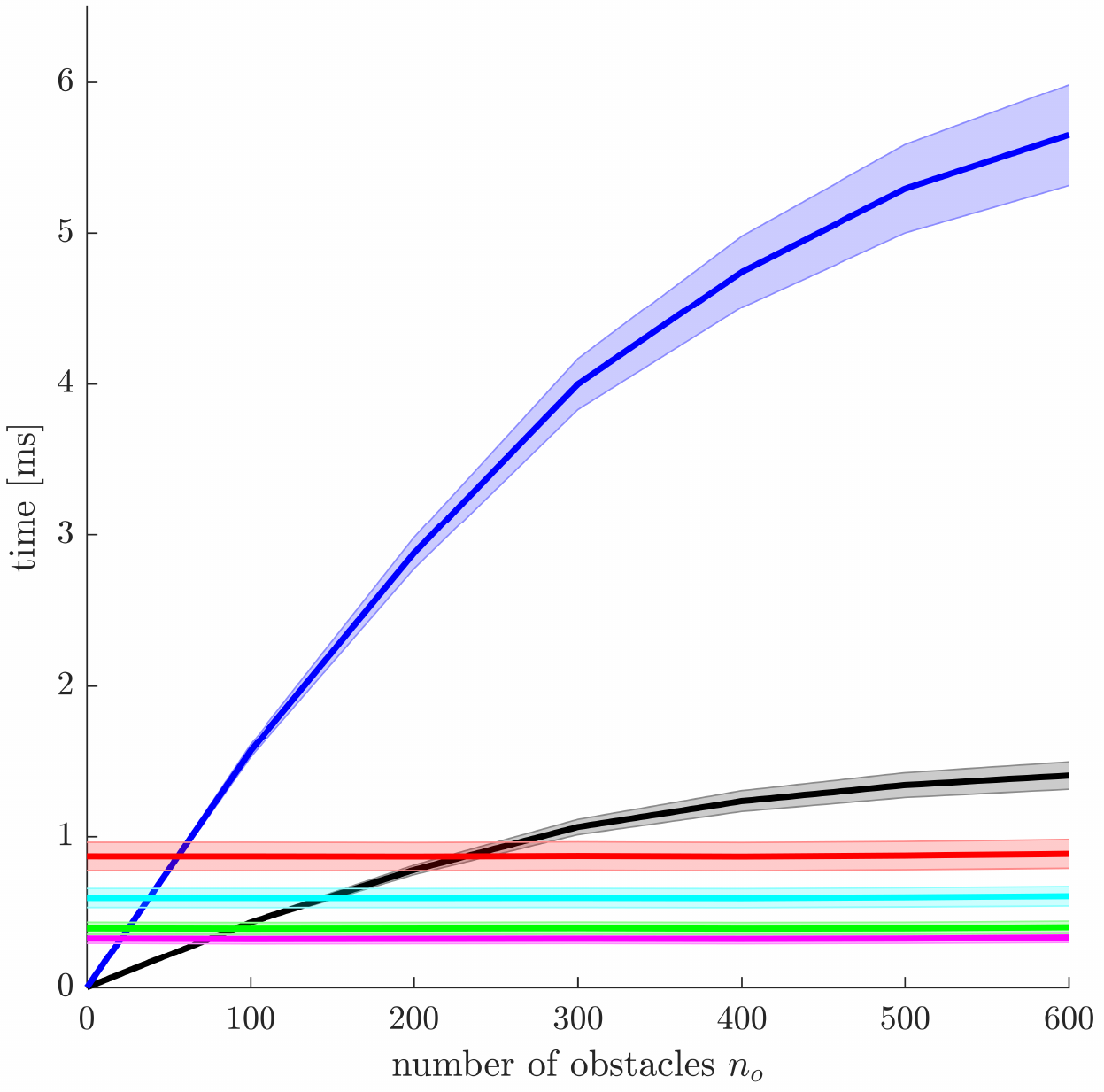}\label{fig:cc_obs_time}}\;
            \subfloat[]{\includegraphics[height=5.5cm, clip]{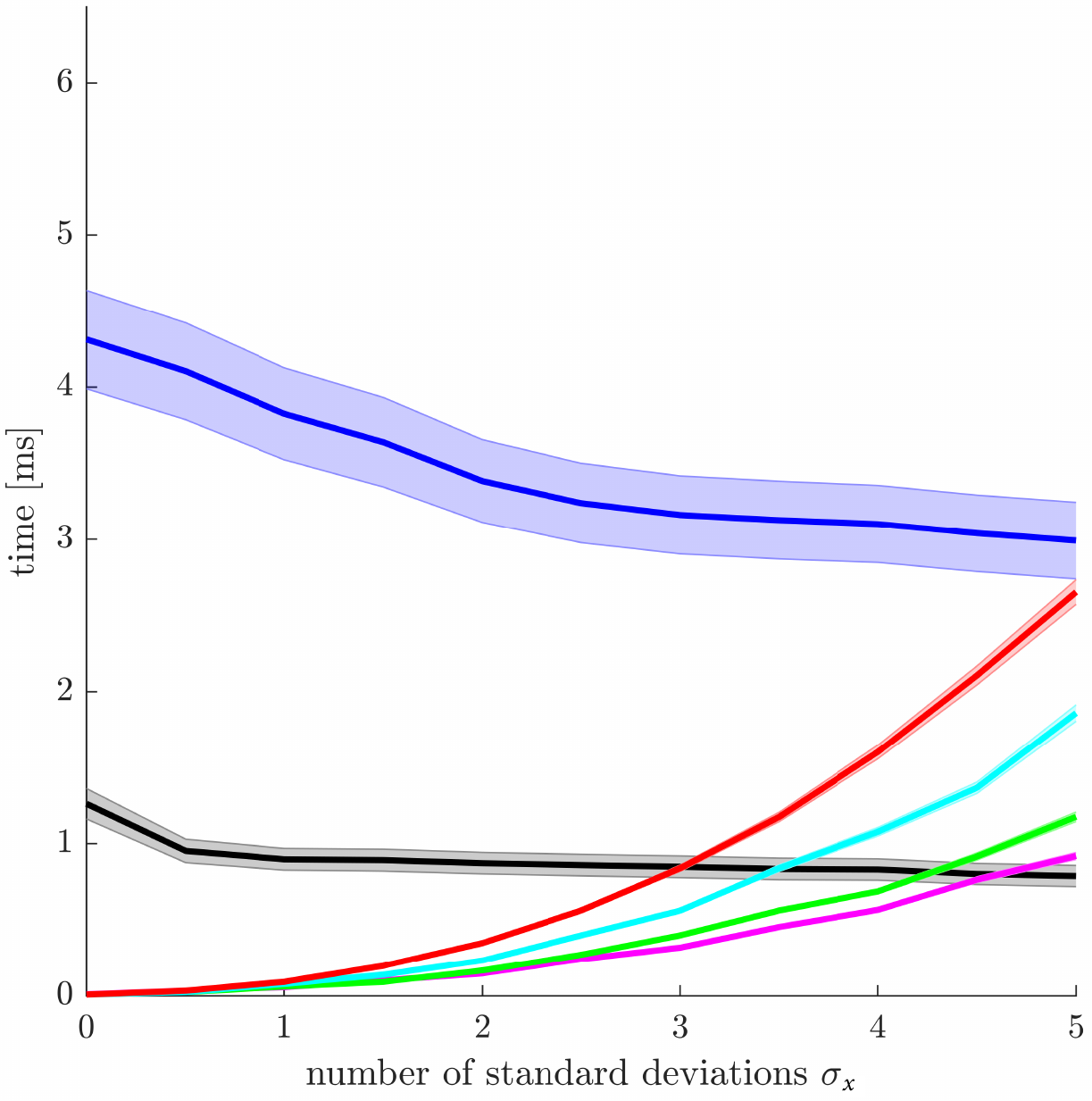}\label{fig:cc_std_time}}\;
            \subfloat[]{\includegraphics[height=5.5cm, clip]{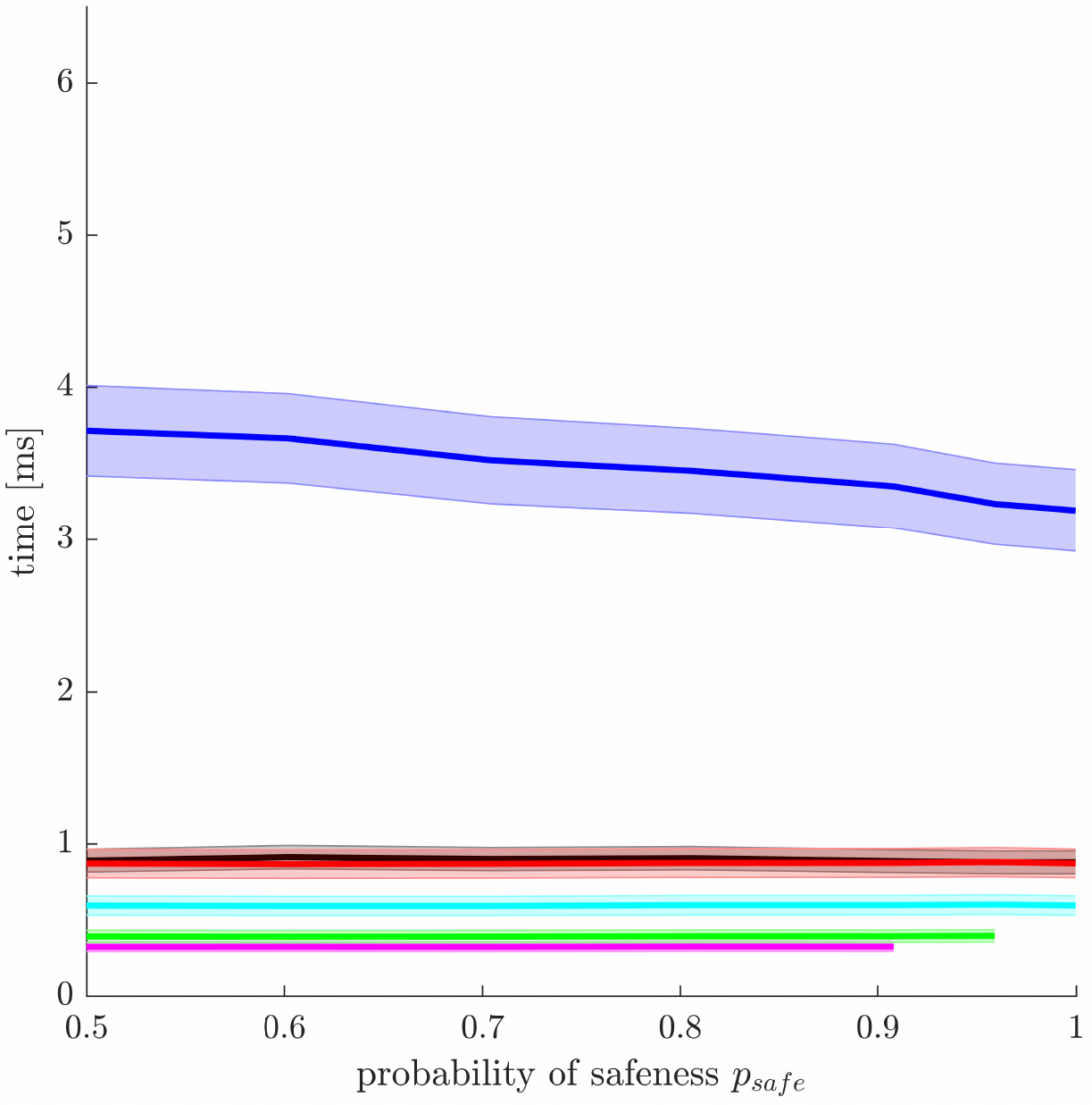}\label{fig:cc_ps_time}}
            \caption{Evaluation of the accuracy (first row) and performance (second row) of the chance constraints formulations~\cite{blackmore2011chance,luders2013robust} and the proposed probabilistic collision checking method with ${\alpha = \{0.90, \; 0.95, \; 0.99, \; 0.999\}}$. The accuracy and performance metrics are represented subject to the number of obstacles $n_o$ in the environment (first column), the state uncertainty ${\cov_\vx = \sigma_x^2 \identity_{3\times3}}$ (second column), and minimum safety probability bound $p_{\safe}$ (third column). The shadowed area corresponds to the variance of the metrics. In the interest of clarity, only one tenth of the variance is displayed.} \label{fig:collision_checking_statistics}
            % https://www.google.com/search?q=matlab+zoom+in+figure&sxsrf=ACYBGNTdOUVkw3wygeRy4SoRTVKNrB3WKw:1568123915385&source=lnms&tbm=isch&sa=X&ved=0ahUKEwi_z8LntMbkAhUZVBUIHaT3ClAQ_AUIEigB&biw=1022&bih=1043#imgrc=F528PYGXRb1uiM:
        \end{figure*}

        Sampling-based planners must be able to analyse the validity of a certain state accurately and efficiently. While accuracy is relevant to avoid discarding regions of the state space which in fact are collision-free, efficiency allows for more space exploration given a limited time budget. However, accurate calculations jeopardise the ability to validate a state rapidly, specially when accounting for uncertainty. In this regard, chance constraints formulations \cite{blackmore2011chance,luders2013robust} offer an interesting accuracy-efficiency trade-off which has proven to be suitable for many motion planning problems in the last decade (see \sref{sec:background}). In fact, chance constraints formulations are still the most widely used probabilistic collision checking method among those state-of-the-art motion planning applications which account for uncertainty (e.g.~\cite{strawser2018approximate,da2019collision}). This motivates the use of chance constraints as the baseline reference to assess the proposed probabilistic collision checking algorithm.
        %\ml{does this mean that the results in subsection 7.2 don't use chance constraint for collision checking?} \erpaar{results in subsection 7.2 use our approach for collision checking. In this sentence I meant that comparing our method against chance constraints is fair given that state-of-the-art planning under uncertainty still uses such an approach.}
        
        % https://www.lexjansen.com/nesug/nesug10/hl/hl07.pdf
        The performance analysis comprises two chance constraints formulations \cite{blackmore2011chance,luders2013robust} and our method with four different parametrisations ${\alpha = \{0.90, \; 0.95, \; 0.99, \; 0.999\}}$. Each method is assessed by its accuracy and efficiency. A method's accuracy is computed as its ability to correctly detect that a state is valid:
        \begin{align}
            \frac{TP}{TP + FN} \in [0, 1],
            \label{eq:cc_accuracy_metric} % this metric is sensitivity
        \end{align}
        where a true positive (TP) indicates that a method's outcome matches the standard of truth\footnote{The standard of truth is approximated by numerical integration of \eref{eq:prob_collision}.}, while a false negative (FN) reflects that the method has mistakenly computed a state as invalid. ${TP + FN}$ is the total number of valid states according to the standard of truth. Therefore, the higher the value of the metric in \eref{eq:cc_accuracy_metric}, the more accurate the method is. For the method's efficiency, the analysis considers the average computation time to process the state validity\footnote{All experiments are performed with an Intel Core i7-7820X CPU @3.60GHz $\times$ 16 with optimised C\texttt{++} implementation for all methods.}. These two metrics are analysed subject to three variables relevant to motion planning problems under uncertainty: (i)~number of obstacles $n_o$ in the environment, (ii)~state uncertainty $\cov_{\vx}$, and (iii)~
        %desired probability of safeness 
        minimum safety probability bound $p_{\safe}$. With this setup, a single case study is parametrised by the triplet ${\langle n_o, \; \sigma_x, \; p_{\safe}\rangle}$. In total, $847$ case studies are retrieved according to the parametrisation span (minimum and maximum values) and discretisation defined in \tref{tab:cc_eval_parameters}.
        
        \begin{table}[t]
            \centering
            \begin{tabular}{lccc}
                \toprule
                & \multirow{2}{*}{\shortstack{Minimum \\ value}} & \multirow{2}{*}{\shortstack{Maximum \\ value}} & \multirow{2}{*}{\shortstack{Discretisation \\ step}} \\
                & & & \\
                \cmidrule{2-4}
                $n_o$ & 0 & 600 & 100 \\
                $p_{\safe}$ & 0.5 & 1.0 & 0.05 \\
                $\sigma_x$ & 0 & 5.0 & 0.5 \\
                \bottomrule
            \end{tabular}
            \caption{Parametrisation for the comparison of probabilistic collision checking methods.}
            \label{tab:cc_eval_parameters}
        \end{table}
        
        Each case study is set up as follows. An environment $\env$ is defined in $\real^3$ with a total of $n_o$ cubical obstacles. In order to have a computational representation of the scene suitable for each method, the environment is encoded as: (i)~a set of linear constraints, where each cubical obstacle is characterised by six constraints, and (ii)~a global occupancy grid map with $0.5m$ resolution. Then, given the known environment $\env$, each probabilistic collision checking method is required to validate, subject to $p_{\safe}$, $10{,}000$ beliefs ${b \sim \N(\hat{\vx}, \, \cov_{\vx})}$. The state estimate ${\hat{\vx} \in \real^3}$ is uniformly sampled over $\X$ and the covariance matrix ${\cov_{\vx} \in \real^{3\times3}}$ is assumed to be diagonal, i.e. ${\cov_\vx = \sigma_x^2 \identity_{3\times3}}$. 
        
        The data extracted from the $847$ case studies is depicted in \fref{fig:collision_checking_statistics}. In the interest of clarity, the corresponding discussion is divided into three parts: accuracy, efficiency and suitability.
     
        % ============
        % acceptance rate
        % ============
        \textbf{Accuracy discussion:}
        the accuracy analysis (first row in \fref{fig:collision_checking_statistics}) depicts that the number of obstacles in the environment is the variable penalising the methods' accuracy the most. This behaviour is due to the methods' conservatism, whose relevance increases with the hardness of the motion planning problem. In other words, the more conservative a method is, the more negatively affected it is. On top of that, the conservatism of chance constraints formulations \cite{blackmore2011chance,luders2013robust} increases with the number of obstacles, whereas our approach accounts for a constant conservatism $\alpha$. This tighter bound allows our method to outperform both chance constraints formulations, even when choosing the most conservative parametrisation ${\alpha=0.9}$. Higher values of $\alpha$ favour accuracy at the cost of more computational expenses (see discussion below). Importantly, the confidence level $\alpha$ of our method should always be set such that ${\alpha \geq p_{\safe}}$, otherwise the constraint in \eqref{eq:pocc_guarantees} will never be satisfied since the analysed part of the space is not sufficient to ensure probabilistic safety. This fact is visible in \fref{fig:cc_ps_acc}, where for ${\alpha < p_{\safe}}$ our method with parametrisation $\alpha$ is not used.
        
        % ============
        % time
        % ============
        \textbf{Efficiency discussion:}
        the efficiency analysis (second row in \fref{fig:collision_checking_statistics}) reflects the expected computational complexity according to the theoretical grounds of each algorithm. This is, chance constraints strategies are fast for scenarios with few number of constraints, but their computational expenses grow linearly as the number of constraints increases. This linear correlation is influenced by the iterative nature of chance constraints, which allows to invalidate a state as soon as ${1 - p_\col(b,~\env) < p_{\safe}}$, i.e. without need to check all constraints. In other words, invalid states involve less time than those which are valid. Consequently, harder planning problems, i.e. those involving more obstacles, higher uncertainties or more restrictive safety guarantees, show a mild deviation towards lower computational time due to the presence of high number of invalid states. In contrast, the computational requirements of our method are uniquely influenced by the state uncertainty $\cov_z$, which determines the number of voxels to include in the calculations (see \sref{sec:planning_pc}). This might restrict the suitability of our approach to systems whose state uncertainty is bounded over time (see discussion below).
        
        %The early stopping strategy also reflects on the variance of the computation time for chance constraints methods.
        
\begin{figure}[t]
            \centering
            \subfloat[\cite{blackmore2011chance}]{\includegraphics[width=0.95\columnwidth, clip]{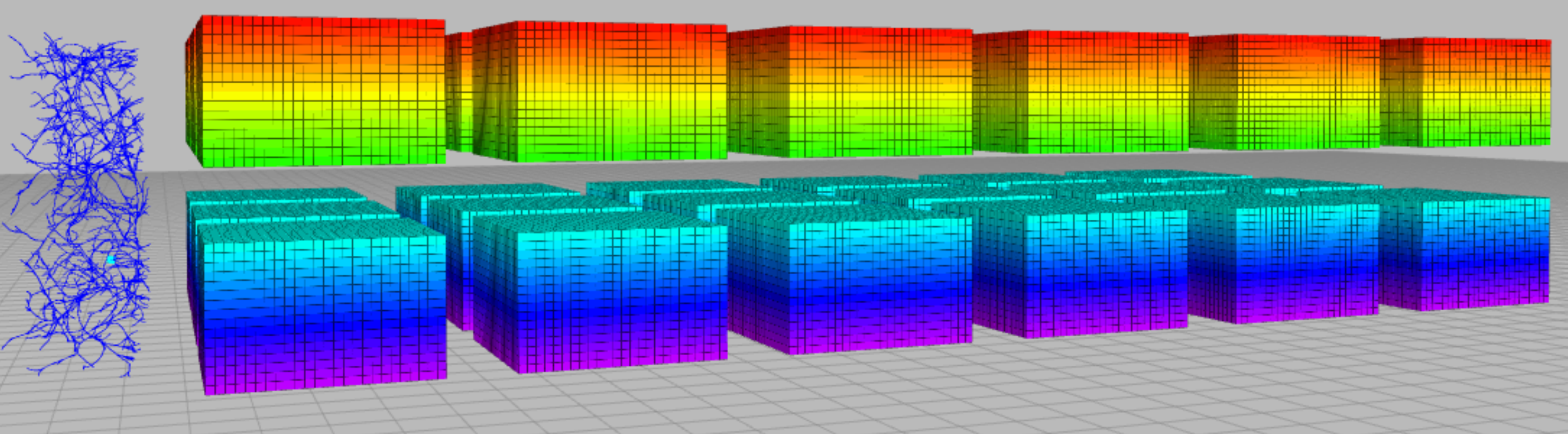}\label{fig:pcc_blakcmore_planning}}
            \\
            \subfloat[\cite{luders2013robust}]{\includegraphics[width=0.95\columnwidth, clip]{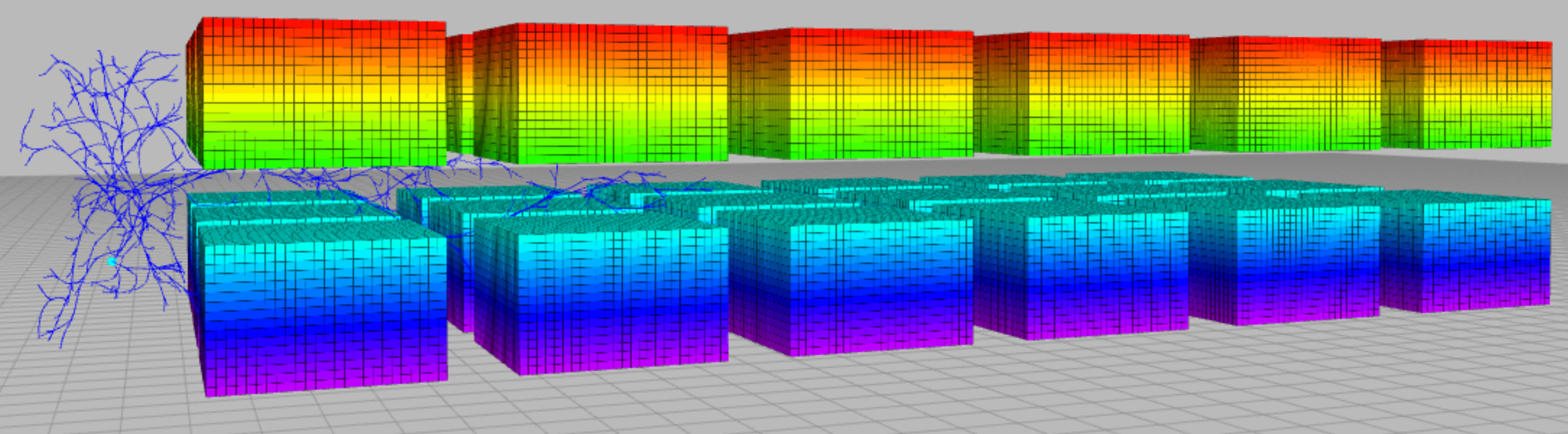}\label{fig:pcc_luders_planning}}
            \\
            \subfloat[This work]{\includegraphics[width=0.95\columnwidth, clip]{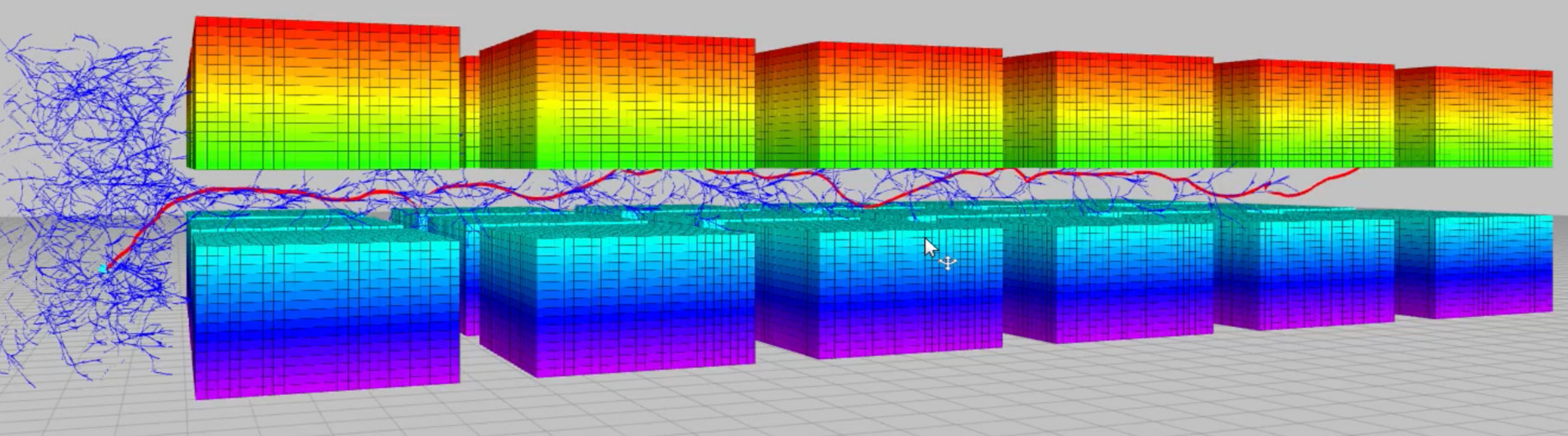}\label{fig:pcc_ours_planning}}
            \caption{Effect of conservatism in the workspace. Where the over conservatism of chance constraints formulations~\cite{blackmore2011chance,luders2013robust} impede finding a solution, the proposed probabilistic collision checking with a fixed conservatism $\alpha$ succeeds on finding a trajectory which transverses the environment with a total of $36$ obstacles.}
            \label{fig:pcc_planning}
        \end{figure}

        % ============
        % closing
        % ============
        % can support this discussion with images of motion planning such as in page 19 of http://acl.mit.edu/papers/Luders10_GNC.pdf 
        % file:///home/ericpairet/Downloads/Karaman_Robust%20Sampling-based.pdf
        \textbf{Suitability discussion:}
        robotic systems operating in uncrowded environments, i.e. very few obstacles sparsely distributed in the space, might find chance constraints to be a suitable alternative. However, the accuracy and efficiency of such approaches scales poorly as the complexity of the motion planning problem increases, i.e. more crowded environments or higher uncertainties. As it can be observed in \mbox{\fref{fig:pcc_blakcmore_planning}-\ref{fig:pcc_luders_planning}}, this behaviour endangers the ability of a planner to find a trajectory through tight apertures or narrow passages, even if one exists. If an alternative route towards the goal exists, the resulting solution will be larger than those trajectories found with less conservative approaches. Moreover, chance constraints require the representation of the environment to be a set of linear constraints, which can be prohibitively expensive to compute online, specially in applications where the environment is incrementally discovered.
        
        In contrast, our approach trades a constant conservatism $\alpha$ in favour of accuracy and performance. This allows to deal with crowded environments efficiently while providing higher accuracy than chance constraints methods. Therefore, as depicted in \fref{fig:pcc_ours_planning}, our probabilistic collision checking method enables a planner to find a solution through the tight corridors where chance constraints methods are over conservatist. However, our method involves higher computation times for highly uncertain states. This limitation might be relevant for systems with unbounded uncertainty, but most robotic systems are endowed with state estimation algorithms which keep the state uncertainty bounded over time. Alternatively, the parameter $\alpha$ can be adjusted to reduce the computation time while still guaranteeing safeness.
        
        On the whole, the presented probabilistic collision checking approach proves to be a suitable strategy for a wide range of motion planning problems under uncertainty, even for those where chance constraints struggle at finding a solution. Moreover, our method is suitable for applications building a representation of the environment online, given that those usually exploit the efficient encoding of occupancy grid maps.
        
    % ===============================
    % ===============================
    % ===============================
    \subsection{Start-to-goal Queries in Undiscovered 3D Environments \label{sec:evaluation_current_framework}}
        \begin{figure}[b]
            \centering
            \subfloat[Perspective view]{\includegraphics[width=0.95\columnwidth, clip]{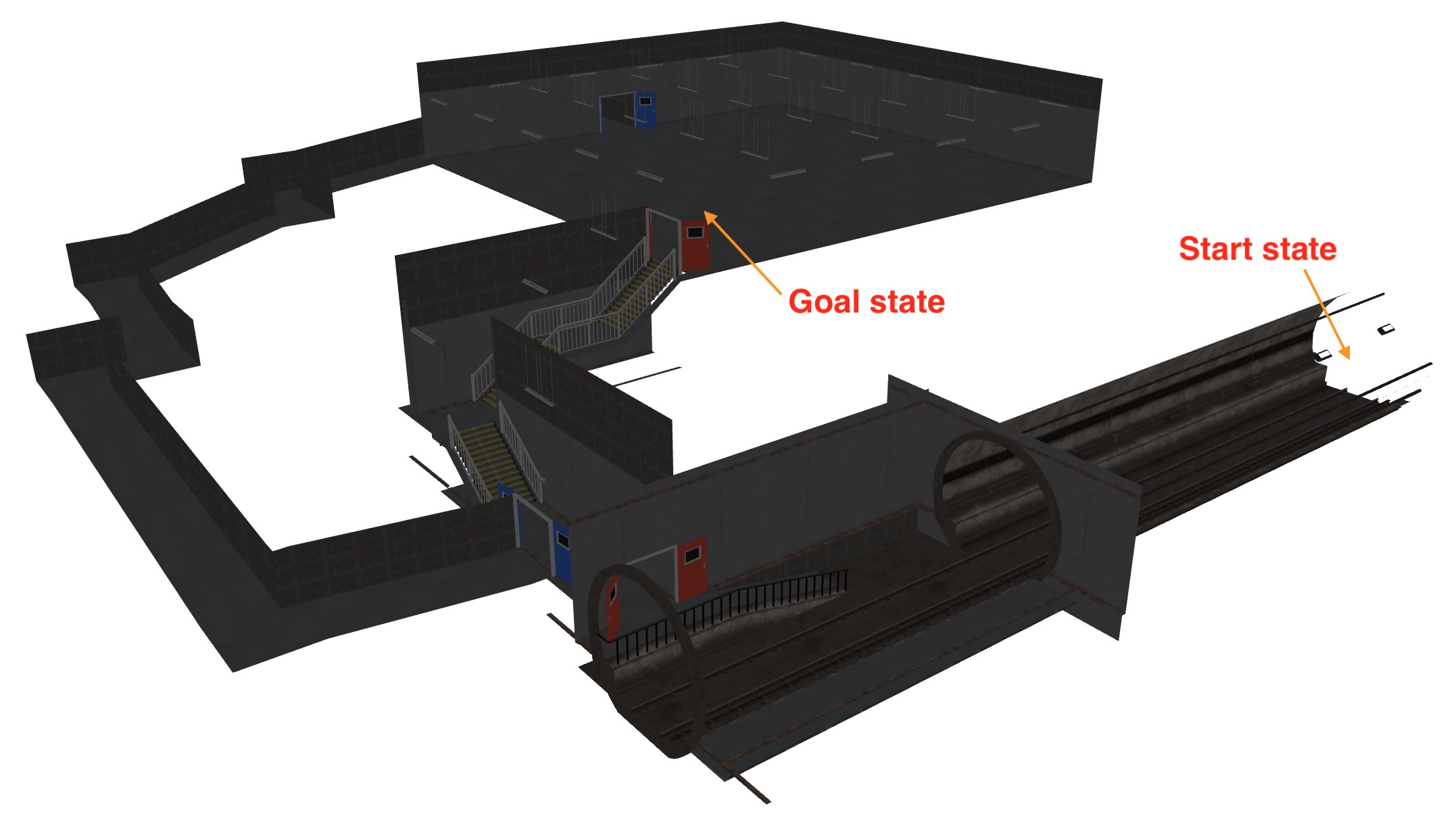}\label{fig:dc_overview}} \\
            \subfloat[40 metres long tunnel]{\includegraphics[width=0.48\columnwidth, clip]{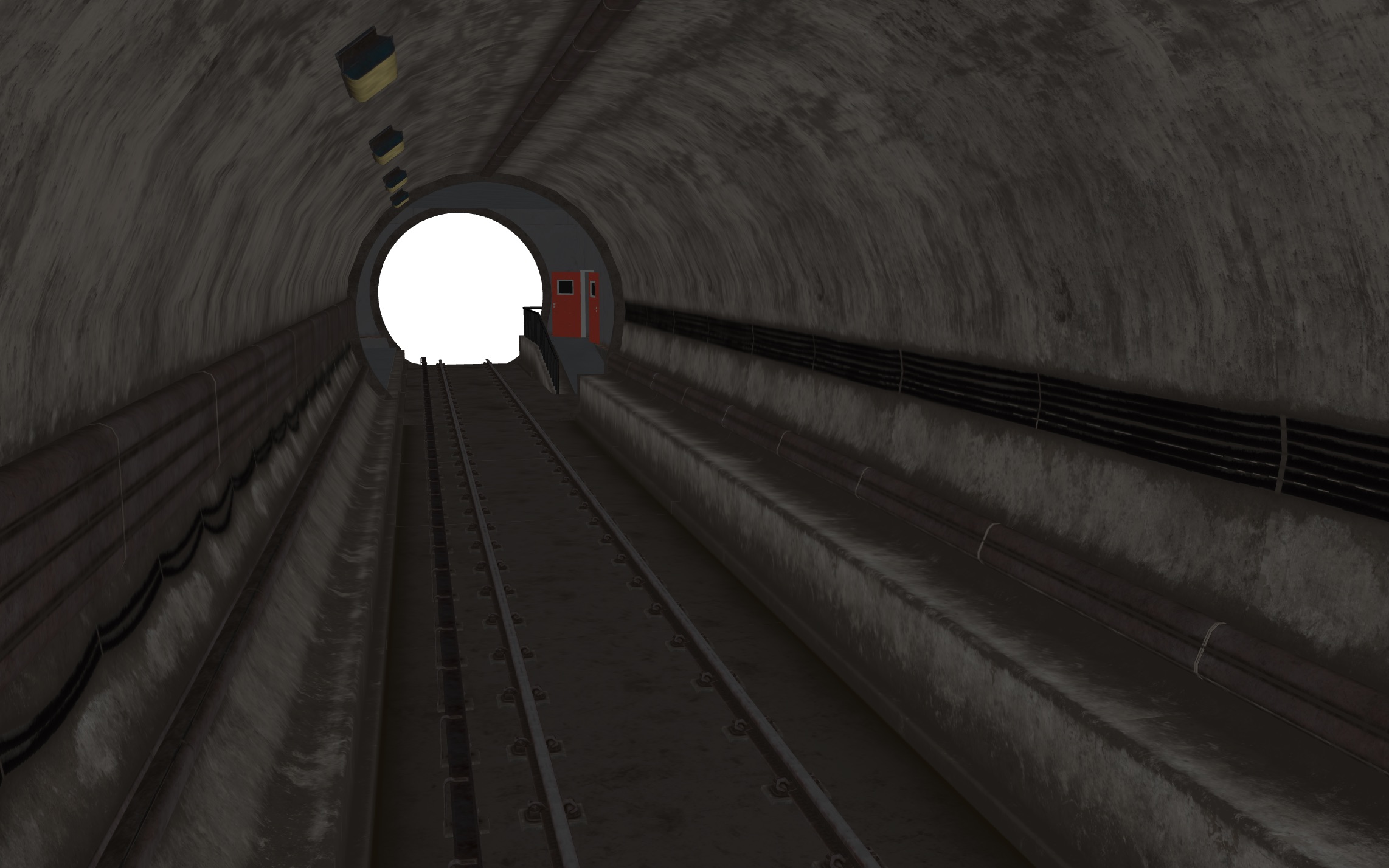}\label{fig:dc_tunnel}} \;
            \subfloat[Entrance to narrow stairwell]{\includegraphics[width=0.48\columnwidth, clip]{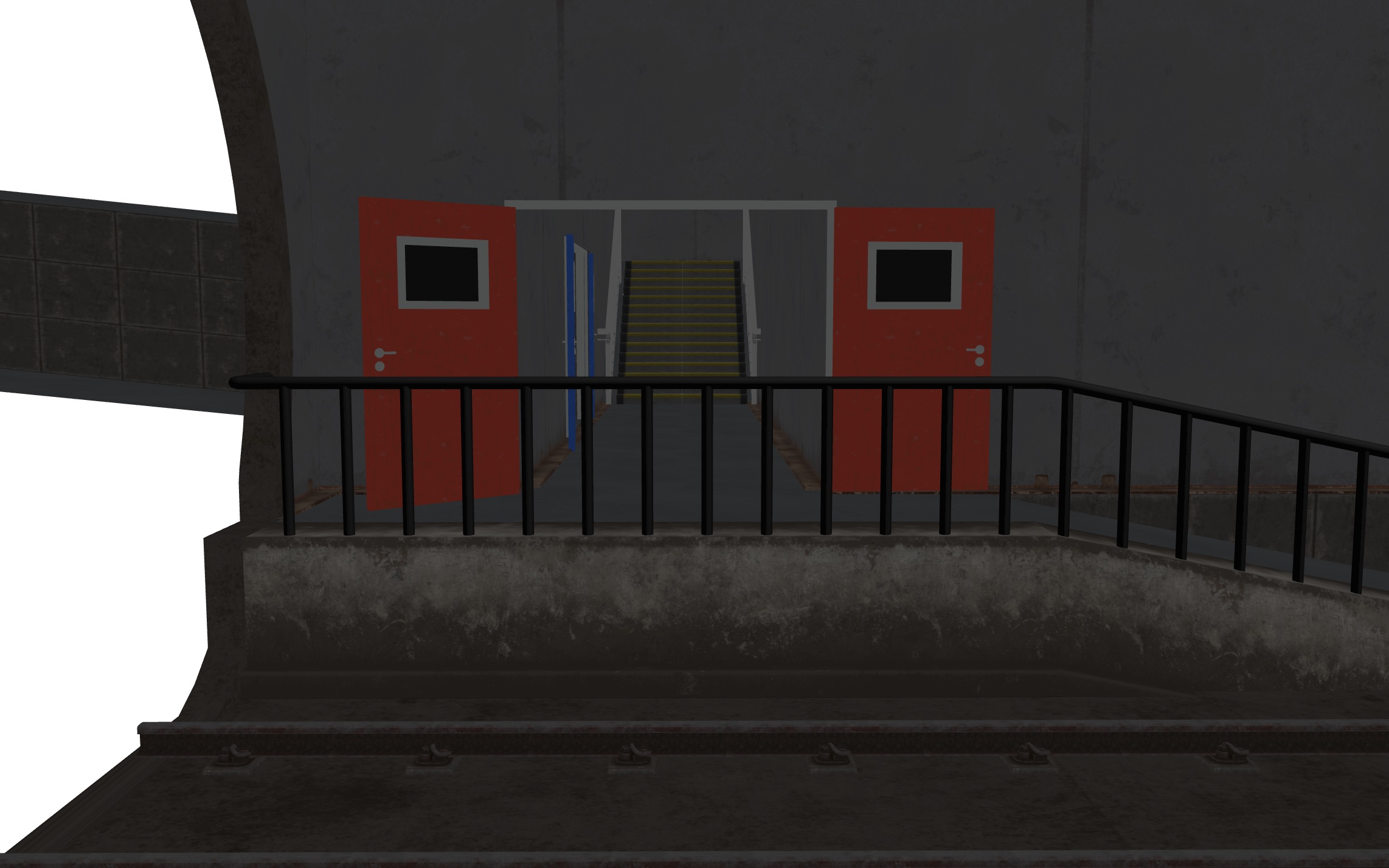}\label{fig:dc_narrow}} \\
            \caption{Urban Stairwell scenario of the DARPA Subterranean Challenge 2019. (a)~Start-to-goal query which requires traversing (b)~a 40 metres long tunnel and (c)~a narrow 25 metres long stairwell. Planning through the stairwell is particularly challenging due to the accumulated localisation uncertainty.}
            \label{fig:dc_scenario}
        \end{figure}
        
        \begin{figure*}[t]
            \centering
            \subfloat[]{\includegraphics[width=4.1cm, clip]{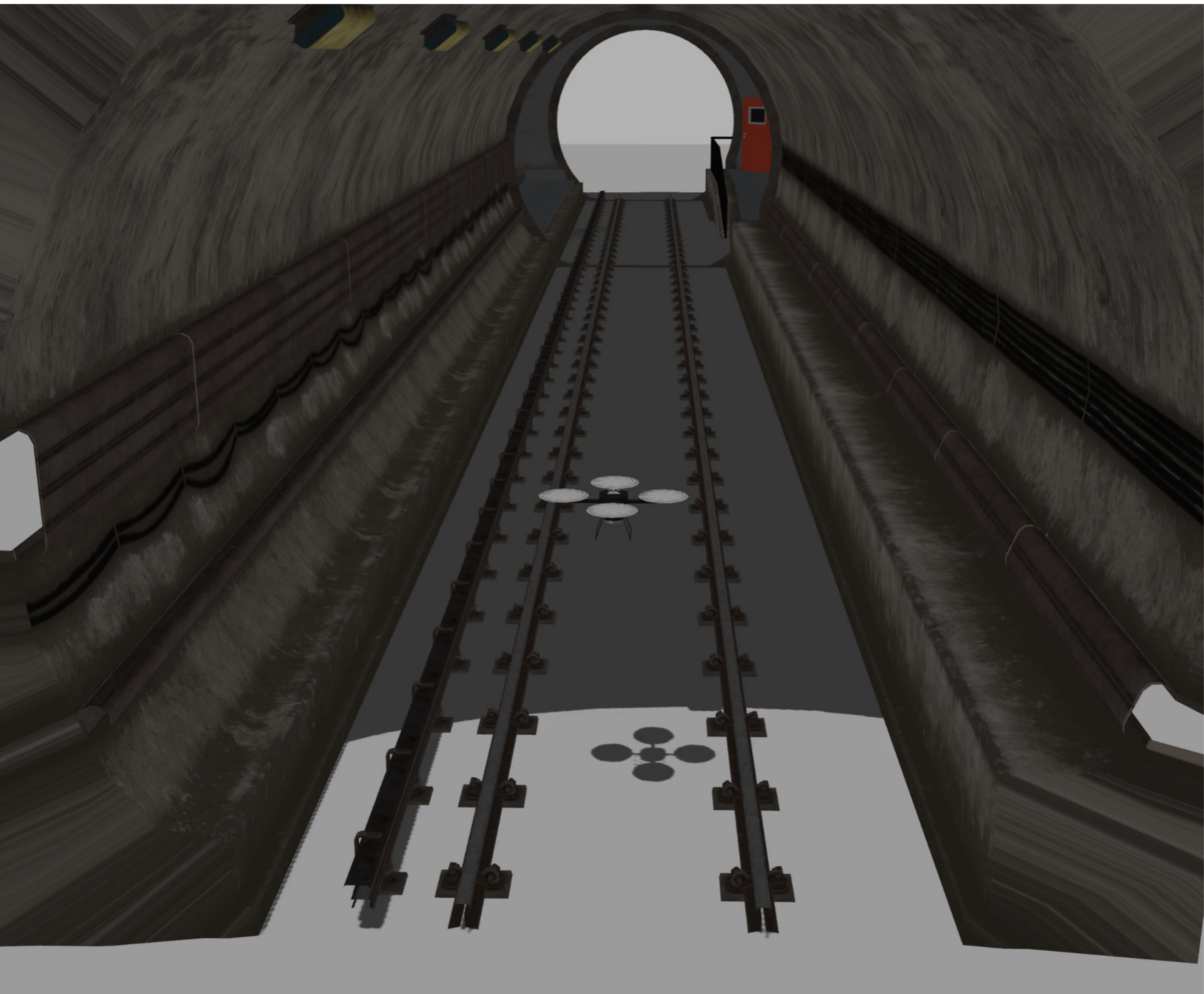}\label{fig:dc_snapshot_1}} \;
            \subfloat[]{\includegraphics[width=4.1cm, clip]{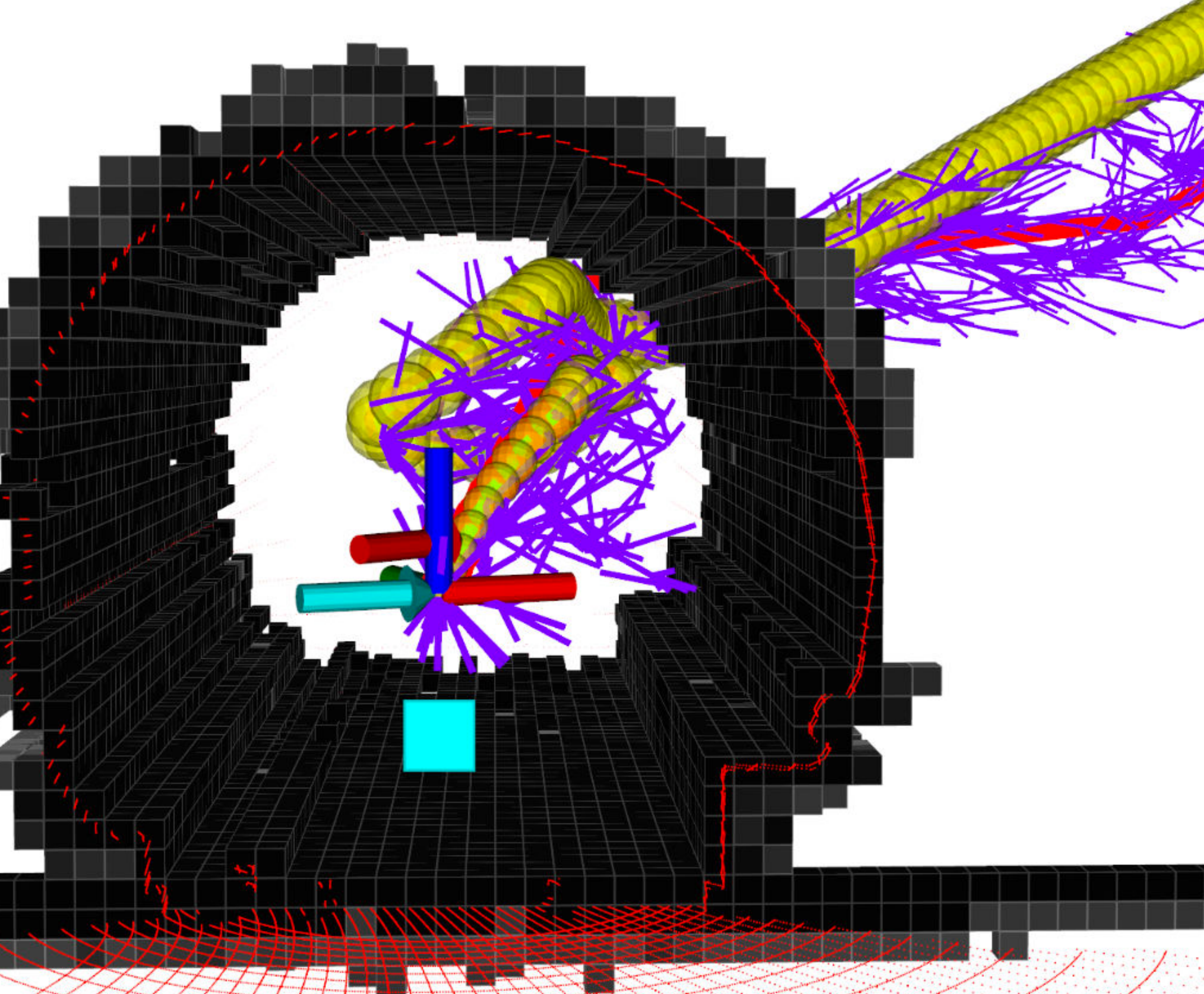}\label{fig:dc_snapshot_2}} \;
            \subfloat[]{\includegraphics[width=4.1cm, clip]{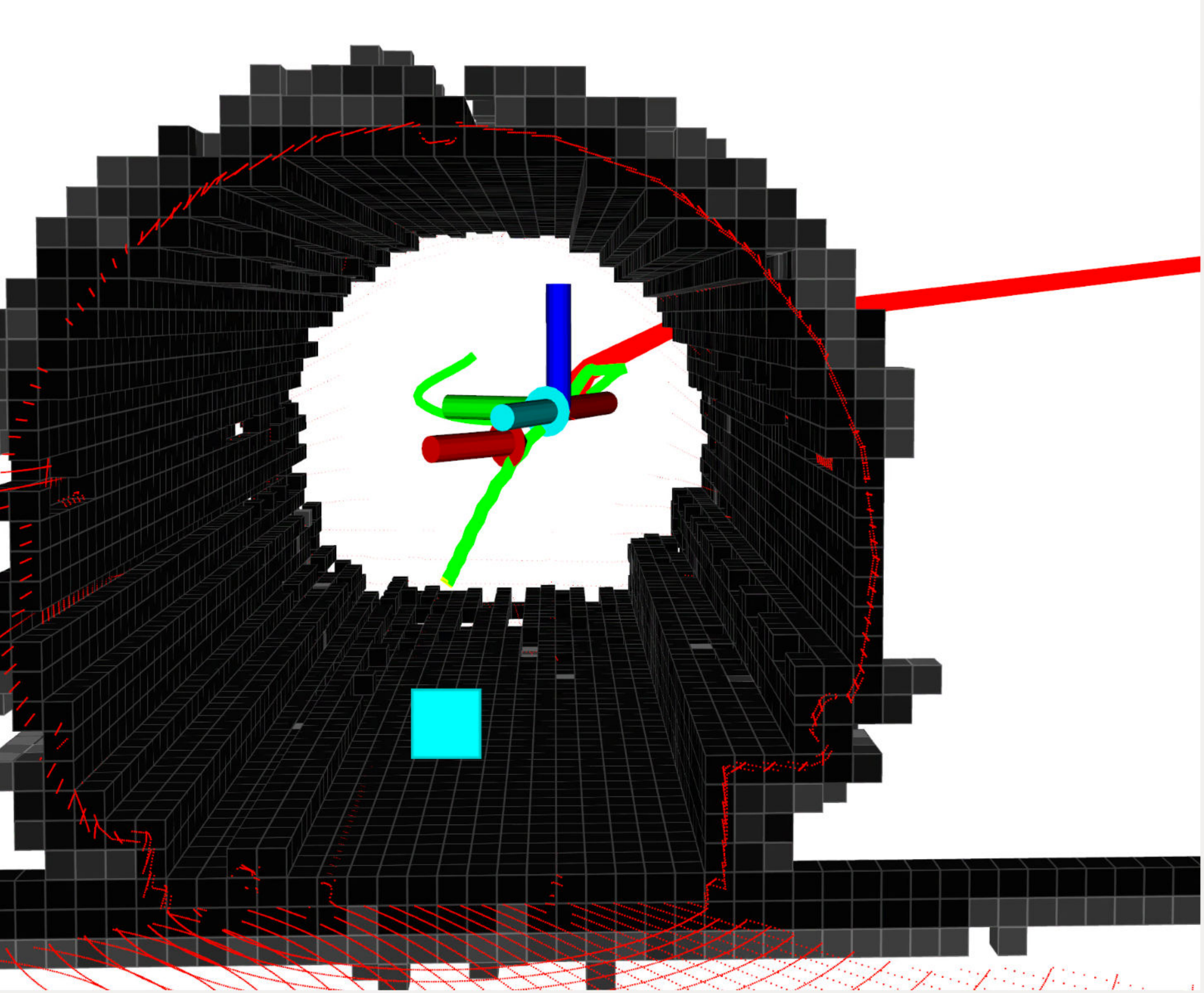}\label{fig:dc_snapshot_3}} \;
            \subfloat[]{\includegraphics[width=4.1cm, clip]{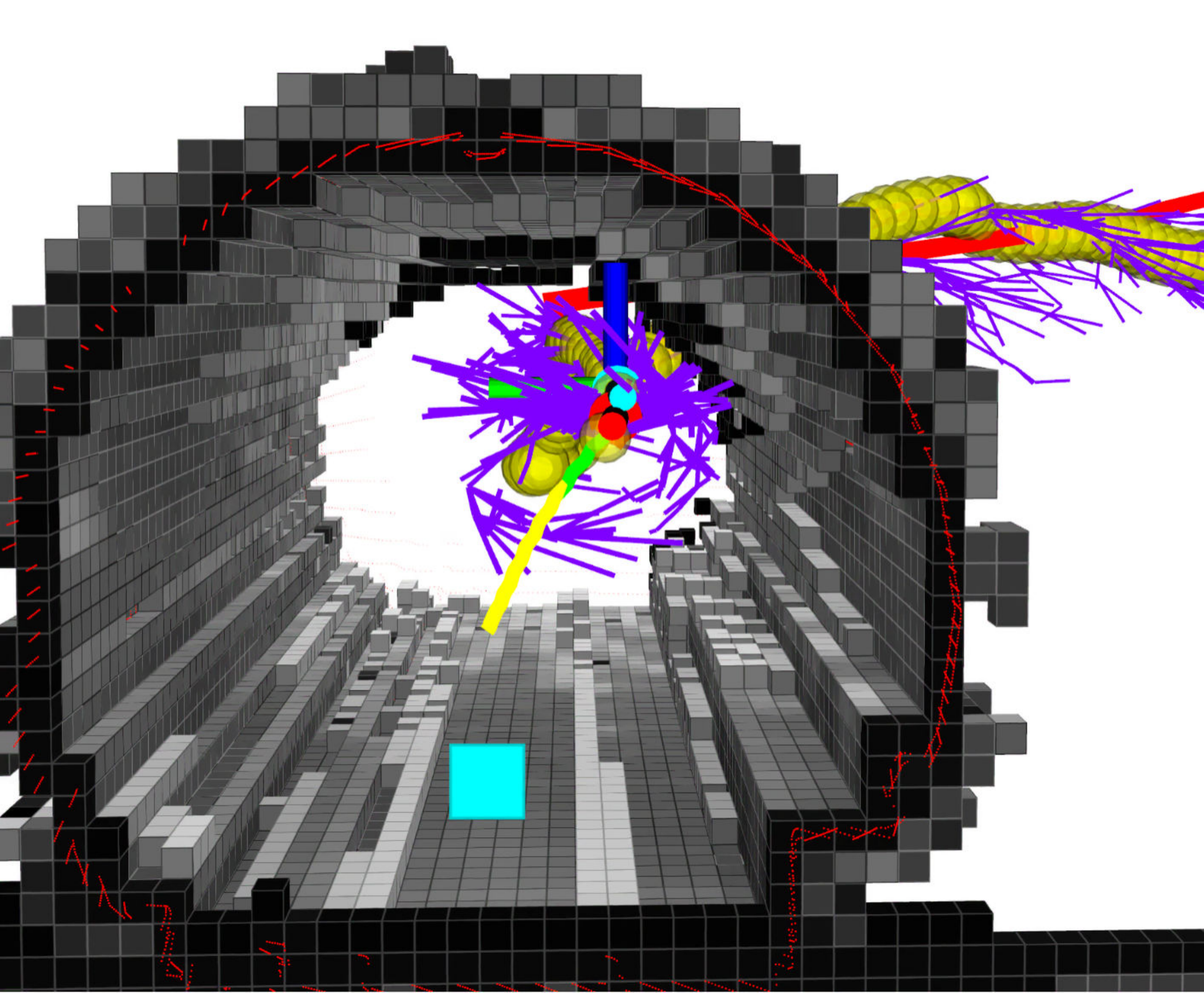}\label{fig:dc_snapshot_4}} \\
            
            \subfloat[]{\includegraphics[width=4.1cm, clip]{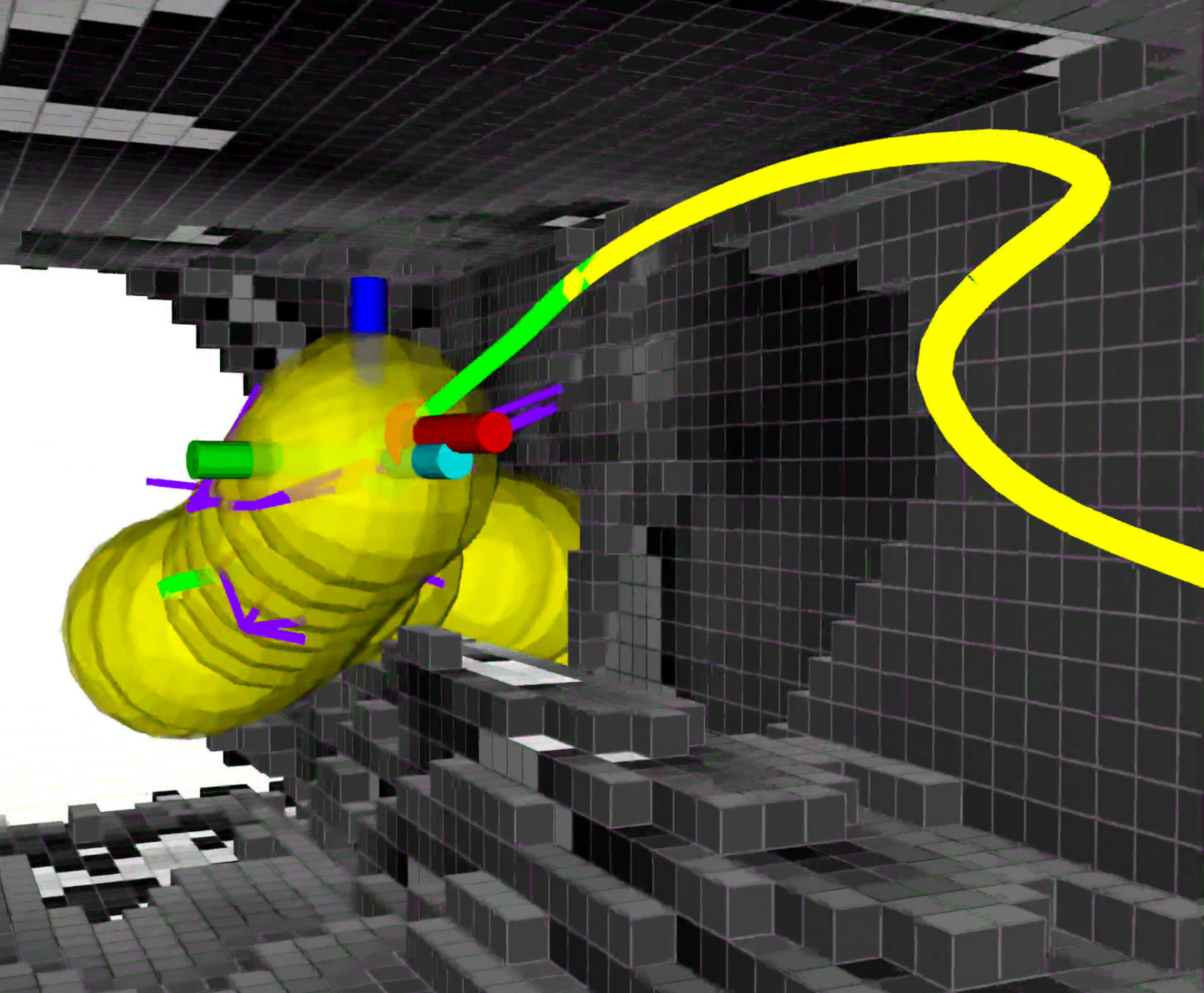}\label{fig:dc_snapshot_5}} \;
            \subfloat[]{\includegraphics[width=4.1cm, clip]{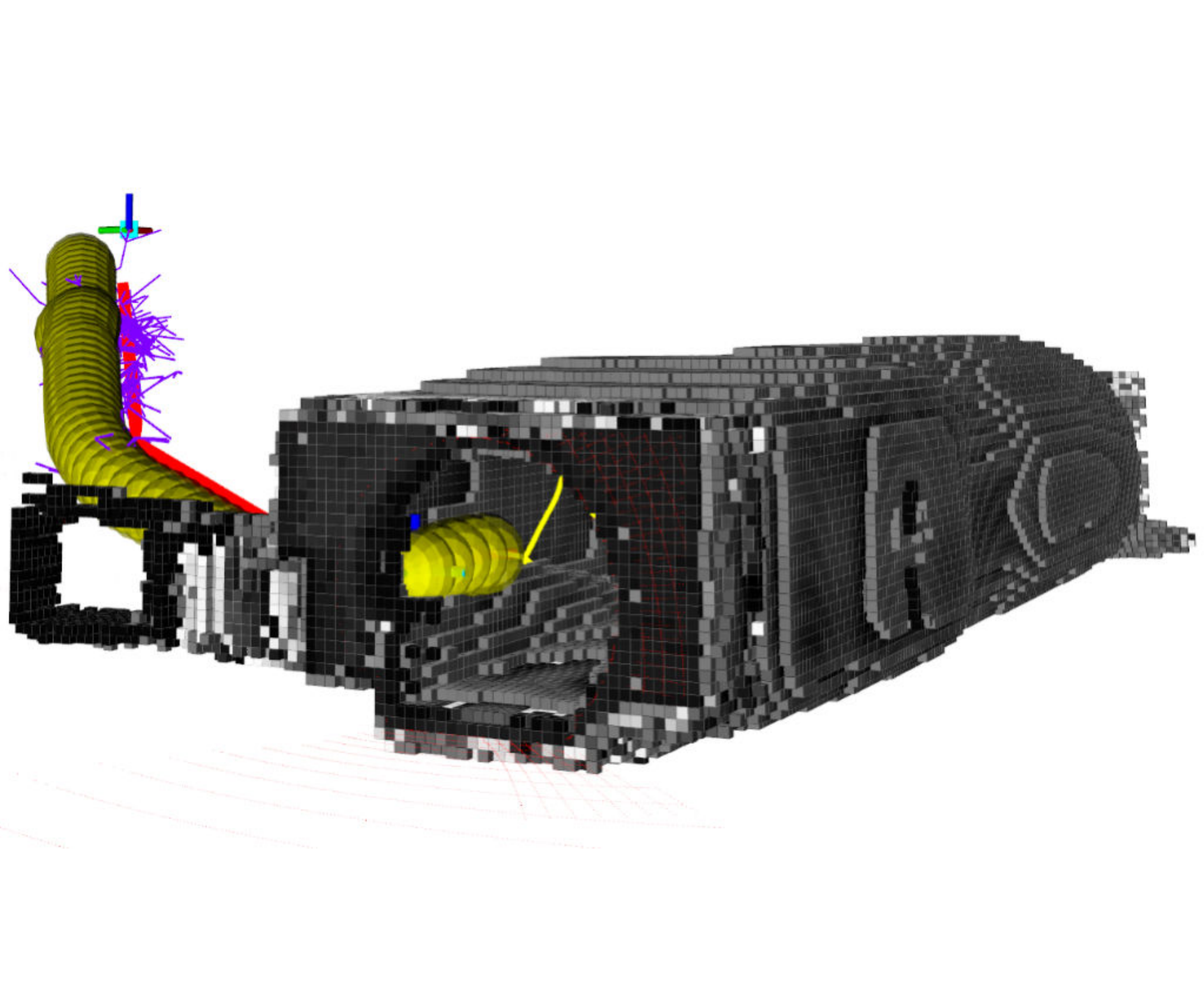}\label{fig:dc_snapshot_6}} \;
            \subfloat[]{\includegraphics[width=4.1cm, clip]{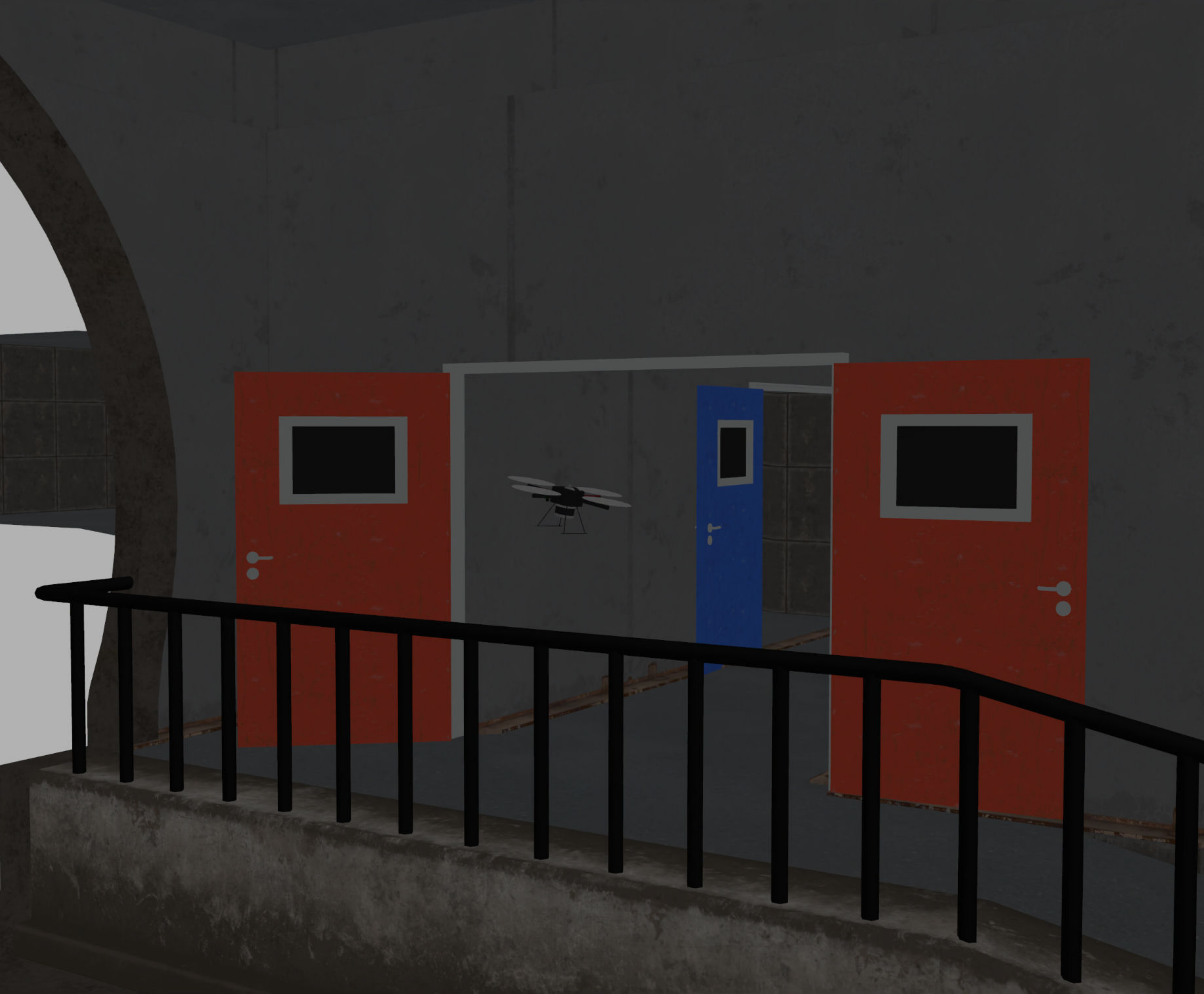}\label{fig:dc_snapshot_7}} \;
            \subfloat[]{\includegraphics[width=4.1cm, clip]{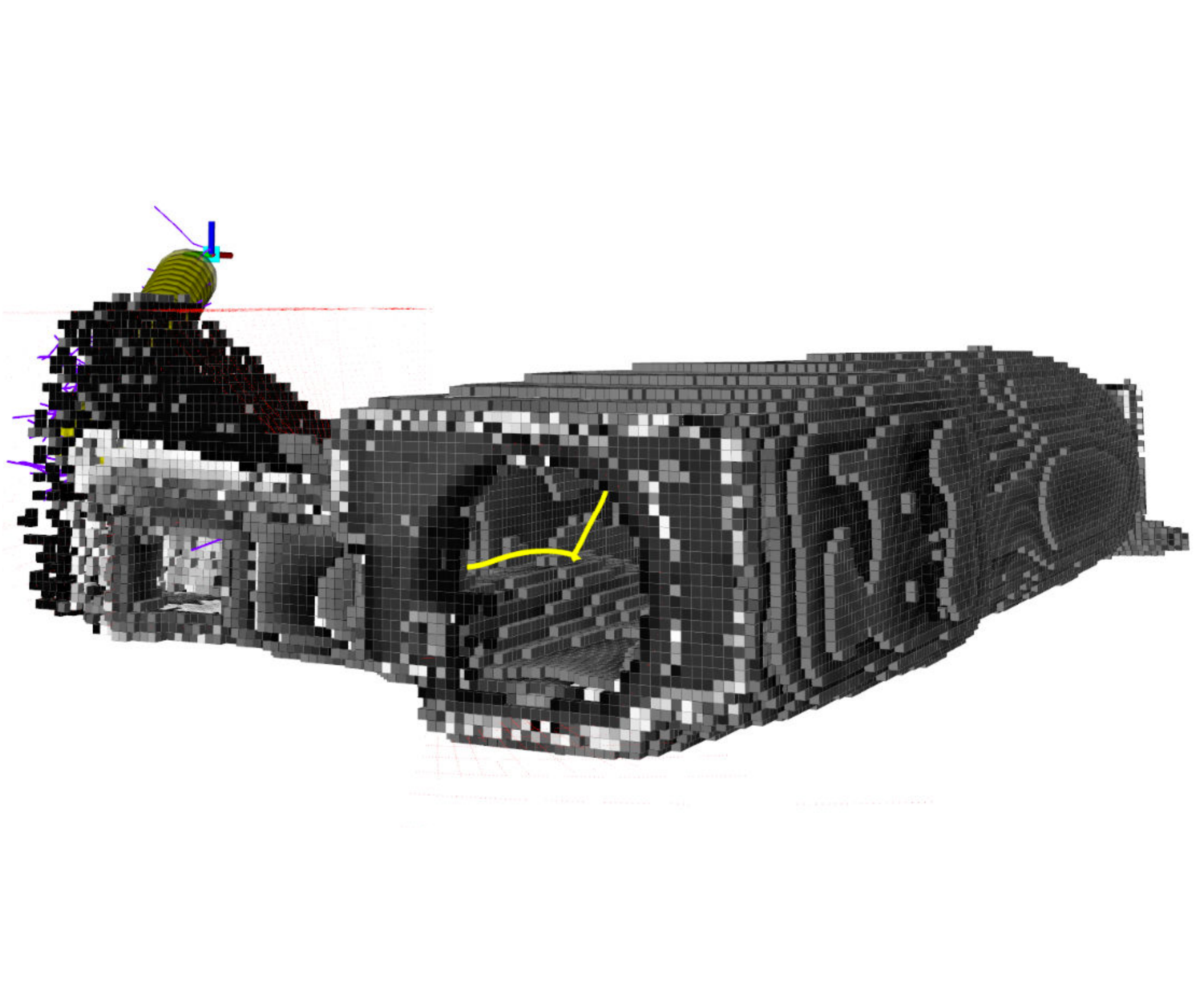}\label{fig:dc_snapshot_8}} \\
            
            \subfloat[]{\includegraphics[width=4.1cm, clip]{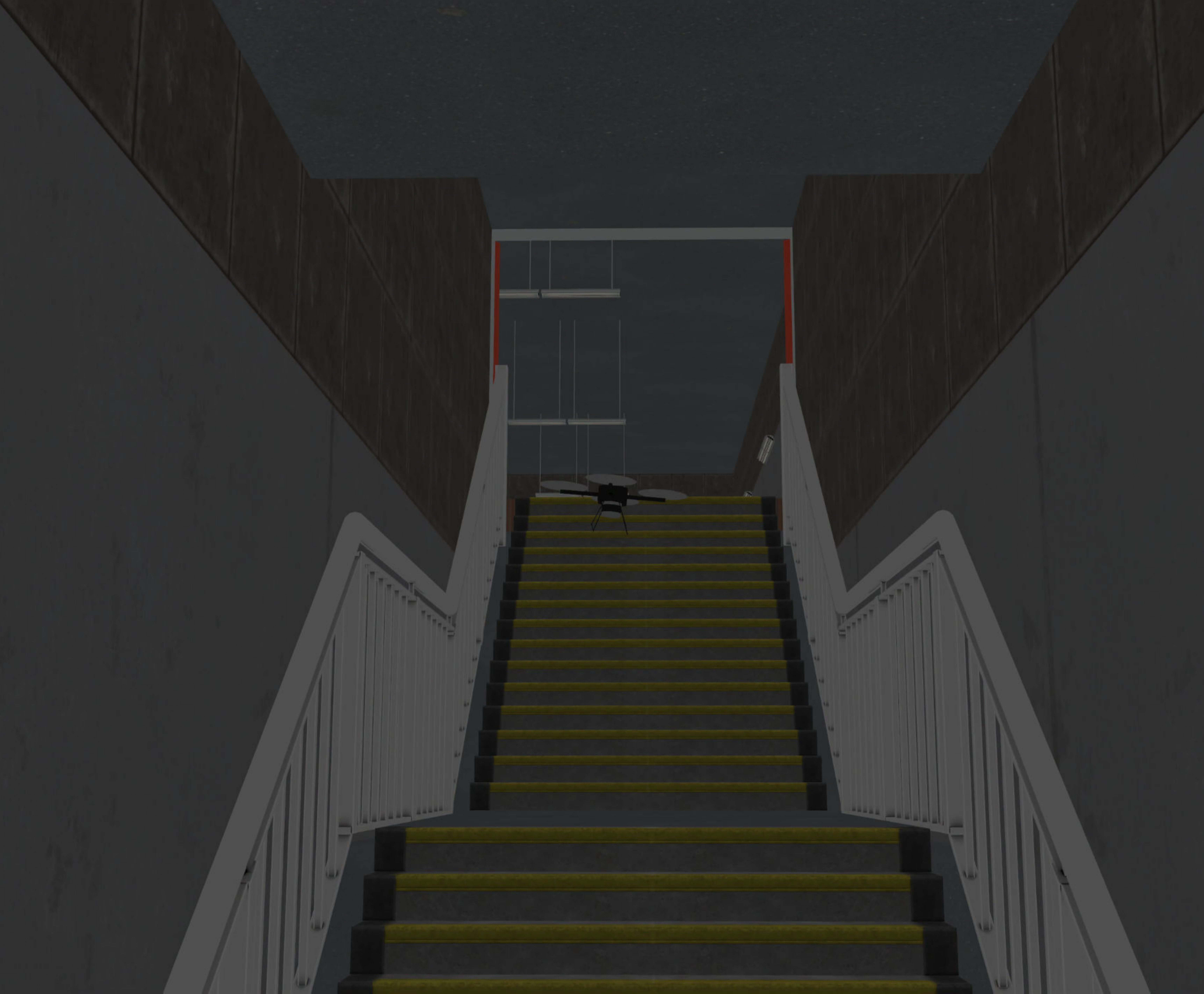}\label{fig:dc_snapshot_9}} \;
            \subfloat[]{\includegraphics[width=4.1cm, clip]{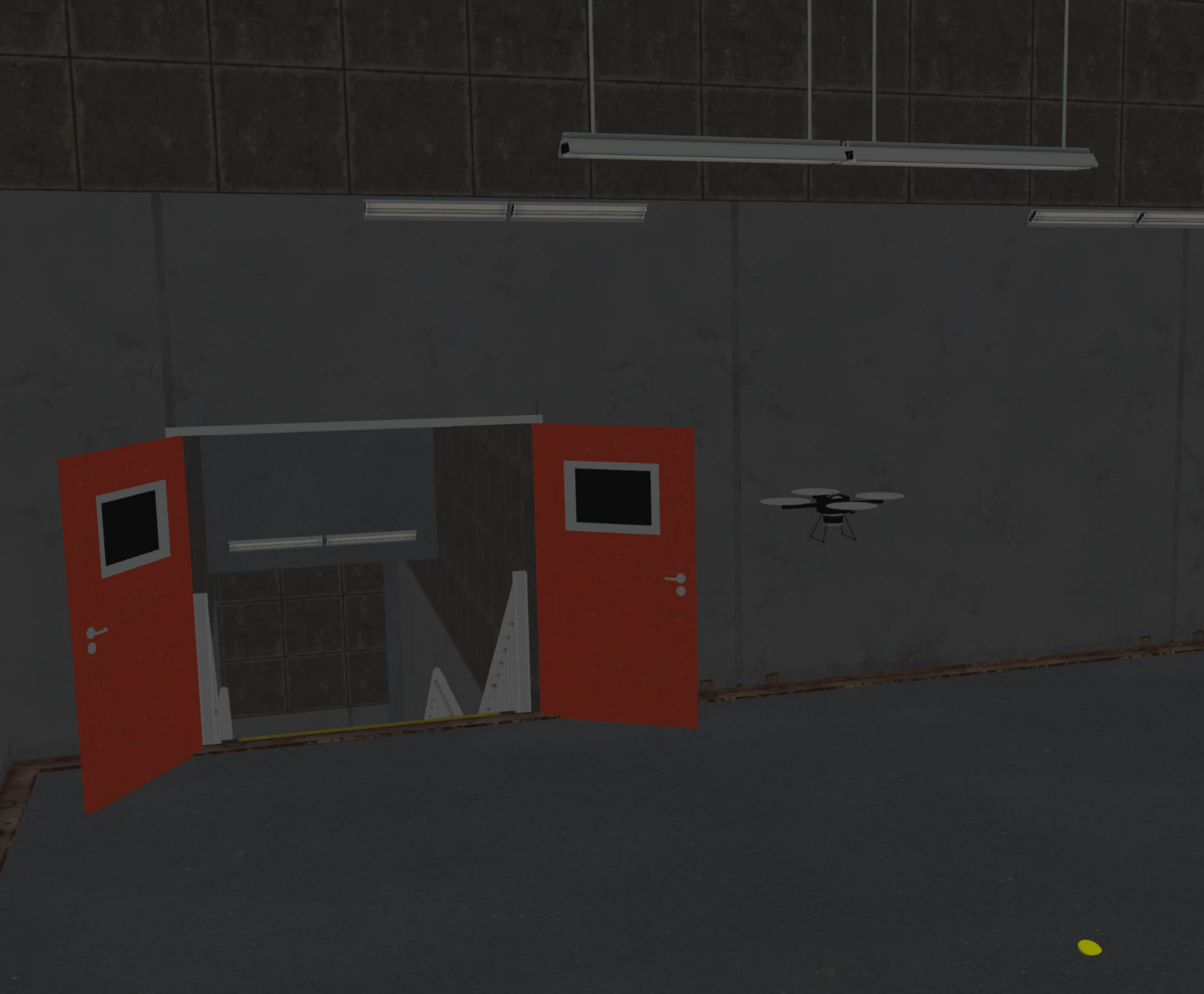}\label{fig:dc_snapshot_10}} \;
            \subfloat[]{\includegraphics[width=4.1cm, clip]{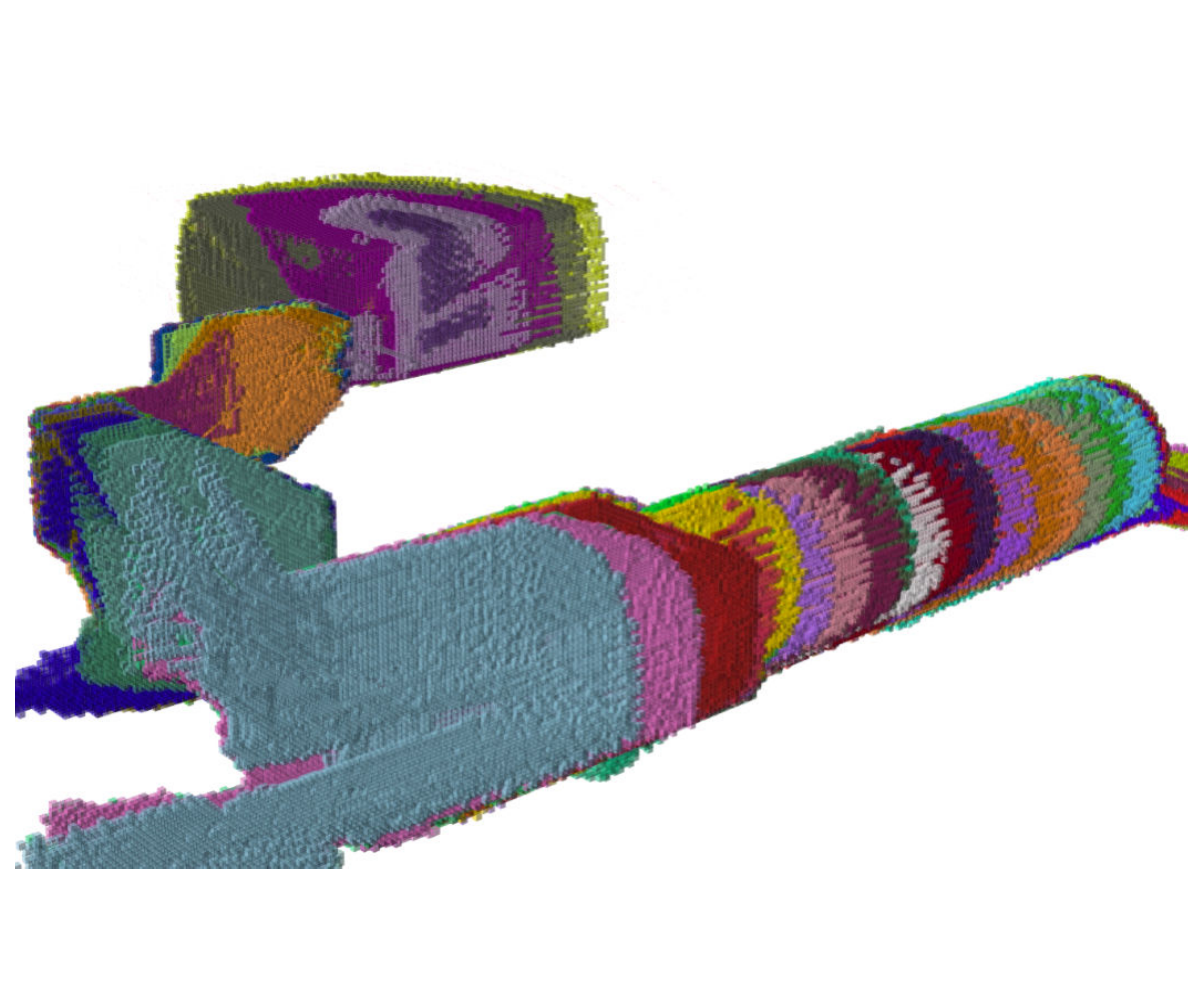}\label{fig:dc_snapshot_11}} \;
            \subfloat[]{\includegraphics[width=4.1cm, clip]{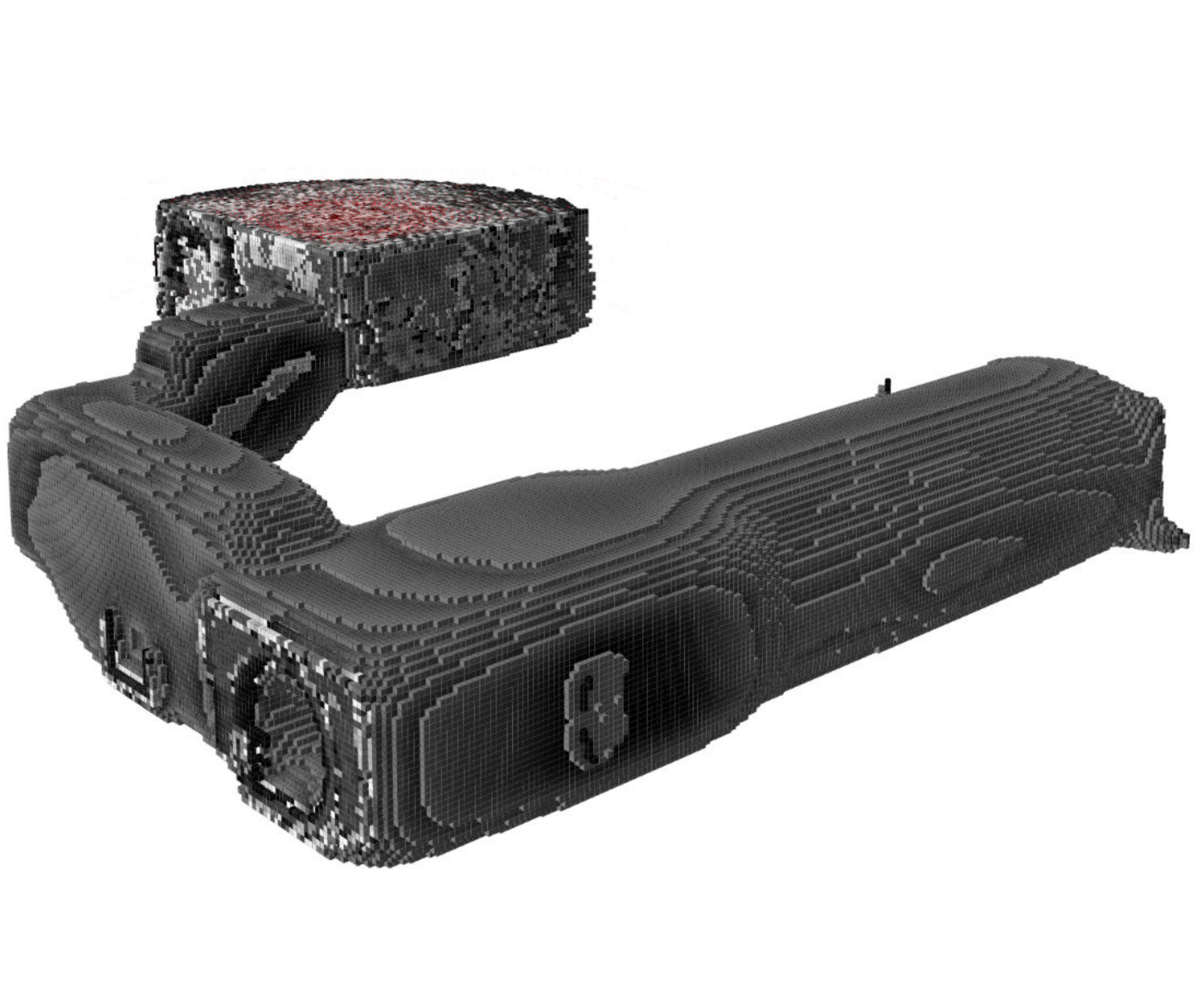}\label{fig:dc_snapshot_12}}
            \caption{Online mapping and planning through the Urban Stairwell scenario of the DARPA Subterranean Challenge 2019. (a)~Initial state of the quadrotor and (b)~first mapping and planning iteration: geometric path (red), kinodynamic tree satisfying the probabilistic safety guarantees $p_\safe=0.95$ (magenta), and resulting trajectory (green) with the associated uncertainty propagation (yellow). (c)~When the previous trajectory is partially invalidated due to the incremental knowledge of the surroundings, the framework finds a new trajectory towards the goal. Note that the previously observed patches of the environment become more uncertain (greyish areas) as the robot moves. (e-f)~The entrance to the narrow stairwell is fully mapped and the framework successfully plans through it despite the considerable accumulated uncertainty. (g-i)~As the robot moves into the stairwell, the framework continues iterating over the mapping-planning process to ensure save navigation until (j)~reaching the goal region. (k)~Incremental set of local maps composing the discovered environment during the mission, and (l)~corresponding cumulative map $F_\X^{\planningframe}$ (only showing those voxels $P(\voxel) > 0.4$ for visualisation purposes).}
            \label{fig:dc_snapshots}
        \end{figure*}
        
        The proposed framework as a whole has been deployed on a simulated quadrotor \acf{UAV}~\cite{2012simpar_meyer} equipped with a 3D \ac{LIDAR}. The considered environment is the Urban Stairwell scenario of the DARPA Subterranean Challenge 2019. This scenario is challenging due to its extensive workspace of ${40.50 \times 50.04 \times 13.69}$ metres and all narrow passages that must be traversed to accomplish the requested start-to-goal motion planning query. \fref{fig:dc_scenario} illustrates the Urban Stairwell scenario altogether with the defined start-to-goal query. For this experiments, the quadrotor's dynamics are approximated to those of a fixed-wing plane as described in \xref{sec:km_fixed_wing} and the surroundings are mapped online from the sensor's data at a resolution of $0.2$ metres. The remaining parameters of the mapping module are as those in the experiments reported in \sref{sec:evaluation_precedent_framework}. During the mission, no localisation updates are considered to test the planner in the most adversarial conditions, i.e., large environmental and localisation uncertainties. The required probabilistic safety guarantees are $p_\safe=0.95$ and the planning time is $\planningtime=1.5$ seconds, distributed as $\planningtimescout=0.3$ seconds and $\planningtimetough=1.2$ seconds. \fref{fig:dc_snapshots} depicts some snapshots\footnote{A complete trial through the Urban Stairwell scenario of the DARPA Subterranean Challenge 2019 can be seen in: \newline %\url{www.dropbox.com/s/y0jpx2a5azi41v2/ijrr.mp4?dl=0}.}. 
        %\url{www.dropbox.com/s/vuqpyo5dy5tzdmw/IJRR20_v2_rs.mp4?dl=0}.}
        \url{https://youtu.be/I5X_QFKDpeI}.}
        of the online mapping and planning procedure in the Urban Stairwell scenario of the DARPA Subterranean Challenge 2019. The proposed framework allows for probabilistically safe autonomous navigation in such a hostile and unknown environment. 
        
        The mesh of the Urban Stairwell scenario is composed of a total of $108{,}512$ faces. Although these faces could be potentially approximated online from the sensor's data and used as linear constraints in~\cite{blackmore2011chance,luders2013robust}, it would imply checking for collision against $30$ times more linear constraints than those considered in \sref{sec:evaluation_pcc}, for which chance constraints methods already showed poor performance due to their over conservatism. Instead, our framework is able to efficiently deal with these complex scenarios, as demonstrated with the experiments conducted in the Urban Stairwell scenario of the DARPA Subterranean Challenge 2019.
    \section{Conclusion} \label{sec:conclusion}
    This paper has presented a novel end-to-end framework, which probabilistically guarantees the robot's safety when navigating in unexplored environments. The proposed approach is twofold: (i)~incrementally maps the vehicle's surroundings to build an uncertain representation of the environment, and (ii)~plans feasible trajectories (according to the robot's kinodynamic constraints) with probabilistic safety guarantees (according to the uncertainties in the vehicle's localisation, motion and mapping). Our proposed approach includes a multi-layered planning strategy which enables for faster exploration of the high-dimensional belief space, while preserving asymptotically optimal and completeness guarantees, and an efficient evaluation and tighter bound on the computation of the probability of collision than other uncertainty-aware planners in the literature. Overall, the framework is capable to deal with high-dimensional problems online while being suitable for systems with limited on-board computation power. Experimentation conducted in simulation shows some of the theoretical qualities of this work. Additionally, simulated and real-world trials on an \ac{AUV} and a quadrotor \ac{UAV} demonstrated the suitability of the framework to guarantee the robot's safety while navigating in unexplored environments and dealing with real-robots constraints.
    
	The framework is not restricted to the presented experimental evaluation nor a specific platform. Any other mobile robot, either terrestrial, maritime or aerial system 
	%with a linearisable equation of motion 
	can benefit from this work. 
	%The proposed framework assumes linearisable systems because its behaviour can be numerically quantified, what is an essential aspect to ensure the safety guarantees pursued in this work. Alternatively, the unscented transform could be used to approximate the state of highly non-linear systems to Gaussian distributions. However, as there are no proofs for the estimation accuracy of the unscented transform, the safety constraints cannot be guaranteed anymore.
	The modularity of the proposed framework allows for multiple extensions and variations. Foremost, although the experimental evaluation of the proposed framework has been conducted considering the worst-case scenario of an open-loop navigation without uncertainty update, the framework can bear with periodic navigation updates as described in \sref{sec:framework_root}. An interesting possible feature that could be added to the framework is the use of the truncation trick, i.e. to uniquely propagate the posterior of the estimation which is in no collision. However, truncating the system's belief involves approximating the posterior to a Gaussian distribution. Another possible extension is leveraging the multi-resolution encoding of Octomaps to check the compliance of the safety guarantee at different resolutions. Formulating this process as a multi-resolution kernel checking could speed up computations even further. Finally, the conducted experimentation pointed out that automatically adjusting the replanning period might be beneficial, as well as studying more intelligent methods to leverage from the lead path or even prior solutions.
	
	% include in discussion:
	% - contingency plan
	% - what to do to not get trapped/or in an inevitable collision state (https://arxiv.org/pdf/1804.05804;Safe)
	% - sensory limitations

    % ===============================
    % ===============================
    % ===============================
    %\subsection{Contingency Plan \label{sec:formulation_contingency}}
        % related
        % http://www.i6.in.tum.de/Main/Publications/Althoff2015b.pdf
    
        % the framework must guarantee that the vehicle does not get to a dead-end. To this aim, avoiding Inevitable Collision States is not sufficient. Instead, a safe state is such allowing for reachibility to a state of the executed path. Then, from this state on, the system can redo the previous path backwards.
        
        % robot should not attempt to achieve a state which has:
        %(1) not seen
        %(2) no contingency plan to a safe state
        %(3) that the contingency plan lies in unexplored space
        
        % NOTE: predicting the field of view does not make sense. In the worst case scenario, whatever is perceived my be occupied space, therefore safety cannot be guaranteed if the framework relies on detecting free space.
        
        % OPTION 1: include these constraints in the state validity checking
        % OPTION 2: only check for the short horizon of the resulting path, i.e. up to where the robot needs to go until next planning cylcle (B^\prime)
        
        % check reachable sets in \cite{majumdar2017funnel}: they mention works which aim at verification and emergency maneuver
        % http://www.i6.in.tum.de/Main/Publications/Althoff2015b.pdf

    \begin{acks}
    The authors are grateful to Michael Mistry and Paola Ard\'on for all support and helpful discussions about this work.
\end{acks}

\begin{dci}
    The authors declare that there is no conflict of interest.
\end{dci}

\begin{funding}
    The authors disclosed receipt of the following financial support for the research, authorship, and/or publication of this article:
    this research was partially supported by the School of Engineering and Physical Sciences (EPS) at Heriot-Watt University, as part of the CDT in Robotics and Autonomous Systems at Heriot-Watt University and The University of Edinburgh. Additionally, this research had been partially supported by the Scottish Informatics and Computer Science Alliance (SICSA), ORCA Hub EPSRC (EP/R026173/1) and consortium partners. Experiments in \sref{sec:evaluation_precedent_framework} were conducted in the Computer Vision and Robotics Institute (VICOROB) at University of Girona with the support of the EXCELLABUST and ARCHROV projects under the Grant agreements H2020-TWINN-2015, CSA, ID: 691980 and DPI2014-57746-C3-3-R, respectively.
\end{funding}

%\begin{sm}
%    \textcolor{red}{TODO}
%\end{sm}

    \appendix
    \section{Kinematic Models} \label{sec:kinematic_models}
    %https://www.researchgate.net/publication/236333580_A_Geometric_Approach_to_Dynamically_Feasible_Real-Time_Formation_Control
    %10D near-hover quadrotor: https://people.eecs.berkeley.edu/~jfisac/papers/FaSTrack.pdf
    %fixed-wing: https://static1.squarespace.com/static/56e4a24bc6fc082c7577a416/t/5718f7377da24fd2759818bd/1461253948078/Hugh_AER_1216_lecture.pdf
    %fixed-wing: https://link.springer.com/referenceworkentry/10.1007%2F978-90-481-9707-1_53

    % ===============================
    % ===============================
    % ===============================
    \subsection{Unicycle System \label{sec:km_auv_2d}}
        For the particular case of a torpedo-shaped \acf{AUV} that operates at a constant depth, i.e. in a \ac{2D} workspace ${\W = \real^2}$, with configuration space $\mathrm{SE}(2)$, the vehicle's motion model can be approximated by a (second-order) unicycle system:
        {\par
        \begin{figure}[H]
            \centering
            \begin{minipage}{0.22\textwidth}
                \centering
                \includegraphics[width=0.85\textwidth]{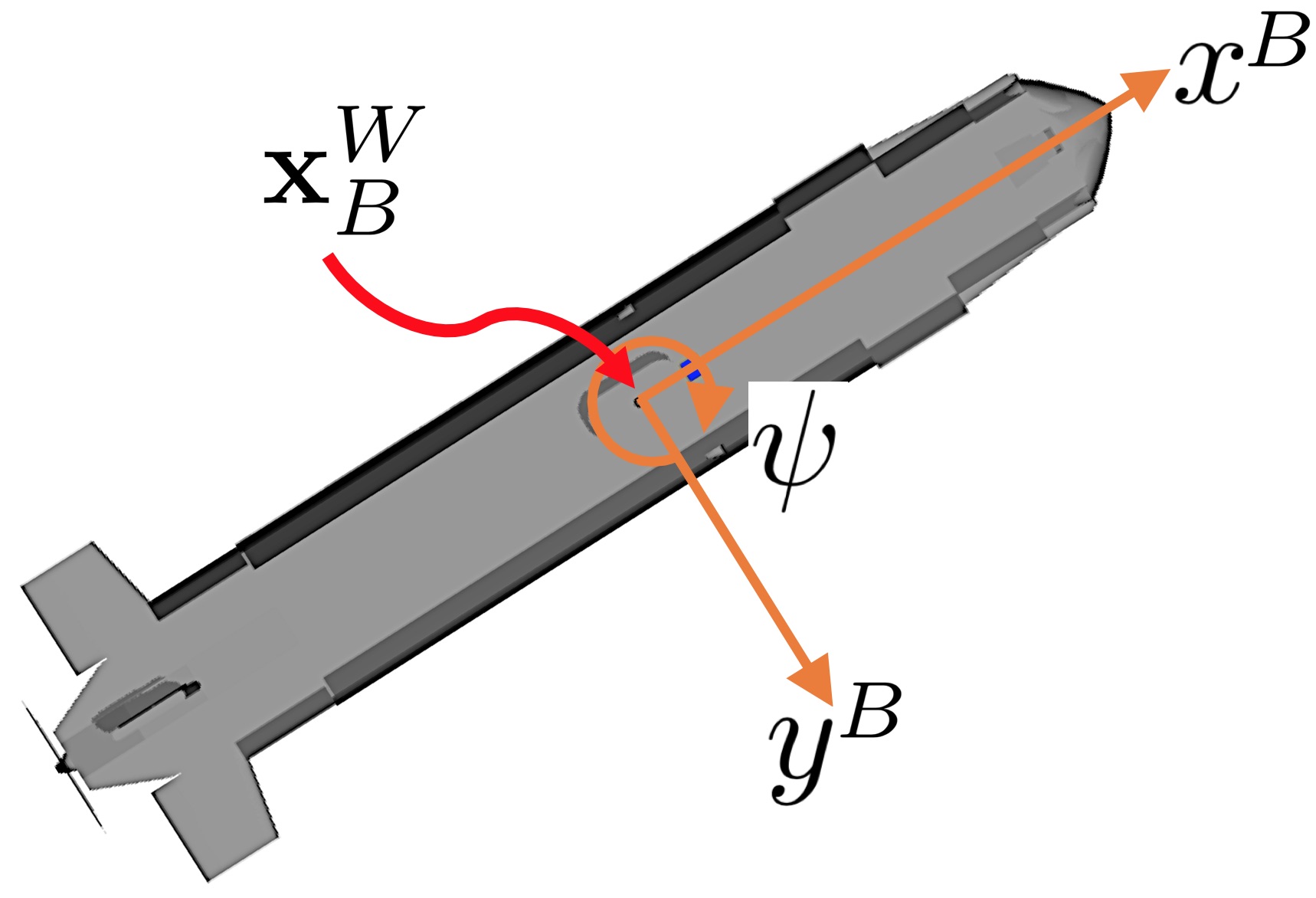}
            \end{minipage}% to prevent a space
            \begin{minipage}{0.22\textwidth}
                \begin{equation*}
                    \begin{aligned} 
                        \dot{x} &= v \; \cos(\yaw), \\
                        \dot{y} &= v \; \sin(\yaw), \\
                        \dot{\yaw} &= \yawvel, \\
                        \dot{v} &= a,
                    \end{aligned}
                \end{equation*}
                \vspace{0.5cm}
            \end{minipage}
        \end{figure}
        \par}
        \noindent where $x$ and $y$ correspond to the Cartesian coordinates of the system with respect to a predefined reference frame, $\yaw$ is the system's orientation around the $z$-axis, and $v$ is the vehicle's forward velocity, $\yawvel$ is the vehicle's turning rate, and $a$ is the acceleration. Thus, the system's state is defined as ${\vx = (x, \; y, \; \yaw, \; v)^T}$, and the system's control input is defined as ${\vu = (\yawvel, \; a)^T}$.
        
        % Although the model above approximates the \ac{AUV}'s behaviour such that it is computationally tractable, they do not consider any desired reference. 
        The model above approximates the \ac{AUV}'s behaviour, but in underwater environment it is subject to uncertain external forces, e.g., current.  To capture this uncertainty in the dynamics, the vehicle motion is modelled as a Gaussian process.
        % In coping with this limitation, 
        The system's motion model is first linearised by using a dynamic feedback linearisation controller as presented in~\cite{de2000stabilization}.
        This technique 
        % (i)~increases the order of the system's state, now denoted by $\vz$, and (ii)~
        (i)~transforms the state of the closed-loop system to ${\vz = (x,\; y,\; \dot{x},\; \dot{y})^T}$, and (ii)~
        applies a \ac{PD} controller on the model to drive the system towards a desired state~$\vr$. Then, the differences between the real system and the linearised closed-loop model can be approximated by a Gaussian distribution, and the closed-loop system can be represented as a Gaussian process as in~\mbox{\eqref{eq:generic_km}-\eqref{eq:generic_cov}} with states ${\vz \in \X = \real^{4}}$, and controls ${\vr \in \U = \X}$.  Thus, the system state $\vz_\timestamp$ is best described by its probability distribution in the belief space~$\B$, i.e. ${\belief_\timestamp = \N(\hat{\vz}_\timestamp, \; \cov_{\vz_\timestamp})}$. The evolution of the belief is then given by the independent propagation of its mean and covariance as:
	    \begin{align}
	        %\begin{split}
	           \label{eq:state_propagation}
	           \hat{\vz}_{\timestamp+1} &= A \hat{\vz}_\timestamp + B \vr_\timestamp, \\
	           \cov_{\vz_{\timestamp+1}} &= A \cov_{\vz_\timestamp} A^T + \cov_{\vw},
	        %\end{split}
	        \label{eq:cov_propagation}
	    \end{align}
	    where ${A \in \real^{4 \times 4}}$ and ${B \in \real^{4\times 4}}$ define the closed-loop linearised equation of motion with the \ac{PD} controller as in~\cite{de2000stabilization}, and $\cov_\vw$ is the covariance of the noise modelling the discrepancies between \eref{eq:state_propagation} and the real system behaviour.

    \subsection{Fixed-wing System \label{sec:km_fixed_wing}}
        %\erpaar{can (3) for the UAV be learnt from the error between the model below and the data from the simulator?}
        %\ml{yes, we can argue that! But, we should try to find a reference for it.}
    
        Although more complex models could be used to represent the motion capabilities of an \ac{AUV} or a quadrotor \acf{UAV} operating in a \ac{3D} workspace ${\W = \real^3}$ with configuration space $\mathrm{SE}(3)$, both vehicle's motion model can be approximated by a fixed-wing system:
        {\par
        \begin{figure}[H]
            \centering
            \begin{minipage}{0.5\columnwidth}
                \centering
                \includegraphics[width=0.75\textwidth]{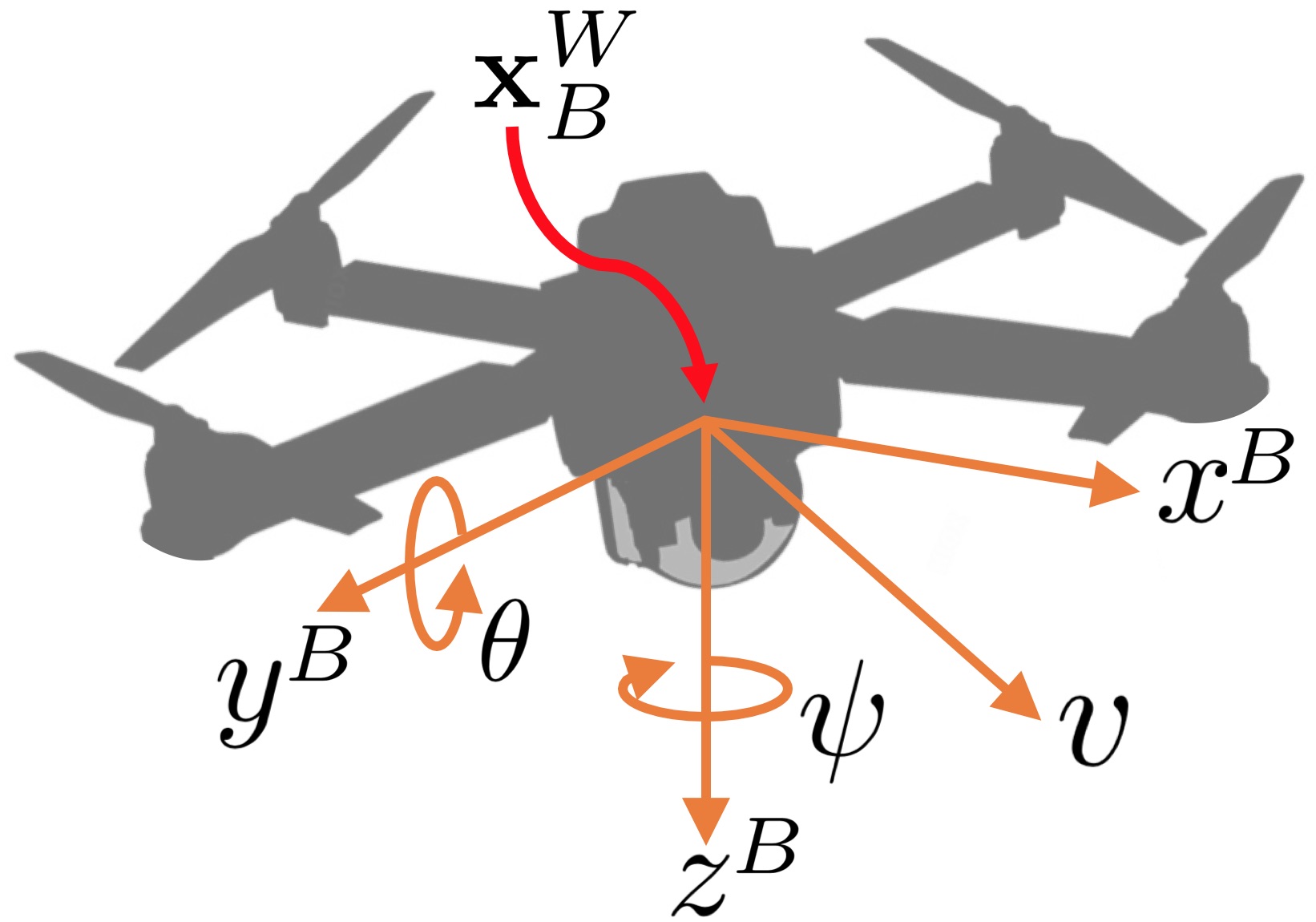}
            \end{minipage}% to prevent a space
            \begin{minipage}{0.5\columnwidth}
                \begin{equation}
                    \begin{aligned} 
                        \dot{x} &= v \; \cos(\yaw) \; \cos(\pitch), \\
                        \dot{y} &= v \; \sin(\yaw) \; \cos(\pitch), \\
                        \dot{z} &= v \; \sin(\pitch), \\
                        \dot{\yaw} &= \yawvel, \\
                        \dot{\pitch} &= \pitchvel, \nonumber
                        %\label{eq:specific_km_all}
                    \end{aligned}
                \end{equation}
                \vspace*{-0.5cm}
            \end{minipage}
        \end{figure}
        \par}
        \noindent where $x$, $y$ and $z$ correspond to the Cartesian coordinates of the system with respect to a predefined reference frame, $\yaw$ and $\pitch$ respectively define the system's orientation around the $z$-axis and $y$-axis,
        $v$ is the vehicle's forward velocity, and $\yawvel$ and $\pitchvel$ are the vehicle's turning rate. Thus, the system's state is defined as $\vx = (x, \; y, \; z, \; \yaw, \; \pitch)^T$, and the system's control input is defined as $\vu = (v, \; \yawvel, \; \pitchvel)^T$.
        
        %\epa{\st{The closed-loop system once applied the dynamic feedback linearisation controller described in \sref{sec:formulation_mot} results with state ${\vz=(x, \; \dot{x}, \; y, \; \dot{y}, \; z, \; \dot{z})^T}$ and desired state ${\vr = (x_r, \; \dot{x}_r, \; y_r, \; \dot{y}_r, \; z_r, \; \dot{z}_r)^T}$.}}
        
        Similar to \cite{hemakumara2013learning}, the Gaussian process describing the \ac{UAV}'s motion model is learnt from simulated data. The training data is extracted from the \ac{UAV}'s simulator producing a varied set of stationary excitations via the control input $\vu$. Relevant control inputs are selected with the above system's model to maximise information on the output system's state $\vx$. 
        
        Further discussion on methods for modelling robots with (partially) unknown dynamics as Gaussian processes is available in \cite{nguyen2009model,jackson2020safety}.
    \section{Gaussian Relationships \label{sec:operators}}
    This appendix summarises the calculation of spatial relationships via the inverse and compound operators. These elemental transformations can be composed to calculate more complex spatial relationships. The interested reader may wish to consult \cite{smith1990estimating} for a more thorough explanation about Gaussian relationships than the brief introduction that follows.
    
    % ===============================
    % ===============================
    % ===============================
    \subsection{Inverse Relationship}
        The inverse relationship $\ominus$ represents the Gaussian relationship $\vx_i^j$ as a function of $\vx_j^i$ as:
        \begin{gather}
            \vx_i^j \coloneqq \ominus \vx_j^i.
        \end{gather}
        
        The first-order estimate of the mean and the covariance of the compounding operation are:
        \begin{align}
            \hat{\vx}_i^j &\approx \ominus \hat{\vx}_j^i, \\
            \cov_{\vx_i^j} &\approx \jacobian_\ominus
            \cov_{\vx_j^i}
            \jacobian_\ominus^T,
        \end{align}
        where
        \begin{gather}
            \jacobian_\ominus \coloneqq 
            \frac{\partial \vx_i^j}{\partial \vx_j^i}.
        \end{gather}
    
    % ===============================
    % ===============================
    % ===============================
    \subsection{Compound Relationship}
        The compounding operation $\oplus$ computes the Gaussian relationship $\vx_k^i$ from two spatial relationships $\vx_j^i$ and $\vx_k^j$ which are arranged head-to-tail as:
        \begin{gather}
            \vx_k^i \coloneqq \vx_j^i \oplus \vx_k^j.
        \end{gather}
        
        The first-order estimate of the mean and the covariance of the compounding operation are:
        \begin{align}
            \hat{\vx}_k^i &\approx \hat{\vx}_j^i \oplus \hat{\vx}_k^j, \\
            \cov_{\vx_k^i} &\approx \jacobian_\oplus
            \begin{bmatrix}
                \cov_{\vx_j^i} & \cov_{\left(\vx_j^i, \vx_k^j\right)} \\
                \cov_{\left(\vx_k^j, \vx_j^i\right)} & \cov_{\vx_k^j}
            \end{bmatrix}
            \jacobian_\oplus^T,
            \label{eq:compounding_covariance}
        \end{align}
        where $\jacobian$ denotes the Jacobian, i.e. the matrix of partial derivatives: 
        \begin{align}
            \jacobian_\oplus &\coloneqq \frac{\partial \vx_j^i \oplus \vx_k^j}{\partial \left(\vx_j^i,\vx_k^j\right)} =
            \frac{\partial \vx_k^i}{\partial \left(\vx_j^i,\vx_k^j\right)} \\
            &=
            \begin{bmatrix}
            \jacobian_{1\oplus} & \jacobian_{2\oplus}
            \end{bmatrix}
            =
            \begin{bmatrix}
            \frac{\partial \vx_k^i}{\partial \vx_j^i} & \frac{\partial \vx_k^i}{\partial \vx_k^j}
            \end{bmatrix}
            .
        \end{align}
        
        If the relationships $\vx_j^i$ and $\vx_k^j$ are independent, i.e. ${\cov_{\left(\vx_j^i, \vx_k^j\right)} = 0}$, \eref{eq:compounding_covariance} can be rewriten as:
        \begin{gather}
            \cov_{\vx_k^i} \approx \jacobian_{1\oplus} \cov_{\vx_j^i} \jacobian_{1\oplus}^T + \jacobian_{2\oplus} \cov_{\vx_k^j} \jacobian_{2\oplus}^T 
        \end{gather}
    %\appendix
\section{$\boldsymbol{\confidencelevel}$-Kernel Construction \label{sec:kernel_construction}}
    % \textcolor{red}{here would be a good part to talk about the size of the robot}
    %https://www.chegg.com/homework-help/questions-and-answers/multivariate-gaussians-general-expression-univariate-gaussian-mean-mu-variance-sigma-2-n-x-q14773913
    % https://en.wikipedia.org/wiki/Confidence_interval#Basic_steps
    % https://en.wikipedia.org/wiki/Cumulative_distribution_function
    % https://en.wikipedia.org/wiki/Quantile_function
    
    A Gaussian distribution $\N(\hat{\vx}, \, \cov_{\vx})$ describing a state's belief~$b$ is continuous and extends over the entire belief space. For the required computations, $b$ is represented on a discrete support $\kernel_\confidencelevel(\cov_{\vx})$, referred to as $\confidencelevel$-kernel, with resolution~$\mapresolution$ and size $\kernelsize_d$ at each dimension~$d$ as defined by:
    \begin{align}
        \kernelsize_d = 2\;\text{ceil}\left(\dfrac{\criticalvalue_\confidencelevel \sigma_d}{\mapresolution}\right) + 1
    \end{align}
    where the kernel size $\kernelsize_d$ is always odd, $\sigma_d$ is the standard deviation of $\cov_{\vx}$ along dimension $d$, and the critical value $\criticalvalue_\confidencelevel$ is computed from the desired confidence level $\confidencelevel \in [0, 1]$ as:
    \begin{align}
        \criticalvalue_\confidencelevel = -\phi^{-1}\left(\dfrac{1}{2}(1 - \confidencelevel)\right)
        \label{eq:critical_value}
    \end{align}
    where $\phi^{-1}(\cdot)$ denotes the quantile function of the \mbox{d-dimensional} Gaussian normal distribution describing the system's belief $b$. \tref{tab:confidence_level_to_critical_value} shows some critical values $\criticalvalue_\confidencelevel$ according to commonly desirable confidence levels $\confidencelevel$ for 1D, 2D and 3D Gaussian distributions.
    
    Noteworthy, the confidence level $\confidencelevel$ involves a trade-off between computational performance and accuracy. On one hand, $\confidencelevel$ bounds the extend of the resulting $\kernel_\confidencelevel(\cdot)$ over the belief space, thus determining the total number of voxels in the kernel and, consequently, having an impact on the computational load of the probabilistic collision checking formulated in \eref{eq:pocc_guarantees}. On the other hand, \eref{eq:pocc_guarantees} introduces a constant conservatism $\confidencelevel$, implying that $\confidencelevel$ must be selected such that ${\confidencelevel > p_\safe}$. Otherwise, the method will not find any valid state. This requirement is implicit in the probabilistic collision checking formulated in \eref{eq:pocc_guarantees}.
    
    The value of each cell $n \in \kernel_\confidencelevel(\cov_{\vx})$ can be drawn from the corresponding Gaussian distribution as $h^{D}\N(\vx \, | \, \hat{\vx}, \, \cov_{\vx})$, where $h^{D}$ is a normalising constant according to the kernel resolution $h$, and $\vx$ is the point coordinate of $n$ referenced at $\hat{\vx}$. For the particular case of a multivariate Gaussian $\N(\hat{\vx}, \, \cov_{\vx})$ with diagonal covariance matrix $\cov_{\vx}$, i.e., its elements can be written as ${\cov_{ij} = \sigma_{i}{^2} \identity_{ij}}$, where $\identity_{ij}$ are the matrix elements of the identity matrix (so ${\identity_{ij} = 0}$ if ${i \neq j}$ and ${\identity_{ij} = 1}$). Then, the multivariate Gaussian with diagonal ${\cov_{ij} = \sigma_{i}{^2} \identity_{ij}}$ factorises into a product of univariate Gaussians as:
    \begin{align} 
        \N(\vx \, | \, \hat{\vx}, \, \cov_{\vx}) = h^{D} \prod_{i=1}^{D} \N(\vx_i \, | \, \hat{\vx}, \, \sigma_{\vx_i}^2) \label{eq:gaussian_factorisation}
    \end{align}
    where for any arbitrary positive definite covariance matrix $\cov_{\vx}$ the resulting distribution is normalised. The property in \eref{eq:gaussian_factorisation} provides a computationally efficient strategy to build any d-dimensional kernel $\kernel(\cdot)$ from 1D Gaussian signals.
    
    It is worth mentioning that the kernel computation can be conducted and stored offline for different kernel sizes. At running time, the planner would uniquely need to retrieve in a look-up table fashion the required kernel. Although this is an option to speed up performance of the presented probabilistic collision checking, the implementation in this work computes the kernels online.
    
    \begin{table}[ht!]
        \centering
        \begin{tabular}{cccccc}
            \toprule
            n-D & \multicolumn{5}{c}{Confidence level $\confidencelevel$ [\%]} \\
            & 85.0 & 90.0 & 95.0 & 99.0 & 99.9 \\
            \cmidrule{2-6}
            1 & 1.4395 & 1.6449 & 1.9600 & 2.5758 & 3.2905\\
            2 & 1.9479 & 2.1460 & 2.4477 & 3.0349 & 3.7169 \\
            3 & 2.3059 & 2.5003 & 2.7955 & 3.3682 & 4.0331 \\
            \bottomrule
        \end{tabular}
        \caption{Critical values $\criticalvalue_\confidencelevel$ computed with \eref{eq:critical_value} subject to different confidence levels $\confidencelevel$ and dimensions $d$ of the Gaussian distribution.}
        \label{tab:confidence_level_to_critical_value}
        % values extracted from https://www.researchgate.net/figure/Magnification-ratios-of-scaled-SDHE-corresponding-to-different-specified-confidence_fig8_273955667 and validated according to the corresponding Gaussian distribution
    \end{table}

    \bibliographystyle{Format/SageH}
    \bibliography{references}

\begin{thebibliography}{65}
\providecommand{\natexlab}[1]{#1}
\providecommand{\url}[1]{\texttt{#1}}
\providecommand{\urlprefix}{URL }
\expandafter\ifx\csname urlstyle\endcsname\relax
  \providecommand{\doi}[1]{DOI:\discretionary{}{}{}#1}\else
  \providecommand{\doi}{DOI:\discretionary{}{}{}\begingroup
  \urlstyle{rm}\Url}\fi

\bibitem[{Agha-Mohammadi et~al.(2014)Agha-Mohammadi, Chakravorty and
  Amato}]{agha2014firm}
Agha-Mohammadi AA, Chakravorty S and Amato NM (2014) {FIRM}: Sampling-based
  feedback motion-planning under motion uncertainty and imperfect measurements.
\newblock \emph{The International Journal of Robotics Research} 33(2):
  268--304.

\bibitem[{Alterovitz et~al.(2007)Alterovitz, Sim{\'e}on and
  Goldberg}]{alterovitz2007stochastic}
Alterovitz R, Sim{\'e}on T and Goldberg KY (2007) The stochastic motion
  roadmap: A sampling framework for planning with markov motion uncertainty.
\newblock In: \emph{Robotics: Science and systems}. pp. 233--241.

\bibitem[{Andert et~al.(2011)Andert, Adolf, Goormann and
  Dittrich}]{andert2011mapping}
Andert F, Adolf F, Goormann L and Dittrich J (2011) Mapping and path planning
  in complex environments: An obstacle avoidance approach for an unmanned
  helicopter.
\newblock In: \emph{2011 IEEE International Conference on Robotics and
  Automation}. IEEE, pp. 745--750.

\bibitem[{Beckers et~al.(2017)Beckers, Umlauft and Hirche}]{beckers2017stable}
Beckers T, Umlauft J and Hirche S (2017) Stable model-based control with
  gaussian process regression for robot manipulators.
\newblock \emph{IFAC-PapersOnLine} 50(1): 3877--3884.

\bibitem[{Bingham et~al.(2010)Bingham, Foley, Singh, Camilli, Delaporta,
  Eustice, Mallios, Mindell, Roman and Sakellariou}]{bingham2010robotic}
Bingham B, Foley B, Singh H, Camilli R, Delaporta K, Eustice R, Mallios A,
  Mindell D, Roman C and Sakellariou D (2010) Robotic tools for deep water
  archaeology: Surveying an ancient shipwreck with an autonomous underwater
  vehicle.
\newblock \emph{Journal of Field Robotics} 27(6): 702--717.

\bibitem[{Blackmore et~al.(2011)Blackmore, Ono and
  Williams}]{blackmore2011chance}
Blackmore L, Ono M and Williams BC (2011) Chance-constrained optimal path
  planning with obstacles.
\newblock \emph{IEEE Transactions on Robotics} 27(6): 1080--1094.

\bibitem[{Carreras et~al.(2015)Carreras, Candela, Ribas, Palomeras, Mag{\'\i},
  Mallios, Vidal, Vidal and Ridao}]{carreras2015testing}
Carreras M, Candela C, Ribas D, Palomeras N, Mag{\'\i} L, Mallios A, Vidal E,
  Vidal {\`E} and Ridao P (2015) {Testing Sparus II AUV, an open platform for
  industrial, scientific and academic applications}.
\newblock \emph{Instrumentation viewpoint} (18): 54--55.

\bibitem[{da~Silva~Arantes et~al.(2019)da~Silva~Arantes, Toledo, Williams and
  Ono}]{da2019collision}
da~Silva~Arantes M, Toledo CFM, Williams BC and Ono M (2019) Collision-free
  encoding for chance-constrained nonconvex path planning.
\newblock \emph{IEEE Transactions on Robotics} 35(2): 433--448.

\bibitem[{Dadkhah and Mettler(2012)}]{dadkhah2012survey}
Dadkhah N and Mettler B (2012) Survey of motion planning literature in the
  presence of uncertainty: Considerations for uav guidance.
\newblock \emph{Journal of Intelligent \& Robotic Systems} 65(1-4): 233--246.

\bibitem[{De~Iaco et~al.(2019)De~Iaco, Smith and Czarnecki}]{de2019learning}
De~Iaco R, Smith SL and Czarnecki K (2019) Learning a lattice planner control
  set for autonomous vehicles.
\newblock \emph{arXiv preprint arXiv:1903.02044} .

\bibitem[{De~Luca et~al.(2000)De~Luca, Oriolo and
  Vendittelli}]{de2000stabilization}
De~Luca A, Oriolo G and Vendittelli M (2000) Stabilization of the unicycle via
  dynamic feedback linearization.
\newblock In: \emph{6th IFAC Symp. on Robot Control}. pp. 397--402.

\bibitem[{Dubins(1957)}]{dubins1957curves}
Dubins LE (1957) On curves of minimal length with a constraint on average
  curvature, and with prescribed initial and terminal positions and tangents.
\newblock \emph{American Journal of Mathematics} 79(3): 497--516.

\bibitem[{Durrant-Whyte and Bailey(2006)}]{durrant2006simultaneous}
Durrant-Whyte H and Bailey T (2006) Simultaneous localization and mapping: part
  i.
\newblock \emph{IEEE robotics \& automation magazine} 13(2): 99--110.

\bibitem[{Frazzoli et~al.(2005)Frazzoli, Dahleh and
  Feron}]{frazzoli2005maneuver}
Frazzoli E, Dahleh MA and Feron E (2005) Maneuver-based motion planning for
  nonlinear systems with symmetries.
\newblock \emph{IEEE transactions on robotics} 21(6): 1077--1091.

\bibitem[{Galceran et~al.(2015)Galceran, Campos, Palomeras, Ribas, Carreras and
  Ridao}]{galceran2015coverage}
Galceran E, Campos R, Palomeras N, Ribas D, Carreras M and Ridao P (2015)
  Coverage path planning with real-time replanning and surface reconstruction
  for inspection of three-dimensional underwater structures using autonomous
  underwater vehicles.
\newblock \emph{Journal of Field Robotics} 32(7): 952--983.

\bibitem[{Hauser and Zhou(2016)}]{hauser2016asymptotically}
Hauser K and Zhou Y (2016) Asymptotically optimal planning by feasible
  kinodynamic planning in a state--cost space.
\newblock \emph{IEEE Transactions on Robotics} 32(6): 1431--1443.

\bibitem[{Hemakumara and Sukkarieh(2013)}]{hemakumara2013learning}
Hemakumara P and Sukkarieh S (2013) Learning uav stability and control
  derivatives using gaussian processes.
\newblock \emph{IEEE Transactions on Robotics} 29(4): 813--824.

\bibitem[{Hern{\'a}ndez et~al.(2016)Hern{\'a}ndez, Moll, Vidal, Carreras and
  Kavraki}]{hernandez2016planning}
Hern{\'a}ndez JD, Moll M, Vidal E, Carreras M and Kavraki LE (2016) Planning
  feasible and safe paths online for autonomous underwater vehicles in unknown
  environments.
\newblock In: \emph{Intelligent Robots and Systems (IROS)}. IEEE, pp.
  1313--1320.

\bibitem[{Hern{\'a}ndez et~al.(2019)Hern{\'a}ndez, Vidal, Moll, Palomeras,
  Carreras and Kavraki}]{hernandez2019planning}
Hern{\'a}ndez JD, Vidal E, Moll M, Palomeras N, Carreras M and Kavraki LE
  (2019) Online motion planning for unexplored underwater environments using
  autonomous underwater vehicles.
\newblock \emph{Journal of Field Robotics} 36(2): 370--396.
\newblock \doi{10.1002/rob.21827}.

\bibitem[{Ho et~al.(2018)Ho, Sodhi, Teixeira, Hsiao, Kusnur and
  Kaess}]{ho2018virtual}
Ho BJ, Sodhi P, Teixeira P, Hsiao M, Kusnur T and Kaess M (2018) Virtual
  occupancy grid map for submap-based pose graph slam and planning in 3d
  environments.
\newblock In: \emph{2018 IEEE/RSJ International Conference on Intelligent
  Robots and Systems (IROS)}. IEEE, pp. 2175--2182.

\bibitem[{Hornung et~al.(2013)Hornung, Wurm, Bennewitz, Stachniss and
  Burgard}]{hornung13auro}
Hornung A, Wurm KM, Bennewitz M, Stachniss C and Burgard W (2013) {OctoMap}: An
  efficient probabilistic {3D} mapping framework based on octrees.
\newblock \emph{Autonomous Robots} \doi{10.1007/s10514-012-9321-0}.

\bibitem[{Hover et~al.(2012)Hover, Eustice, Kim, Englot, Johannsson, Kaess and
  Leonard}]{hover2012advanced}
Hover FS, Eustice RM, Kim A, Englot B, Johannsson H, Kaess M and Leonard JJ
  (2012) Advanced perception, navigation and planning for autonomous in-water
  ship hull inspection.
\newblock \emph{The International Journal of Robotics Research} 31(12):
  1445--1464.
\newblock \doi{10.1177/0278364912461059}.

\bibitem[{Hsu et~al.(1997)Hsu, Latombe and Motwani}]{hsu1997path}
Hsu D, Latombe JC and Motwani R (1997) Path planning in expansive configuration
  spaces.
\newblock In: \emph{Robotics and Automation, 1997. Proceedings., 1997 IEEE
  International Conference on}, volume~3. IEEE, pp. 2719--2726.

\bibitem[{Huynh et~al.(2012)Huynh, Karaman and Frazzoli}]{huynh2012incremental}
Huynh VA, Karaman S and Frazzoli E (2012) An incremental sampling-based
  algorithm for stochastic optimal control.
\newblock In: \emph{Robotics and Automation (ICRA), 2012 IEEE International
  Conference on}. IEEE, pp. 2865--2872.

\bibitem[{Jackson et~al.(2020)Jackson, Laurenti, Frew and
  Lahijanian}]{jackson2020safety}
Jackson J, Laurenti L, Frew E and Lahijanian M (2020) Safety verification of
  unknown dynamical systems via gaussian process regression.
\newblock \emph{arXiv preprint arXiv:2004.01821} .

\bibitem[{Janson et~al.(2018)Janson, Schmerling and Pavone}]{janson2018monte}
Janson L, Schmerling E and Pavone M (2018) Monte carlo motion planning for
  robot trajectory optimization under uncertainty.
\newblock In: \emph{Robotics Research}. Springer, pp. 343--361.

\bibitem[{Katz and Some(2003)}]{katz2003nasa}
Katz DS and Some RR (2003) Nasa advances robotic space exploration.
\newblock \emph{Computer} 36(1): 52--61.

\bibitem[{Kavraki et~al.(1996)Kavraki, Svestka, Latombe and
  Overmars}]{kavraki1996probabilistic}
Kavraki LE, Svestka P, Latombe JC and Overmars MH (1996) Probabilistic roadmaps
  for path planning in high-dimensional configuration spaces.
\newblock \emph{IEEE transactions on Robotics and Automation} 12(4): 566--580.

\bibitem[{LaValle and Kuffner~Jr(2001)}]{lavalle2001randomized}
LaValle SM and Kuffner~Jr JJ (2001) Randomized kinodynamic planning.
\newblock \emph{The International Journal of Robotics Research} 20(5):
  378--400.

\bibitem[{LaValle and Sharma(1995)}]{lavalle1995framework}
LaValle SM and Sharma R (1995) A framework for motion planning in stochastic
  environments: modeling and analysis.
\newblock In: \emph{Proceedings of 1995 IEEE International Conference on
  Robotics and Automation}, volume~3. IEEE, pp. 3057--3062.

\bibitem[{Le~Ny and Pappas(2009)}]{le2009trajectory}
Le~Ny J and Pappas GJ (2009) On trajectory optimization for active sensing in
  gaussian process models.
\newblock In: \emph{Proceedings of the 48h IEEE Conference on Decision and
  Control held jointly with 2009 28th Chinese Control Conference}. IEEE, pp.
  6286--6292.

\bibitem[{Li et~al.(2016)Li, Littlefield and Bekris}]{li2016asymptotically}
Li Y, Littlefield Z and Bekris KE (2016) Asymptotically optimal sampling-based
  kinodynamic planning.
\newblock \emph{The International Journal of Robotics Research} 35(5):
  528--564.

\bibitem[{Lin and Saripalli(2014)}]{lin2014path}
Lin Y and Saripalli S (2014) Path planning using 3d dubins curve for unmanned
  aerial vehicles.
\newblock In: \emph{2014 international conference on unmanned aircraft systems
  (ICUAS)}. IEEE, pp. 296--304.

\bibitem[{Liu and Ang(2014)}]{liu2014incremental}
Liu W and Ang MH (2014) Incremental sampling-based algorithm for risk-aware
  planning under motion uncertainty.
\newblock In: \emph{2014 IEEE International Conference on Robotics and
  Automation (ICRA)}. IEEE, pp. 2051--2058.

\bibitem[{Luders et~al.(2010)Luders, Kothari and How}]{luders2010chance}
Luders B, Kothari M and How J (2010) Chance constrained {RRT} for probabilistic
  robustness to environmental uncertainty.
\newblock In: \emph{AIAA guidance, navigation, and control conference}. p.
  8160.

\bibitem[{Luders et~al.(2013)Luders, Karaman and How}]{luders2013robust}
Luders BD, Karaman S and How JP (2013) Robust sampling-based motion planning
  with asymptotic optimality guarantees.
\newblock In: \emph{AIAA Guidance, Navigation, and Control (GNC) Conference}.
  p. 5097.

\bibitem[{Luna et~al.(2014)Luna, Lahijanian, Moll and Kavraki}]{Luna:AAAI:2014}
Luna R, Lahijanian M, Moll M and Kavraki LE (2014) Optimal and efficient
  stochastic motion planning in partially-known environments.
\newblock In: \emph{The Twenty-Eighth AAAI Conference on Artificial
  Intelligence}. Quebec City, Canada, pp. 2549--2555.

\bibitem[{Majumdar and Tedrake(2017)}]{majumdar2017funnel}
Majumdar A and Tedrake R (2017) Funnel libraries for real-time robust feedback
  motion planning.
\newblock \emph{The International Journal of Robotics Research} 36(8):
  947--982.

\bibitem[{Meyer et~al.(2012)Meyer, Sendobry, Kohlbrecher, Klingauf and von
  Stryk}]{2012simpar_meyer}
Meyer J, Sendobry A, Kohlbrecher S, Klingauf U and von Stryk O (2012)
  Comprehensive simulation of quadrotor uavs using ros and gazebo.
\newblock In: \emph{3rd Int. Conf. on Simulation, Modeling and Programming for
  Autonomous Robots (SIMPAR)}. p. to appear.

\bibitem[{Moravec and Elfes(1985)}]{moravec1985high}
Moravec H and Elfes A (1985) High resolution maps from wide angle sonar.
\newblock In: \emph{Proceedings. 1985 IEEE International Conference on Robotics
  and Automation}, volume~2. IEEE, pp. 116--121.

\bibitem[{Morgenthal and Hallermann(2014)}]{morgenthal2014quality}
Morgenthal G and Hallermann N (2014) Quality assessment of unmanned aerial
  vehicle (uav) based visual inspection of structures.
\newblock \emph{Advances in Structural Engineering} 17(3): 289--302.

\bibitem[{Murphy et~al.(2008)Murphy, Tadokoro, Nardi, Jacoff, Fiorini, Choset
  and Erkmen}]{murphy2008search}
Murphy RR, Tadokoro S, Nardi D, Jacoff A, Fiorini P, Choset H and Erkmen AM
  (2008) Search and rescue robotics.
\newblock \emph{Springer handbook of robotics} : 1151--1173.

\bibitem[{Nguyen-Tuong et~al.(2009)Nguyen-Tuong, Seeger and
  Peters}]{nguyen2009model}
Nguyen-Tuong D, Seeger M and Peters J (2009) Model learning with local gaussian
  process regression.
\newblock \emph{Advanced Robotics} 23(15): 2015--2034.

\bibitem[{Pairet et~al.(2018)Pairet, Hern{\'a}ndez, Lahijanian and
  Carreras}]{pairet2018uncertainty}
Pairet {\`E}, Hern{\'a}ndez JD, Lahijanian M and Carreras M (2018)
  Uncertainty-based online mapping and motion planning for marine robotics
  guidance.
\newblock In: \emph{2018 IEEE/RSJ International Conference on Intelligent
  Robots and Systems (IROS)}. IEEE, pp. 2367--2374.

\bibitem[{Palmieri et~al.(2016)Palmieri, Koenig and Arras}]{palmieri2016rrt}
Palmieri L, Koenig S and Arras KO (2016) Rrt-based nonholonomic motion planning
  using any-angle path biasing.
\newblock In: \emph{2016 IEEE International Conference on Robotics and
  Automation (ICRA)}. IEEE, pp. 2775--2781.

\bibitem[{Park et~al.(2012)Park, Pan and Manocha}]{park2012itomp}
Park C, Pan J and Manocha D (2012) Itomp: Incremental trajectory optimization
  for real-time replanning in dynamic environments.
\newblock In: \emph{Twenty-Second International Conference on Automated
  Planning and Scheduling}.

\bibitem[{Patil et~al.(2012)Patil, Van Den~Berg and
  Alterovitz}]{patil2012estimating}
Patil S, Van Den~Berg J and Alterovitz R (2012) Estimating probability of
  collision for safe motion planning under gaussian motion and sensing
  uncertainty.
\newblock In: \emph{2012 IEEE International Conference on Robotics and
  Automation}. IEEE, pp. 3238--3244.

\bibitem[{Pini{\'e}s and Tard{\'o}s(2007)}]{pinies2007scalable}
Pini{\'e}s P and Tard{\'o}s JD (2007) Scalable {SLAM} building conditionally
  independent local maps.
\newblock In: \emph{Intelligent Robots and Systems, 2007. IROS 2007.} IEEE, pp.
  3466--3471.

\bibitem[{Pini{\'e}s and Tard{\'o}s(2008)}]{pinies2008large}
Pini{\'e}s P and Tard{\'o}s JD (2008) Large-scale slam building conditionally
  independent local maps: Application to monocular vision.
\newblock \emph{IEEE Transactions on Robotics} 24(5): 1094--1106.

\bibitem[{Pivtoraiko and Kelly(2011)}]{pivtoraiko2011kinodynamic}
Pivtoraiko M and Kelly A (2011) Kinodynamic motion planning with state lattice
  motion primitives.
\newblock In: \emph{2011 IEEE/RSJ International Conference on Intelligent
  Robots and Systems}. IEEE, pp. 2172--2179.

\bibitem[{Plaku(2015)}]{plaku2015region}
Plaku E (2015) Region-guided and sampling-based tree search for motion planning
  with dynamics.
\newblock \emph{IEEE Transactions on Robotics} 31(3): 723--735.

\bibitem[{Plaku et~al.(2010)Plaku, Kavraki and Vardi}]{plaku2010motion}
Plaku E, Kavraki LE and Vardi MY (2010) Motion planning with dynamics by a
  synergistic combination of layers of planning.
\newblock \emph{IEEE Transactions on Robotics} 26(3): 469.

\bibitem[{Prats et~al.(2012)Prats, P{\'e}rez, Fern{\'a}ndez and
  Sanz}]{prats2012open}
Prats M, P{\'e}rez J, Fern{\'a}ndez JJ and Sanz PJ (2012) An open source tool
  for simulation and supervision of underwater intervention missions.
\newblock In: \emph{Intelligent Robots and Systems (IROS), 2012 IEEE/RSJ
  International Conference on}. IEEE, pp. 2577--2582.

\bibitem[{Reeds and Shepp(1990)}]{reeds1990optimal}
Reeds J and Shepp L (1990) Optimal paths for a car that goes both forwards and
  backwards.
\newblock \emph{Pacific journal of mathematics} 145(2): 367--393.

\bibitem[{Scherer et~al.(2008)Scherer, Singh, Chamberlain and
  Elgersma}]{scherer2008flying}
Scherer S, Singh S, Chamberlain L and Elgersma M (2008) Flying fast and low
  among obstacles: Methodology and experiments.
\newblock \emph{The International Journal of Robotics Research} 27(5):
  549--574.

\bibitem[{Smith et~al.(1990)Smith, Self and Cheeseman}]{smith1990estimating}
Smith R, Self M and Cheeseman P (1990) Estimating uncertain spatial
  relationships in robotics.
\newblock In: \emph{Autonomous robot vehicles}. Springer, pp. 167--193.

\bibitem[{Strawser and Williams(2018)}]{strawser2018approximate}
Strawser D and Williams B (2018) Approximate branch and bound for fast,
  risk-bound stochastic path planning.
\newblock In: \emph{2018 IEEE International Conference on Robotics and
  Automation (ICRA)}. IEEE, pp. 7047--7054.

\bibitem[{{\c{S}}ucan et~al.(2012){\c{S}}ucan, Moll and
  Kavraki}]{sucan2012the-open-motion-planning-library}
{\c{S}}ucan IA, Moll M and Kavraki LE (2012) The {O}pen {M}otion {P}lanning
  {L}ibrary.
\newblock \emph{{IEEE} Robotics \& Automation Magazine} 19(4): 72--82.
\newblock \doi{10.1109/MRA.2012.2205651}.

\bibitem[{Sun et~al.(2015)Sun, Patil and Alterovitz}]{sun2015high}
Sun W, Patil S and Alterovitz R (2015) High-frequency replanning under
  uncertainty using parallel sampling-based motion planning.
\newblock \emph{IEEE Transactions on Robotics} 31(1): 104--116.

\bibitem[{Van Den~Berg et~al.(2011)Van Den~Berg, Abbeel and
  Goldberg}]{van2011lqg}
Van Den~Berg J, Abbeel P and Goldberg K (2011) Lqg-mp: Optimized path planning
  for robots with motion uncertainty and imperfect state information.
\newblock \emph{The International Journal of Robotics Research} 30(7):
  895--913.

\bibitem[{Vidal et~al.(2019)Vidal, Moll, Palomeras, Hern{\'a}ndez, Carreras and
  Kavraki}]{vidal2019online}
Vidal E, Moll M, Palomeras N, Hern{\'a}ndez JD, Carreras M and Kavraki LE
  (2019) Online multilayered motion planning with dynamic constraints for
  autonomous underwater vehicles.
\newblock In: \emph{2019 International Conference on Robotics and Automation
  (ICRA)}. IEEE, pp. 8936--8942.

\bibitem[{Webb and van~den Berg(2012)}]{dustin2012}
Webb DJ and van~den Berg J (2012) Kinodynamic {RRT}*: Optimal motion planning
  for systems with linear differential constraints.
\newblock \emph{CoRR} abs/1205.5088.

\bibitem[{Whitcomb et~al.(2000)Whitcomb, Yoerger, Singh and
  Howland}]{whitcomb2000advances}
Whitcomb L, Yoerger DR, Singh H and Howland J (2000) Advances in underwater
  robot vehicles for deep ocean exploration: Navigation, control, and survey
  operations.
\newblock In: \emph{Robotics Research}. Springer, pp. 439--448.

\bibitem[{Yguel et~al.(2008)Yguel, Aycard and Laugier}]{yguel2008update}
Yguel M, Aycard O and Laugier C (2008) Update policy of dense maps: Efficient
  algorithms and sparse representation.
\newblock In: \emph{Field and Service Robotics}. Springer, pp. 23--33.

\bibitem[{Youakim et~al.(2020)Youakim, Cieslak, Dornbush, Palomer, Ridao and
  Likhachev}]{youakim2020multirepresentation}
Youakim D, Cieslak P, Dornbush A, Palomer A, Ridao P and Likhachev M (2020)
  Multirepresentation, multiheuristic a* search-based motion planning for a
  free-floating underwater vehicle-manipulator system in unknown environment.
\newblock \emph{Journal of Field Robotics} .

\end{thebibliography}

    %\pagebreak
    %\input{Sections/miscellaneous.tex}
\end{document}